\documentclass[sigconf,nonacm=true,screen=true,review=false,urlbreakonhyphens=true]{acmart}

\usepackage[utf8]{inputenc}
\usepackage{multirow} 
\usepackage{graphicx} 
\usepackage{dsfont}
\usepackage{amsmath,amsthm,amsfonts}
\usepackage{algorithm,algpseudocode}

\title{Fair mapping}
\date{}

\begin{document}

\author{S\'ebastien Gambs}
\email{gambs.sebastien@uqam.ca}
\affiliation{%
  \institution{Universit\'e du Qu\'ebec \`a Montr\'eal}
  \city{Montr\'eal}
  \country{Canada}}

\author{Rosin Claude Ngueveu}
\email{ngueveu.rosin-claude@courrier.uqam.ca}
\affiliation{%
  \institution{Universit\'e du Qu\'ebec \`a Montr\'eal}
  \city{Montr\'eal}
  \country{Canada}}

\begin{abstract}
To mitigate the effects of undesired biases in models, several approaches propose to pre-process the input dataset to reduce the risks of discrimination by preventing the inference of sensitive attributes.
Unfortunately, most of these pre-processing methods lead to the generation a new distribution that is very different from the original one, thus often leading to unrealistic data. 
As a side effect, this new data distribution implies that existing models need to be re-trained to be able to make accurate predictions.
To address this issue, we propose a novel pre-processing method, that we coin as fair mapping, based on the transformation of the distribution of protected groups onto a chosen target one (the privileged distribution), with additional privacy constraints to prevent the inference of sensitive attributes.
More precisely, we leverage on the recent works of the Wasserstein GAN framework to achieve the optimal transport of data points coupled with a discriminator enforcing the protection against attribute inference.
Our proposed approach, preserves the interpretability of data and can be used without any operational overhead (model re-traning and hyperparameters search).
In addition, our approach can be specialized to model existing state-of-the-art approaches, thus proposing a unifying view on these methods.
Finally, several experiments on real and synthetic datasets demonstrate that our approach is able to hide the sensitive attributes, while limiting the distortion of the data and improving the fairness on subsequent data analysis tasks.
\end{abstract}


\newcommand{\ber}{BER}
\newcommand{\sac}{SAcc}
\newcommand{\demoPar}{DemoParity}
\newcommand{\eqgap}{EqGap}
\newcommand{\consistency}{yNN}
\newcommand{\distance}{Dist}
\newcommand{\fid}{Fid}
\newcommand{\optfid}{Optim}
\newcommand{\classification}{\textit{Pc}}
\newcommand{\diversity}[0]{Diversity}
\newcommand{\catdam}[0]{Cat}
\newcommand{\accuracy}[0]{Acc}
\newcommand{\baserate}[0]{BRate}

\newcommand{\attgan}[0]{AttGAN}
\newcommand{\wgan}[0]{WGAN}
\newcommand{\gansan}[0]{GANSan}
\newcommand{\dirm}[0]{DIRM}
\newcommand{\fairmapping}[0]{FM}
\newcommand{\fairmappingdd}[0]{FM2D}

\newcommand{\protectedmapped}[0]{P.M}
\newcommand{\allmapped}[0]{A.M}

\newcommand{\origdecision}[0]{Y.O}
\newcommand{\mappeddecision}[0]{Y.M}

\newcommand{\update}[0]{red}
\newcommand{\noupdate}[0]{black}
\def\ros#1{\textcolor{red}{#1}}

\newcommand{\approachsize}[0]{0.2}
\def\ros#1{\textcolor{black}{#1}}
\renewcommand{\update}[0]{black}

\maketitle

\section{Introduction}
\label{intro}

In recent years, machine learning models have become ubiquitous, from their use on personal devices such as our phone to banking applications in which they are used for the assessment of applicants of a credit card~\cite{bazarbash2019fintech,wang2020comparative,bao2019integration}, for summarizing the data into valuable information that helps in the retention of customers or for fraud detection~\cite{chitra2013data}.
Machine learning models are also deployed in health settings, in which they assist medical practitioners in the early detection of diseases or psychological disorders~\cite{zhan2018using,torous2018smartphones}.

These models usually require a significant amount of data to be trained.
Unfortunately, the data on which they are trained often incorporate historical or social biases (\emph{i.e.}, data influenced by historically biased human decisions or social values)~\cite{pessach2020algorithmic,suresh2019framework,mehrabi2019survey}.
This can lead to a form of representational harm (\emph{e.g.}, stereotype) towards a particular group of the population.
Thus, if the machine learning model integrate this bias in its structure and is deployed in high stakes decision systems in which its predictions are put into effect~\cite{Dickson2016,DeBrusk2018,Kent2019}, this will only exacerbate its effect and could lead to discrimination.

To mitigate the impact of negative and undesired biases, several approaches for fairness enhancement consider pre-processing the dataset~\cite{aivodji2021local,kamiran2012data,calmon2017optimized}, by transforming it such that the underlying discrimination is removed.
These researches often tackle the problem of \emph{indirect discrimination}, that arises when the model makes decision by exploiting correlations between sensitive attributes (\emph{e.g.}, gender or religious beliefs~\cite{holmes2005anti,donohue2007antidiscrimination}) and the rest of the profile.

Unfortunately, these approaches are often designed only for datasets in which there are two groups to considered for discrimination, making it usable only for a single binary sensitive attribute.
In addition, the transformed dataset is often obtained by modifying the distributions of each group, mapping them towards an intermediate one that introduce enough distortion to satisfy the fairness constraints or find a new representation of the data.
The drawback is that such an intermediate distribution might be one that does not exist in the real world, and does not necessarily account for all existing correlations in the dataset.

For instance, consider a dataset that exhibits a strong correlation between the sensitive attribute \emph{ethnic origin}, and the attribute \emph{education level} and a low correlation between the \emph{degree obtained} and \emph{ethnic origin}.
By merging distributions based on the sensitive attribute to enhance fairness, we might observe some discrepancies in which an individual in \textit{High-School} end up with a \textit{Master} or \textit{PhD} degree.
Such transformation could diminish the usability of the approach, as the difference in statistics could lead to misinterpretation when the dataset is used in subsequent analysis tasks (\emph{i.e}, deciding for school resources' allocation).
Finally, those approaches relies on the sensitive attribute to guide the transformation process, as the model has to know the particular group membership of a given input data point before applying the transformation~\cite{feldman2015certifying}.

In this paper, we propose an approach called Fair Mapping ($\fairmapping{}$) to address the above issues.
Our approach is inspired from the \attgan{} framework~\cite{he2019attgan} and leverages the \textit{Wasserstein Generative Adversarial Networks} (\wgan{})~\cite{arjovsky2017wasserstein} to perform the optimal transport (\emph{i.e.}, the one having the lowest cost in terms of modifications to transform a distribution into another one) of an input data distribution onto a chosen target one, to which we add privacy constraints to prevent the inference of the sensitive attribute.
By transporting the input distributions (\emph{e.g.}, the distributions of the protected groups as defined by the values of the sensitive attributes) onto a target one (which we called the privileged group distribution), our approach preserves the realistic aspect of the dataset as the target distribution is known to exist.
Moreover, the transformation does not require the knowledge of the sensitive attribute at test time and can also be used as a discrimination detection mechanism in which one could observe if a model prediction would change had a given individual been in another group, (as shown in~\cite{black2020fliptest} and in situation testing~\cite{luong2011k}).
Finally, the optimal transport on a chosen target distribution has the additional benefit of introducing the minimal amount of modifications necessary to prevent the inference of sensitive attributes while not modifying members of the target group.

Our contributions can be summarized as follows:
\begin{itemize}
\item We introduce a preprocessing technique called Fair Mapping that solves the problem of indirect discrimination by preventing the inference of the sensitive attribute.
In contrast to prior works that requires both the transformation of the privileged and protected groups\ros{~\cite{xu2019fairgan+,yu2021fair,zhou2021improving,calmon2017optimized}}, our approach only transform the protected groups.
\item By transforming the data onto a chosen distribution, our mapping preserves the realistic aspect of the resulting dataset, as the target distribution exists in practice and is known.
Also, this dataset remains interpretable in the sense that it does not change the representation space.
This means that our approach can be used in the context of discrimination discovery or for counterfactuals, as done previously in FlipTest~\cite{black2020fliptest}.
Similarly, if a model is already trained for a specific task on the original dataset, our approach does not require the re-training of the model, which removes the computational overhead of many preprocessing techniques.

\item Once the model is trained, our approach does not require access to the sensitive attribute to apply the transformation, making it suitable for situations in which users do not want to disclose their group membership.
In addition, Fair Mapping only introduces the necessary modifications for a data point to change its group membership if it does not belong to the target group, while members of that target group are unmodified.
Furthermore, our approach can be seen as a generalised version of other state-of-the-art approaches that leverage the adversarial training to prevent the inference of the sensitive attribute.
\item Finally, our experiments on synthetic and real world datasets demonstrate that Fair Mapping is able to prevent the inference of the group membership while reducing the discrimination as measured with standard fairness metrics such as the demographic parity and equalized odds.

\end{itemize}

The outline of the paper is a follows.
First, in Section \ref{sec:background}, we introduce the notations as well as the main fairness metrics used throughout this paper.
Afterwards in Section \ref{sec:related_work}, we review the related work on fairness-enhancement methods that are the most relevant to our work, before describing our approach Fair Mapping in Section~\ref{sec:fairmapping}. 
\ros{Then in Section~\ref{sec:soaupdate}, we demonstrate the generic aspect of the fair mapping framework by deriving other approaches from the literature as well as extending state-of-the-art approaches from it.
Finally, we present our experimental setting as well as the obtained results in Section~\ref{sec:experiments} before concluding in Section~\ref{sec:conclusion}.}

\section{Background}
\label{sec:background}

In this section, we introduce the notations as well as the fairness metrics used throughout the rest of this paper.

\subsection{Notations}


We consider a dataset $R$ composed of $N$ records, each described by $d$ attributes.
Each record $r_{i}$ ($i \in \{1, \ldots, N\}$) corresponds to the profile of a particular individual and is composed of
three types of attributes:
\begin{itemize}

\item \emph{Sensitive} attributes $S_{j}$ ($j \in \{1, \ldots, k\}$) are those through which discrimination may arise.
They can either be \emph{binary} or \textit{multivalued}.
In our context, we consider for instance the following attributes as sensitive ones: \emph{gender}, \emph{ethnic origin} and \textit{age}.

\item A binary \emph{decision} attribute $Y$, typically representing the prediction made by a machine learning model, which will be used in decision-making process (\emph{e.g.}, being accepted or rejected for a loan application).

\item Other \emph{non-discriminatory} attributes $A$, which are used by the machine learning model to predict $Y$ and can be correlated with any of the sensitive attributes $S_{j}$ or their combinations.
\end{itemize}

\ros{As the dataset can be composed of several sensitive attributes, multiple groups can be identified leading to the notion of intersectional fairness~\cite{foulds2020intersectional}.
Among these groups, we will consider the group with the highest advantage among all groups (\emph{e.g}, the highest rate of positive decisions) as the privileged group (\emph{e.g.}, \emph{White-Male}), which we coined as $R_{priv}$. 
In contrast, the other groups defined by the combination of sensitive attributes will be considered as protected ones (\emph{e.g.}, \emph{White-Female}, \emph{Black-Male} and \emph{Black-Female}) and denoted by $R_{prot}$.
Slightly abusing the notation, we will transform the dataset by combining multiple sensitive attributes into a single one $S$ that can take $k$ different values. 
In this attribute, the privileged group will be associated with the sensitive value $s_{1}$ and a datapoint $r_{i}$ from such group will be denoted as $r_{i}^{1}$.
All other values $j \in \{2, \ldots, k\}$ will represent the protected groups.}

To ease the reading of the paper, Table~\ref{tab:notation-table} summarizes the notations used throughout the paper. 
\begin{table}[h!]
    \centering
    \caption{Summary of notations and symbols.}
    \label{tab:notation-table}
    \resizebox{\columnwidth}{!}{%
        \begin{tabular}{p{0.25\linewidth} | p{0.8\linewidth}}
            \hline
            Symbol & Definition  \\
            $R$  &  Dataset \\
            $N$  &  Number of records (\emph{i.e.}, rows or individuals) of the dataset\\
            $d$  &  Number of attributes of the dataset \\
            $S_{j}$  & $j^{th}$ sensitive attribute \\
            $r_{i}^{j}$ ($i \in \{1, \ldots, N\}$, $j \in \{1, \ldots, k\}$)  &  Description of the $i^{th}$ row or individual $i$ of $R$, with sensitive attribute $S_{j}$\\
            $Y$  & Decision attribute \\
            $A$  & Attributes that are neither sensitive of the decision one \\
            $R_{priv}$ & Subset of data from the privileged group \\
            $R_{prot}$ & Subset of data from the protected group \\
            $\ber$  & Balanced Error Rate \\
            $\ber{}_{rc_{prv}}$ & $\ber$ computed with the reconstructed privileged group and the protected group\\
            $\ber{}_{og_{prv}}$ & $\ber$ computed with the original privileged group and the protected group\\
            $\sac$  & Sensitive attribute prediction Accuracy \\
            $\demoPar$ & Demographic Parity \\
            $\eqgap$  & Equalised Odds \\
            $\classification{}_{prot}$ & Proportion of individual in the protected group predicted as belonging to the privileged group\\
            $\fid$  & Fidelity \\
            $\fid_{priv}$  & Fidelity computed with respect to the privileged group \\
            \attgan  & Facial Attribute Editing by Only Changing What You Want~\cite{he2019attgan} \\
            \wgan  & Wasserstein generative adversarial networks~\cite{arjovsky2017wasserstein} \\
            \gansan  & Generative Adversarial Network Sanitization~\cite{aivodji2021local} \\
            \dirm  & Disparate Impact Remover~\cite{feldman2015certifying} \\
            \fairmapping  & FairMapping (our approach) \\
            \fairmappingdd  & FairMapping with 2 discriminators \\
            $C$  & Classifier in $\fairmapping{}$ outputting a vector of probabilities $ C(r_{i}^{t}) = \langle c_{i}^{1}, \ldots, c_{i}^{t}, \ldots, c_{i}^{k}  \rangle \: ; \: c_{i}^{j} = C(r_{i}^{t})^{j} \: ; \: \sum_{j=1}^{k} c_{i}^{j} = 1$ for a given point $r_{i}^{t}$ with sensitive attribute $S_{t}$ \\
            $D$  & Discriminator for the sensitive attribute in $\fairmapping{}$ with similar to $C$ \\
            $D_{std}$  & Standard GAN discriminators, distinguishing real from generated data \\
            $G_{Enc}$  & Encoder of $\fairmapping{}$ \\
            $G_{Dec}$  & Decoder of $\fairmapping{}$ \\
            
            $\lambda_{C}$  & Classification weight \\
            $\lambda_{D}$  & Protection weight \\
            $\lambda_{D_{std}}$ & GAN weight \\
            $\lambda_{R}$  & weight of the identity operation \\
            \hline
        \end{tabular}
        }
\end{table}
\subsection{Fairness metrics}

In the fairness literature, there are mainly three families of fairness notions: \emph{individual fairness},
\emph{group fairness} 
 and \emph{fairness through the prevention of inference of the sensitive attribute}.

In a nutshell, \emph{individual fairness} states that \emph{similar} individuals should receive a \emph{similar treatment}.
This notion, also called fairness through awareness~\cite{dwork2012fairness}, requires the specification of a similarity measure between pairs of individuals.
Various approaches have been proposed to define and select the similarity measure in different
contexts~\cite{zemel2013learning,yurochkin2020sensei,ruoss2020learning,xue2020auditing}.
The level of individual fairness of a particular machine learning model can be quantified using metrics such as the
\emph{consistency}~\cite{zemel2013learning}.

In contrast to individual fairness, \emph{group fairness} relies more on the statistical properties of groups in the dataset~\cite{binns2020apparent}.
More precisely, group fairness is generally satisfied when a statistical measure is equalized across groups defined by the sensitive attributes and can generally be computed from the confusion matrix associated to a particular machine learning model.
We refer the interested reader to the following survey~\cite{verma2018fairness} reviewing some of these metrics.
Among them, the \emph{demographic parity} \demoPar{}~\cite{berk2018fairness} and the \emph{equalized odds}
\eqgap{}~\cite{hardt2016equality} are among the most used ones:
\begin{equation}
    \demoPar{} = \lvert P(\hat{Y} \mid S=s_1) - P(\hat{Y} \mid S=s_0) \rvert \leq \epsilon
    \label{eq:dempar}
\end{equation}
\begin{equation}
    \eqgap{} = Pr(\hat{Y} = 1 \mid S=s_0, Y=y) - Pr(\hat{Y} = 1 \mid S=s_1, Y=y) \leq \epsilon
    \label{eq:eqod}
\end{equation}
in which $Y$ and $\hat{Y}$ are the original decision attribute and the prediction made by a classifier, and $S$ is the sensitive attribute.
The Demographic parity ensures that groups as defined by the sensitive attribute(s) (\emph{e.g.}, in Equation~\ref{eq:dempar}, groups are defined by the values $S=s_0$ and $S=s_1$) receive almost the same rate of positive (or negative) decisions, up to a tolerance threshold $\epsilon$.
As for the Equalized odds, the group fairness is achieved if the true positive rates (respectively the false positive rates) of groups does not differ by a difference greater than the threshold $\epsilon$.

\emph{Fairness through the prevention of the inference of the sensitive attribute} is another family of fairness notion introduced in~\cite{feldman2015certifying,zemel2013learning,aivodji2021local} that rely on the fact that discrimination arises due to \emph{the possibility of inferring the values of sensitive attributes}.
Unfortunately, removing sensitive attributes from the dataset is not enough as they might be correlated with other attributes.
For instance, $S$ could potentially be expressed as $f(A, Y)$, for $f$ a non-linear function.
To prevent such inferential risk, this notion of fairness aims at modifying the original dataset such that the sensitive attribute is hidden by removing it as well as its correlations with other attributes.

\ros{To quantify the protection of the sensitive attribute, most approaches in this category rely on the accuracy of prediction of
\sac{} and the \emph{Balanced Error Rate} (\ber{})~\cite{feldman2015certifying,xu2018fairgan}.}
\begin{equation}
    \begin{split}
        &\ber{}(f(A, Y), S)  \\
        &= \dfrac{1}{\lvert S \rvert} \left(\sum_{i=1}^{\lvert S \rvert}P(f(A, Y) \neq s_{i} \mid S = s_{i})\right)\\
        & = \dfrac{1}{\lvert S \rvert}\left(\sum_{i=1}^{\lvert S \rvert} 1 - P(f(A, Y) = s_{i} \mid S = s_{i})\right)\\
        & = 1 - \dfrac{1}{\lvert S \rvert}\left(\sum_{i=1}^{\lvert S \rvert} P(f(A, Y) = s_{i} \mid S = s_{i})\right)\\
        & = 1 - \dfrac{1}{k}\left(\sum_{i=1}^{k} P(f(A, Y) = s_{i} \mid S = s_{i})\right),
    \end{split}
    \label{eq:ber}
\end{equation}
in which $S$ is the sensitive attribute, $Y$ the decision one, $A$ the rest of the attributes characterizing a record and $k$ the number of sensitive groups.
If $f$ is a model that, for each data point $r_{i}^{t}$ output its probability of belonging to the $k$ different groups, $f(r_{i}^{t}) = \langle p_{i}^{1}, \ldots, p_{i}^{t}, \ldots, p_{i}^{k}  \rangle$, $t$ being the correct group, we expect $p_{i}^{t} = f(r_{i}^{t})^{t} = P(f(r_{i}^{t}) = s_{t}) = 1$. 
In this situation, the $\ber{}$ can be written as in Equation~\ref{eq:ber-simp}.
\begin{equation}
    \ber{}(f(A, Y), S) = 1 - \dfrac{1}{k}\sum_{j=1}^{k}\left(\dfrac{\sum_{r_{i}^{j} \in R \mid S=s_{j}} f(r_{i}^{j})^{j}}{\| S=s_{j} \| }\right).
    \label{eq:ber-simp}
\end{equation}
\ros{where $\| S=s_{j} \| $ define the size of the group with sensitive value $S=s_{j}$. More precisely, this generalized balanced error rate measures the error rate in predicting the sensitive attribute in each group before averaging those error rates for all groups.
Its value ranges from $0$ (\emph{i.e.}, perfect prediction as it is equivalent to an accuracy of $1$) to $1$ with $\dfrac{k-1}{k}$ being the optimal protection value, as it means that the predictor behave essentially as a random guess within each group (\emph{cf.} Appendix~\ref{ber-demo}).}

\ros{Ideally, the protection of the sensitive attribute requires maximizing the \ber{} to its optimal value while minimizing the accuracy up to the proportion of the most present group in the dataset. 
Note that in the a binary sensitive attribute scenario, achieving a $\ber{}$ of $1$ (which correspond to an accuracy of $0$) do not protect the sensitive information as the prediction could be simply reversed to achieve perfect prediction. 
In the multi-valued setting, the $\ber{}$ becomes harder to interpret, as the value of the $\ber{}$ does not gives insight on the model behaviour in each group. Similarly, achieving the lowest accuracy in the multi-attribute setting does not necessarily means that the predictions correspond to some random guessing: the classifier could predict $s_{2}$ instead of $s_{1}$, $s_{3}$ instead of $s_{2}$ and $s_{1}$ instead of $s_{3}$. The predictions preserves some information about the group membership. 
Thus, both metrics are necessary to quantify the protection achieved by a particular approach. 
For instance, in Section~\ref{sec:adult2-org-results}, we show that a higher value of $\ber{}$ does not necessarily means a lower accuracy $\sac{}$ in the multi-valued setup.}
Later in Section~\ref{sec:group-fair-priv}, we also discuss the relationship between the group fairness and the fairness by the prevention of inference of the sensitive attribute.

With these fairness notions defined, in the next section we review the existing works in the literature that are the closest to ours.

\section{Related Work}
\label{sec:related_work}

There is a growing body of literature aiming at the detection of discrimination and the enhancement of fairness in machine learning models and decision-making processes.
These approaches can be categorised into three main families: data pre-processing~\cite{aivodji2021local,kamiran2012data,calmon2017optimized}, model in-processing~\cite{kearns2018preventing,kamishima2012fairness,agarwal2018reductions} and model post-processing~\cite{lohia2019bias}. 
The pre-processing algorithms transformed the input data by producing a modified version satisfying some fairness constraints or preventing the inference of the sensitive attribute, making the generated dataset useful for many subsequent data analysis tasks.
In-processing techniques, also known as algorithms modification approaches, modify the learning process of the algorithm by introducing some fairness constraints that the algorithm has to satisfy during its training, in addition to other standard objectives such as accuracy.
Such constraints are often introduced as a form of regularization of the original objective~\cite{kamishima2012fairness,bechavod2017learning}. 
Finally, the post-processing approaches consist in modifying the standard learning algorithm output to reach the fairness requirements.

In this research, we will mainly discuss techniques related to the fairness pre-processing family, as our approach falls within the same category.
More precisely, the related work section is divided into three parts, the first one discussing the related work on fairness pre-processing approaches as well as a few methods that are designed to handle multiple groups.
The second part briefly reviews the Generative Adversarial Networks (GANs) and the attribute transfer approaches mostly used in the context of facial features editing, as our approach is inspired from those researches and relies on the same procedure to transport and transform a given datapoint onto a target group.
The last part will introduce the related work on research focusing on the application of the optimal transport theory in the context of GANs as well as fairness.

We refer the interested reader to the following surveys~\cite{pessach2020algorithmic,mehrabi2019survey,suresh2019framework,rabaeyfairness,verma2018fairness,jones2020metrics} that propose a more in-depth review of fairness definitions, approaches, and metrics used in this burgeoning research field.

\subsection{Fairness pre-processing approaches}

To the best of our knowledge, there are three seminal works that proposed a pre-processing technique to enhance fairness : \emph{Data preprocessing}~\cite{kamiran2012data}, \emph{Learning Fair Representations (LFR)}~\cite{zemel2013learning} and \emph{Disparate Impact Remover (DIRM)}~\cite{feldman2015certifying}.

\emph{Seminal works.} In~\cite{kamiran2012data}, the authors proposed the \emph{suppression} and the \emph{massaging} of the dataset as a pre-processing technique to reduce the discrimination in the input training dataset.
The suppression consists in finding and removing attributes that are highly correlated with the sensitive one, while the massaging changes the label of some individuals based on a ranking obtained with a Naïve Bayes classifier. 
The rank of each profile is computed based on the probability of a naïve Bayes classifier to assign a positive decision to this particular profile.
Unfortunately, the suppression can introduce a high level of distortion in the dataset and is highly dependent on the existing form of correlations, while the massaging does not necessarily prevent the inference of the sensitive attributes.
Indeed, it is possible that more complex classifiers can still discriminate even tough the fairness constraints are met.

In LFR~\cite{zemel2013learning}, the authors learn a fair representation of the dataset by mapping each point of the dataset onto a set of prototypes.
Each prototype has an equal probability of representing either the privileged group or the protected one.
In addition, the mapping preserves the existing correlations with the decision attribute while maintaining enough information that enables the reconstruction of the original dataset.
A careful choice of prototypes is required in this approach as they act as the representatives of the population and the approach relies on the distance between each profile and the set of prototypes.
As a consequence, a set of prototypes closer to the privileged group will induce a lower quality of data reconstruction, as the approach will compensate for their proximity with the privileged population.
In addition, a lower number of prototypes might improve the protection of the sensitive attribute but lower the reconstruction quality, while a larger number leads to a better reconstruction at the cost of the protection of the sensitive attribute.

Following the same direction, in~\cite{feldman2015certifying} the authors propose a framework that builds the conditional distribution of each of the dataset attributes based on the sensitive attribute before translating them towards a median distribution.
Unfortunately, this approach is a linear application that does not consider complex correlations that arises with the combinations of several attributes to infer the sensitive one.
Moreover, as it requires the construction of the cumulative distribution function and cannot take into account categorical attributes.

\emph{Advanced approaches.} More advanced techniques have also appeared in recent years.
In~\cite{calmon2017optimized}, Calmon, Wei, Vinzamuri, Ramamurthy and Varshney have learned an optimal randomized mapping for removing group-based discrimination while limiting the distortion introduced at profiles and distributions levels to preserve utility.
The definition of penalty weights for any non-acceptable transformation, makes the approach complex to define, as the relationship between attributes is often not fully understood.
The overall approach is therefore difficult to use in practice, especially on a dataset with very large number of attribute-values combinations, which would imply the definition of large number of constraints making the problem infeasible.
Furthermore, the meaning of each of the given penalties might also be difficult to grasp and there could be numerous non-acceptable transformations.

In~\cite{aivodji2021local}, the authors proposed an approach called \gansan{} that, from a given input profile, produces a new one that still lives in the same representational space as the original one, from which the inference of the value of the sensitive attribute is prevented while also limiting the distance between the original and transformed profiles.
This approach can successfully prevent the inference of the sensitive attributes in various scenarios, thus enhancing the fairness of a model learnt on this data. 
One of the strength of the approach is the ability to locally protect the user data, without having to rely on a centralized solution to protect the sensitive feature.
Similarly to \gansan, our approach preserves the space of the original data while introducing a limited amount of modifications to hide the sensitive attribute.
However, while \gansan{} leaves the choice of the intermediate distribution to the protection mechanism, our approach transport the data distribution onto a chosen target group, thus increasing its interpretability.
Moreover, our framework does not require the sensitive attribute as input during its test phase, in contrast to \gansan.

Another preprocessing approach that preserves the space of original data is FairGan~\cite{xu2018fairgan}, which generates a new data distribution in order to protect the user sensitive information.
FairGan+~\cite{xu2019fairgan+} extends this framework with the introduction of a classifier during the training procedure, whose objective is to maximize the accuracy with respect to a chosen task.
Both approaches differ from ours by the fact that they do not allow for a wide range of subsequent data analysis tasks but are rather used to build a fair classification mechanism (by training a classifier on the generated new distribution).
As a consequence, they cannot be used to protect new input profiles locally on-the-fly. Similarly to \gansan, the intermediate distribution is chosen by the generation process.

\emph{Modification to the machine learning pipeline.} As the data preprocessing approaches consist in the modification of the dataset to satisfy specific fairness constraints, it also encompasses preprocessing techniques that modifies a dataset as part of the machine learning training pipeline.

In FairPreprocessing~\cite{yu2021fair}, the author presents an alternative version of the reweighting algorithm~\cite{kamiran2012data}, assigning weights to different records in the dataset based on their respective group membership and class labels to learn a fairer machine learning model.
This alternative can handle multiple attributes and also consider the group size when assigning weights in the dataset.
Assuming that the dataset is composed of fair decisions, they also showed that a machine learning model exhibit unfairness due to the difference in class size (mostly minimizing the error on the most seen groups) and class labels (\emph{i.e.}, the model favours the conditional distribution of the decision that will appear most frequently).
Similarly to the feature selection approach, this research focuses on building of a fairer machine learning, and thus does not necessarily prevent the inference of the sensitive attribute, as ours does.

The authors of~\cite{zhou2021improving} have proposed a lossless data debiasing technique that oversamples the underrepresented groups such that the discrepancy between the privileged distribution and the underrepresented one is bounded.
The oversampling is carried out in the protected group by generating synthetic samples with either more positive outcomes if the proportion of positive decisions is higher in the privileged group or negative samples otherwise.
As a consequence, only the proportion of the less represented group is augmented.
This approach was also motivated by the fact that fairness enhancement techniques induce a loss of information, often trading off accuracy with fairness, and necessitate the fairness definition and metric to be hard-coded.
This approach, even though it is theoretically justified and does not introduce modification of the original set, provides no guarantees with respect to the protection of the user sensitive information from undesirable inferences.


\emph{Handling multiple attributes.} There also exist a few preprocessing approaches that are designed to handle multiple attributes.
For instance, the framework defined in~\cite{creager2019flexibly} exploits the disentangled nature of representations obtained with variational auto-encoders (VAEs) to produce a new representation of the input dataset in which the sensitive attributes are decorrelated between them as well as with respect to other attributes.
Once generated, if one or more sensitive attributes are considered not suitable for a task, the user can remove their associated latent dimension with the guarantee that other dimensions do not include information about this attribute.
Just as other approaches that are based on VAEs, this approach, even though it can handle many sensitive attributes, changes the representation of the dataset in contrast to ours that preserves the original data space.

In \textit{generating fair universal representation}~\cite{kairouz2019censored}, the authors leverages the adversarial learning approach to generate new representations of the data that prevent the inference of multiple sensitive attributes.
Their approach consists in an encoder limiting the amount of distortion up to a certain threshold $\tau$, while the adversary is trained on the encoder output to maximize its inference of the sensitive attributes.
In addition, they have proven that if the encoder achieves statistical parity (or another fairness notion) in the prediction of the sensitive attribute, a classifier trained on the obtained representation to predict a different task would also achieve the same fairness property with respect to the sensitive attributes.
As their approach relies on the limitation of the distortion up to a threshold $\tau$, the choice of this threshold is difficult to make as a standard user might not be able to interpret its value, as they could vary across different datasets.
For instance, we expect that a high threshold would induce a perfect protection while minimizing the utility, while a very low bound might not be enough to prevent the sensitive inference.
Thus, the value of $\tau$ plays an important role in the preservation of the original representational space.
In particular, a large value would nearly produce outputs corresponding to a new representation of the dataset.



\emph{Privacy protection mechanisms.} Preventing the inference of the sensitive attribute directly echoes some privacy researches.
For instance, early work such as~\cite{Ruggieri2014} have shown that anonymization methods developed to achieve the $t$-closeness privacy model~\cite{t_closeness} can be used as a preprocessing approach to control discrimination as there is a close relationship between $t$-closeness and group fairness.

In a more recent work~\cite{boutet2021dysan}, the authors have proposed a dynamic sanitization mechanism whose objective is to prevent the inference of sensitive attributes from data collected through sensors devices while maintaining most of the data utility.
From a set of pre-trained sanitizing models, the system dynamically selects the model achieving the desired trade-off between utility and privacy for each incoming batch of data and for each user, in contrast to having a single model for all.
Similarly,~\cite{pittaluga2019learning} have designed a procedure based on adversarial training to learn a private encoding of images while allowing the prediction of desirable features.
The focus here is that explicitly given sensitive information can be derived from several cues such as the background, the foreground, and several aspects of the image.
The protection mechanism must account for such cues by training an encoder outputting a representation from which an adversary cannot infer the sensitive attribute.
Another task classifier can also be used to augment the utility preserved.

Another research that leverages generative adversarial networks~\cite{goodfellow2014generative} of generative models is~\cite{romanelli2019generating}, which describes a mechanism to create a dataset preserving the original data space, while obtaining an optimal privacy protection in the context of location privacy.
This mechanism minimizes the mutual information between the sensitive attribute and the prediction made on the decision attribute by a classifier, while respecting a bound on the utility of the dataset.

\subsection{Transferring attributes using generative adversarial networks}
\label{sec:attgan}

Since their conception, Generative Adversarial Networks (GANs)~\cite{goodfellow2014generative} have been applied in a variety of contexts.
The success of such approach resides in their ability to model complex distributions (\emph{i.e.}, pictures or videos), which can be used for various purposes, such as sampling the distribution or translating the learned distribution into another.
The main idea of GANs is to learn the distribution from which a set of data points have been sampled by using two different models with antagonist objectives.
More precisely, the generator model aims at transforming a given random noise into a data point that follows the distribution to learn, while the discriminator model is used to assess the correctness of the transformation by quantifying the closeness of the transformed data point to the known samples of the distribution.
From seminal works such as CycleGan~\cite{zhu2017unpaired}, different approaches have emerged to learn to transfer an image attributes~\cite{zhang2019multi}.
In particular, several works have been proposed in the literature to transfer properties specific to one group onto others, such as sunglasses, haircut, eyes or colour on pictures that do not originally have one.
We refer the reader to the surveys on GANs and facial attributes manipulations~\cite{zheng2020survey,jabbar2021survey} for a more detailed analysis.

One notable work in this direction is \attgan{}~\cite{he2019attgan}, 
whose objective is to learn the minimum amount of modifications needed to add a feature to an input image while retaining most of its original features unperturbed.
For instance, it could correspond to a change in the hair colour of an individual in a picture while preserving the identity and pose of that individual.
\attgan{} \ros{is composed of several models, including an encoder, which produces a latent representation capturing most of the information about the features to edit or change, as well as a decoder, which produces the final image with the desired features edited. 
In addition, a classifier is use to verify whether the edition has taken place on the final image, thus ensuring that the feature predicted by the classifier corresponds to the desired value of the feature (\emph{e.g.}, the new hair color). 
In addition, a discriminator, which is trained to predict whether a given image is coming from the original distribution with no features edited or from the generator output, 
also ensures that the image produced containing the new edited features belong to the same distribution as the original data. 
The use of the discriminator is also justified by the fact that the edited image does not have any known ground truth to which the output of the decoder can be compared to. 
Thus, if no features are edited, the decoder should reproduce the original image through the reconstruction process.
\attgan{} claims that its feature editing procedure introduces the minimal amount the information loss in the latent representation.
Such property is appealing to our context, as our objective is to limit the amount of perturbations required to achieve the mapping, thus transferring the properties of a known distribution with a limited loss.}

\ros{Our approach shares some similarities with \attgan{}, since it transfers the properties of the privileged group onto the protected groups using similar models, in the sense that this transfer can be seen as an edit of some features of the protected group.
Nonetheless, there are major differences that distinguish our approach from \attgan{}. 
First, our objective is to prevent inference of the sensitive attribute. 
Thus, even though we ``\textit{edit}'' some features of the protected group, we also prevent the sensitive attribute to be inferred. 
Second, our approach uses a single model for the encoder and decoder, which reduces the information loss in our procedure.
In addition, the classifier is pre-trained instead of being trained in conjunction with other models as done in \attgan{}. The pre-training step reduces the computational cost of our approach by reducing the training time and the memory cost of training the classifier. 
Finally, while \attgan{} edit features from any group to another (\emph{e.g.}, the protected group to the privileged group and vice-versa), our approach is only interested in the transformation of the protected group.}


\subsection{Optimal transport}
\label{sec:opt-tranport}

\ros{There are several related works that have applied the optimal transport theory in the context of GANs.
The pioneering work in this direction is the Wasserstein GANs (WGAN)~\cite{arjovsky2017wasserstein} in which the authors use the Earth Mover (EM) distance (also called Wasserstein-1 distance) to transport one distribution onto another.
The EM distance can be defined as the minimum cost in terms of movements of probability mass of transforming a given distribution into another (hence the term optimal transport).
However, the computation of this minimum is computationally expensive~\cite{cuturi2013sinkhorn,solomon2015convolutional,arjovsky2017wasserstein} ($O(N^3logN)$ complexity for a $N$-bins histogram~\cite{shirdhonkar2008approximate}).
Thus, the authors have optimized an approximation of the EM and proved that the optimized version considered is sound.
Such approximation requires that the function measuring the distance between the two given distributions is $k$-Lipschitz. The $k$-Lipschitz property enforce a limit on how fast the function $f$ can change, and is defined in equation~\ref{eq:k-lipschiz} for two values $z_{1}$ and $z_{2}$
\begin{equation}
    \mid f(z_{1}) - f(z_{2}) \mid \leq K \mid z_{1} - z_{2} \mid
    \label{eq:k-lipschiz}
\end{equation}
More precisely, the problem that aim to solve \wgan{} is to find the infimum of Equation~\ref{eq:em-problem}, which corresponds to estimate the joint distribution $\gamma$ among the set of all possible joint distribution $\Pi(p_x, p_\theta)$ that minimise the \textit{efforts} (cost of transport $\times$ transport distance) required to transform the distribution $p_\theta$ into the real data distribution $p_x$.}
\begin{equation}
    \underset{\theta}{argmin} W(p_{x}, p_{z}) = \underset{\theta}{argmin} \underset{\gamma \in \Pi(p_x, p_\theta)}{inf} \underset{(x, y) \sim \gamma}{\mathbb{E}} [\mid\mid x - y \mid\mid_{2}]
    \label{eq:em-problem}
\end{equation}
As the problem is intractable, the authors relied on the Kantorovich-Rubinstein duality~\cite{edwards2011kantorovich}, which transforms Equation~\ref{eq:em-problem} into Equation~\ref{eq:em-kanton-dual}. 
\begin{equation}
    \begin{split}
        W(p_{x}, p_{z}) &= \underset{\gamma \in \Pi(p_x, p_\theta)}{inf} \underset{(x, y) \sim \gamma}{\mathbb{E}} [\mid\mid x - y \mid\mid_{2}]\\
                        &= \underset{\mid\mid f \mid\mid_{L} \leq 1}{sup} \left[\underset{x \sim p_x}{\mathbb{E}}[f(x)] - \underset{y \sim p_\theta}{\mathbb{E}}[f(y)] \right]
    \end{split}
    \label{eq:em-kanton-dual}
\end{equation}
\ros{Thus, solving the $EM$ problem amounts to finding the supremum $f$ under the Lipschitz constraints ($\mid\mid f \mid\mid_{L} \leq 1$), which can be carried out by parametrising the function $f$ as a neural network and ensuring through various mechanisms that $f$ belongs to the set of Lipschitz functions.
For instance, the original paper used the clipping of gradient of $f$ to reach the Lipschitz constraints.}
\ros{Since many other approaches have been developed to improve WGAN, either by improving on the Lipschitz constraint~\cite{gulrajani2017improved,liu2019wasserstein} or by changing the space of functions to consider when solving the original problem, thus avoiding the restrictive $k$-Lipschitz constraint~\cite{wu2018wasserstein}.}

\ros{Our approach is inspired from WGAN and its improvements.
However, while the application of optimal transport minimizes the cost of moving one distribution onto another, it does not ensure the unpredictability of the sensitive attributes since the mapped data points could be located on a specific portion or modes of the target data space.
In our approach, we address this issue by adding a constraint on the protection of the sensitive attribute.}

In another work~\cite{gordaliza2019obtaining}, the authors have proposed a fairness repair scheme, which also relies on the optimal transport theory to protect the sensitive attribute. 
The approach prevents the inference of the sensitive attribute by translating the distributions onto their Wasserstein barycentre.
Their approach, which is the closest work to ours, mainly differs by the choice of the target distribution.
Indeed, their approach mapped the data onto an intermediate distribution, which ensures the minimal cost in terms of displacements, while ours transports the data onto a chosen target distribution.
Mapping the data into a known distribution ensures that the data corresponds to realistic datapoints, since the target distribution will be chosen to represent realistic data of a specific group.
In addition, it makes the transformation more interpretable and thus easier to understand.

Another notable work that is also based on optimal transport and WGAN is FlipTest~\cite{black2020fliptest}.
The objective of this approach is to uncover subgroups discrimination by leveraging the optimal transport of each data point to their corresponding version in another group.
In other words, for a given profile whose sensitive attribute is $S=s_i$, the corresponding version of that profile in the group with sensitive attribute value $S=s_i$ is identified for $i\neq j$.
Using such mapping, a classification mechanism exhibits discrimination if, for an individual, the outcome obtained using the original and the mapped version of the profile are different.
FlipTest was focused exclusively on discrimination detection, and it is not clear how to adapt it easily to prevent the inference of sensitive attributes. 
In contrast, our objective is to achieve fairness by mapping each profile to a target group \emph{while} hiding the sensitive attributes.

\color{\update}
\subsection{Relationship with causal fairness}
As mentioned, our approach aims at improving the fairness by suppressing the information about the sensitive attribute in a dataset through a mapping of the data onto a known target distribution. 
The mapping onto the target distribution corresponds the same process that can be used to generate counterfactuals, which corresponds to the profiles from one distribution map to their equivalent in a different distribution. 
As such, a connection can be made between our approach and techniques developed in the causal fairness literature. 

More precisely, the causal fairness literature explores how to improve the fairness of machine learning models through causal graphs. 
One of the assumption here is the existence of an underlying data structure explaining the generative process of the data, and the objective is to satisfy fairness metrics derived from the graph such as the total effect~\cite{xu2019achieving}.
For instance, causal fairness aims to ensure that the predicted decision of an individual is invariant to its group membership. 
In other words, given an individual profile and its counterfactual version, a classifier is causally fair if the predicted decision with the original profile is equal to the predicted decision with the counterfactual. 
To achieve this property, some approaches rely on the training of the classifier on a causally fair dataset. 

For example in~\cite{xu2019achieving}, the causally fair dataset is obtained by the leveraging the \textit{CausalGANs}~\cite{kocaoglu2017causalgan} framework, which is a network of GANs that are built to mimic the given causal graph. 
This network of GANs enables the easier generation of counterfactuals by fixing the value of the sensitive attribute. 
Afterwards, the fairness constraints are enforced by using a second networks of GANs ensuring that the distribution of the decision attribute obtained with the original data is indistinguishable from the distribution obtained with the counterfactual data. 
In another more recent work, \cite{kim2021counterfactual} leverages the Variational AutoEncoders framework to build the fair dataset by enforcing the fair generation of the decision distribution through minimization of the squared difference between the original decision distribution and the decision distribution of counterfactuals. 
Both approaches can also be used for counterfactual data generation.

While our approach also generates counterfactuals, our objectives are different from those in the causal fairness literature. 
In particular, since our objective is the protection of the sensitive attribute, our approach is not designed to handle causal fair metrics and objectives.
\color{\noupdate}
In the following section, we describe our approach, Fair Mapping, which is inspired from the AttGan and Fliptest frameworks described above, with some additional features to increase the privacy and utility.
\section{Fair mapping}
\label{sec:fairmapping}

In this section, we introduce our preprocessing approach, called \textit{Fair Mapping}, whose objective is to learn to transform any given input distribution of the dataset into a chosen target one. 
To be precise, our objective is to protect the sensitive attribute(s) by learning a mapping function that transforms the data point from any group (protected groups or the privileged one) into a version that belong to the target distribution, but from which it is difficult to infer the sensitive attribute(s).
Once trained, we envision two potential use cases of the resulting mapping function.
\begin{itemize}
\item The mapping function can be used by a centralized data curator to provide the same treatment of a specific group to other individuals in the dataset, thus reducing the risk of differential treatment across groups.
\item The user could apply locally the mapping function to sanitize his profile before publication, to ensure that the sensitive attributes are protected from attribute inference attack while also benefiting from the same treatment (privileges but also possibly disadvantages) as the privileged group.
In this situation, the mapping function could be provided by an independent entity to users concerned about the misuse of their data or who fear that they might be subject to discrimination due to their group membership.
\end{itemize}

The objective of our approach differs from the fairness preprocessing techniques existing in the literature mentioned in Section~\ref{sec:related_work} on several points.
First, our approach has to map all datapoints to the chosen privileged distribution, ensuring that all groups benefit from the same treatment, in contrast to mapping the data onto an intermediate distribution close to the median one. 
As a consequence, the transformation preserves the realistic aspect of the data since the privileged distribution is one that exists in the real world, in contrast to the intermediate one that neither represents the protected groups nor the privileged one.
Additionally, the mapping onto a known distribution provides a mean to validate the transformation process, by verifying that the data from the privileged distribution are not modified significantly through our mapping procedure. 
Finally, the transformed data points can also be used for discrimination detection by observing whether a given decision would change had an individual of the protected group been in the privileged distribution (as carried out in FlipTest~\cite{black2020fliptest}). 

\subsection{Overview}
\label{subsec:fair:overview}
As mentioned in the introduction, our objective is to learn a \emph{mapping function} that can be applied to different groups of the dataset such that the amount of modifications introduced by the transformation is minimal, all transformed datapoints belong to the privileged distribution (thus, share the same privileges), and such that the sensitive attribute cannot be inferred. 
These objectives can respectively be translated into the following properties of \textit{identity}, \textit{transformation} and \textit{protection}.

\begin{enumerate}
\item \emph{Identity property}. Ideally, the transformation should not modify the profiles of users that already belong to privileged distribution $R_{priv}$.
In fact, as the objective is to map any given datapoint of the dataset onto the privileged distribution by finding its corresponding \textit{privileged} version, the transformation of a datapoint $r_i$ from the privileged distribution that yields the least amount of modification is the data $r_i$ itself. 
Based on this observation, the identity property can be used to measure the quality of data mapping, for two main reasons.
First, any optimal mapping that produces the least amount of modifications over the whole dataset will also produce the least amount of modifications in the mapping of the privileged group. 
Second, the converse of the previous assertion, which would be that \textit{the transformation with the least amount of modifications on the privileged group will also provide the least amount of modifications on the protected groups}, is not necessarily true. This is due to the fact that the protected groups are not identical to the privileged one, as such, obtaining the least amount of modification on the privileged group will only give a general idea of the amount of modifications we can expect on the protected group (since both data shares the same structure), but not the exact value. 
Nevertheless, as the privileged version of a data point from a protected group that requires the least amount of modifications is not known \emph{a priori}, it cannot be accessed to compute the quality of the mapping to ensure that we have obtained the best transformation possible. 
Thus, we have to rely on a surrogate to evaluate our transformation quality.

\item \emph{Transformation property}. \emph{From the original data distribution perspective, a transformed data point should be predicted as part of the privileged group.}
In practice, this property means that any classifier trained on the original data to infer the sensitive attribute should predict that any transformed datapoint belongs to the privileged group distribution. 
As the identity property ensures that the privileged group is not modified, this property could be limited to the protected groups data $R_{prot}$. Thus, any transformed data point from a protected group should be predicted as belonging to the target group distribution.
If this property is achieved, models already trained on the original data for a specific task do not need to be retrained when using the transformed data. 
Indeed, as such models are trained on datasets that already contain the privileged distribution. 
Thus, by mapping all the data points onto the privileged distribution, we transform the data onto a distribution already seen by the trained models, thus requiring no further changes.

\item \emph{(Privacy) protection property}. \emph{The sensitive attribute should be hidden.}
Assuming that there is more than one protected group, the mapping of those groups onto the privileged distribution should produce an output in which all protected groups are indistinguishable. 
In fact, mapping the protected distributions onto the privileged one is not enough to ensure the protection of the sensitive attribute. 
To illustrate this, consider the situation in which the privileged distribution corresponds to a bimodal Gaussian distribution (\emph{i.e}, a Gaussian distribution with two peaks). 
A transformation that exclusively maps the first protected group only onto the first mode of privileged distribution mode while mapping the other protected groups onto the second mode privileged distribution would satisfy the transformation objective (as every single mode is part of the bimodal distribution), but a classifier would still able to build a decision frontier between both modes (\emph{cf.} Figure~\ref{fig:incorrect-mapping}).

\begin{figure}[h!]
        \includegraphics[width=1\linewidth, height=0.1\textheight]{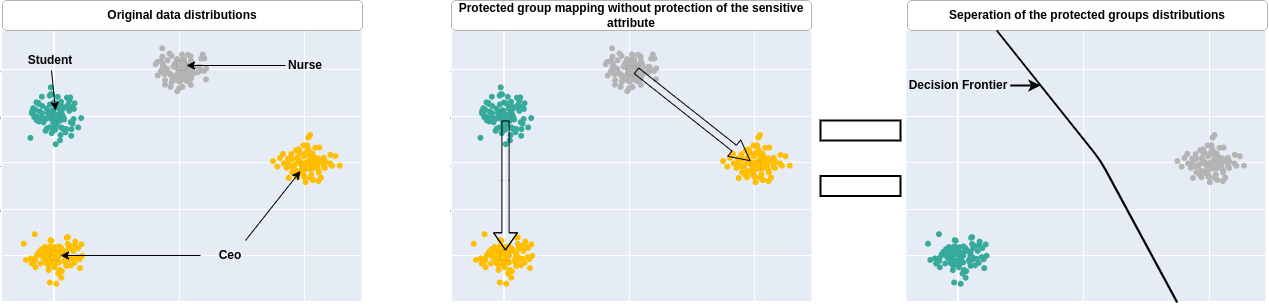}
    \caption{Mapping of the protected groups (\textit{nurse} and \textit{students}) onto the privileged distribution \textit{ceo}. The transformed data belong to the target distribution, but the sensitive attribute is not protected. 
    }
    \label{fig:incorrect-mapping}
\end{figure}
Similarly, a transformation mechanism that maps all protected groups onto the first mode of the privileged distribution would ensure that all protected groups are indistinguishable from each other.
However, they would still be distinguishable from the second mode of the privileged distribution, and thus distinguishable from the privileged distribution, which means that a classifier could also build a decision frontier between the transformed protected and the privileged distributions. 

The protection property aims to prevent these situations by removing any dissimilarities between the privileged distribution and the protected groups transformed distribution. 
In the previous example, the protection ensures that each of the protected group distribution is also bimodal Gaussian with the same statistics as the privileged distribution (\emph{cf.} Figure~\ref{fig:correct-mapping}).
\begin{figure}[h!]
        \includegraphics[width=0.5\linewidth, height=0.15\textheight]{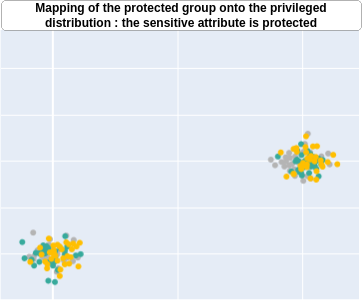}
    \caption{Mapping of the protected groups (\textit{nurse} and \textit{students}) onto the privileged distribution \textit{ceo}, with the sensitive attribute protected. 
    }
    \label{fig:correct-mapping}
\end{figure}
\end{enumerate}

Our approach \fairmapping{} is inspired by \attgan{} and leverages the \wgan{} to achieve our objectives. 
The building blocks of our approach are the transformation model $G_{w}$, the discriminator $D$, the critic $D_{std}$ and the classifier $C$. 
The use of the \wgan{} (instantiated in our framework with the two players game between $G_{w}$ and $D_{std}$) ensures that only the modification necessary for the transformation are introduced during the transformation of the protected data, as shown in Section~\ref{sec:opt-tranport}. 

\begin{figure}[ht]
        \includegraphics[width=1\linewidth, height=\approachsize\textheight]{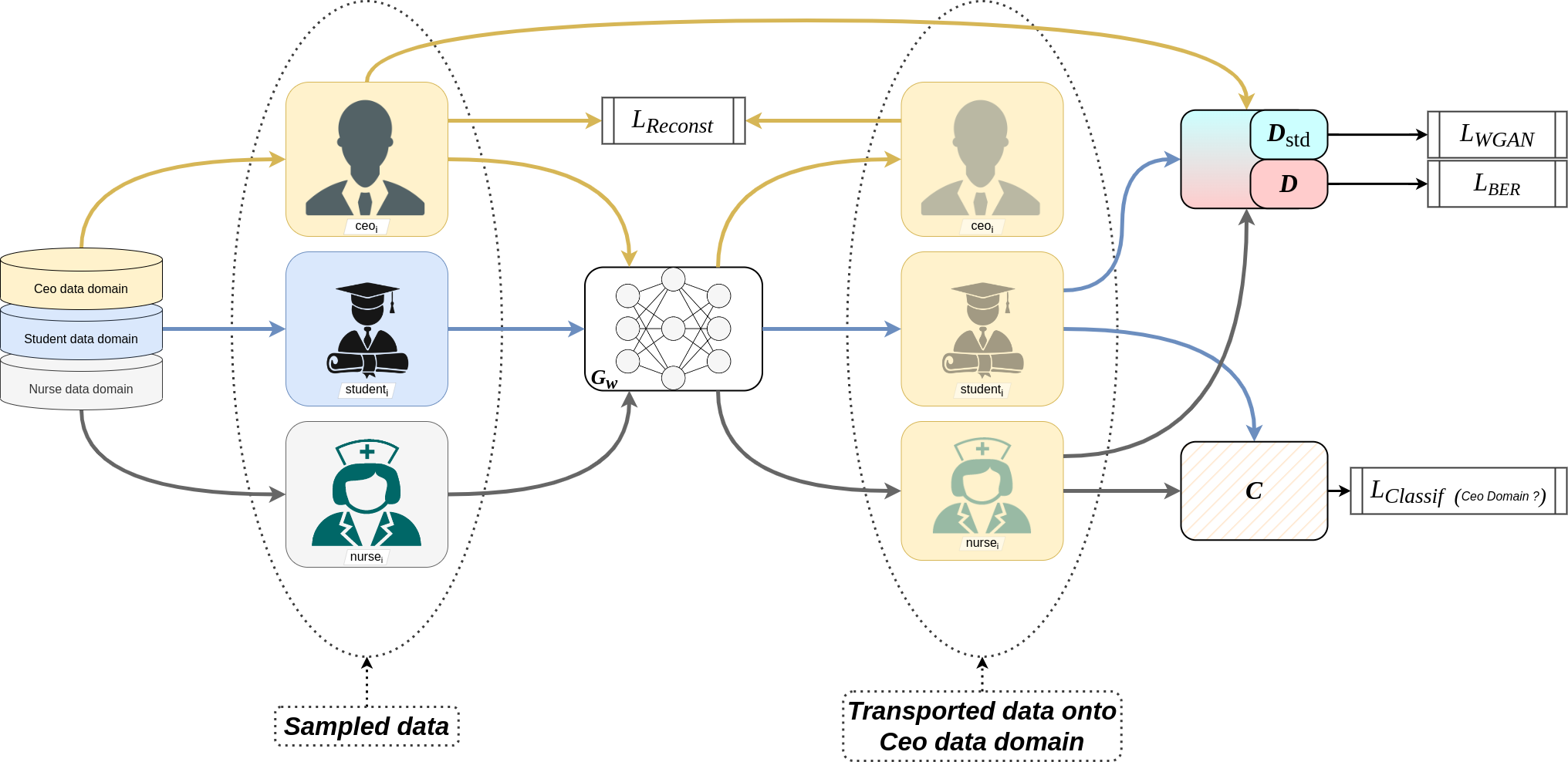}
    \caption{FairMapping (\fairmapping{}) approach. The objective is to map all the datapoints considered as protected (\emph{i.e.}, student data domain and nurse data domain) to the \textit{ceo} data domain (represented with the yellow background) such that they all datapoints share the same advantages (\emph{e.g.}, in case of a loan application) as the \textit{ceo}, such that the transformed student and transformed nurse are considered as part of the privileged group from the classifier \textit{C} point of view, and such that the discriminator \textit{D} is unable to distinguish the \textit{ceo}, from the \textit{nurse} and the \textit{student}, and vice-versa. 
    }
    \label{fig:fairmaping}
\end{figure}

Going further in the example described in the previous figures, consider a dataset composed of three sensitive classes, $s_{1} = ceo$, $s_{2} = nurse$ and $s_{3} = student$, with individuals from each domain applying for a loan.
To summarize, the objectives of \fairmapping{} in this scenario (Figure~\ref{fig:fairmaping}) is to train a model $G_{w}$ that can be used to transform datapoints from all classes onto the \textit{ceo} domain $R_{ceo}$, such that (1) $G_{w}$ performs as the identity function for any data from \textit{ceo} domain, (2) the individuals from the \textit{nurse} and the \textit{student} groups benefit from the same advantages and (3) such that all group transformations are undistinguishable from each other. 
The high level overview of our approach \fairmapping{} consists in the following steps : 
\begin{enumerate}
    \item For each datapoint $r_{i}^{t}$ in any domain $t \in \{ceo, student, nurse\}$, generating the transformed version $\bar{r}_{i}^{t} = G_{w}(r_{i}^{t})$.
    \item Achieving that $\bar{r}_{i}^{t} = r_{i}^{t}$ if $t$ equals $ceo$ (\emph{i.e.}, identity property).
    \item Ensuring, if $t$ is not equal to $ceo$, that $\bar{r}_{i}^{t}$ belongs to the $ceo$ domain using the classifier $C$ and the critic $D_{std}$, and that the transformation has been carried out with a low amount of modifications.
    \item Verifying, with the discriminator $D$, that all of the model $G_{w}$ outputs are indistinguishable from each other.
    \item Updating the models based on the previous observations.
\end{enumerate}

\subsection{Training procedure}
Hereafter, we describe the models used in our approach as well as their respective training procedure. 
More precisely, we first focus on the training of the classifier $C$, which is executed before any other models, followed by the training of the discriminator $D$, the critic $D_{std}$ and finally the generator $G_{w}$. 
These three last models are trained in an alternate manner, similarly to the classical training of GANs.

\subsubsection{Training of the classifier $C$}
The training procedure of \fairmapping{} start with the training of classifier $C$ to predict the sensitive attribute. 
For each datapoint $r_{i}^{t}$ with sensitive attribute value of $s_{t}$, $C$ outputs the probabilities $c_{i}^{j}$, which is the probability that $r_{i}^{t}$ belongs to the sensitive group $S_{j}$. 
Let this output vector be denoted by $ C(r_{i}^{t}) = \langle c_{i}^{1}, \ldots, c_{i}^{t}, \ldots, c_{i}^{k}  \rangle$ such that $ \: c_{i}^{j} = C(r_{i}^{t})^{j} \:$ and $  \: \sum_{j=1}^{k} c_{i}^{j} = 1 \: $.
$C$ is trained using the original data until convergence, with the objective of maximizing the accuracy of predicting the sensitive attribute. As such, $C$ is trained to minimize the loss of accuracy, which is defined as follows:
\begin{equation}
    L_{Classif} = L_{Acc} = 1 - \dfrac{\sum_{r_{i}^{j} \in R} C(r_{i}^{j})^{j}}{\| R\|}. 
    \label{eq:acc-loss}
\end{equation}
with $\| R \|$ defining the size of the training dataset.
Remark that since $C$ is trained on the original data, its training procedure of $C$ is independent of the transformation process and thus the other models. 
$C$ can thus be trained \emph{a priori} before its use in the transformation process.

\subsubsection{Training of the critic $D_{std}$ and the discriminator $D$}
The critic $D_{std}$ corresponds to the critic used for the training of \wgan{}.
This model takes as input the original data from the privileged distribution as well as the transformed protected data, and outputs a value corresponding to a measure of the \textit{distance} between them (Equation~\ref{eq:d-std}). 
Similarly to \wgan{}, the objective is to learn a critic model maximizing the computed \textit{distance} between the privileged group and the transformed protected groups $L_{D_{std}}$. 
Finding such critic ensures that the amount of modification introduced for the transformation of the protected data is minimal.
\begin{equation}
    \begin{split}
        L_{WGAN} = L_{D_{std}} &= D_{std}(R_{priv}) - D_{std}(G_{w}(R_{prot})) \\
                    &= \dfrac{\sum_{r_{i} \in R_{priv}} D_{std}(r_{i})}{\| R_{priv} \|} - \dfrac{\sum_{r_{j} \in R_{prot}} D_{std}(G_{w}(r_{j}))}{\| R_{prot} \|}.
    \end{split}
    \label{eq:d-std}
\end{equation}
where $\| R_{priv} \|$ and $\| R_{prot} \|$ being respectively the size of the privileged and of the protected groups.
Similarly to the classifier $C$, the discriminator $D$ is trained to predict the sensitive attribute, with the exception that $D$ infers the sensitive attribute based on the transformed protected data (obtained through the output of the model $G_{w}$) and the original privileged data. 
More precisely, $D$ is trained to maximize the identification of each transformed protected group as well as the original privileged group. 
The loss function for training $D$ is similar to that of $C$, which is the minimization of the accuracy loss $L_{Acc_{D}}$ (Equation~\ref{eq:d-acc-loss}).
$D$ can also be trained with the objective of minimizing the $L_{BER_{D}}$ (Equation~\ref{eq:d-min-ber-simp}), which corresponds to the minimization of the prediction error in each group.
We will refer to the discriminator loss as $L_{D}$ : $L_{D} = L_{Acc_{D}}$ or $L_{D} = L_{BER_{D}}$.
\begin{equation}
    L_{Acc_{D}} = 1 - \dfrac{\sum_{r_{i}^{1} \in R_{priv}} D(r_{i}^{1})^{1} + \sum_{r_{i}^{j} \in R_{prot}} D(G_{w}(r_{i}^{j}))^{j}}{\| R \|}.
    \label{eq:d-acc-loss}
\end{equation}.
\begin{equation}
    \begin{split}
        L_{BER_{D}} = &1 \\- &\dfrac{1}{k} \left(\dfrac{\sum_{r_{i}^{1} \in R_{priv}} D(r_{i}^{1})^{1}}{\| R_{priv} \| } + \sum_{j=2}^{k}\dfrac{\sum_{r_{i}^{j} \in R_{prot} \mid S=s_{j}} D(G_{w}(r_{i}^{j}))^{j}}{\| S=s_{j} \| }\right).
    \end{split}
    \label{eq:d-min-ber-simp}
\end{equation}

In case of multiple groups, we have also computed the mutual information ($MI$) between the prediction of the sensitive attribute $\bar{S}$ with $D$ and its real value $S$. 
The mutual information is introduced as a regularization of $D$, to further enhance the correct prediction of multiple groups. 
The computation of the mutual information is carried out similarly as in~\cite{romanelli2019generating}, with the objective of maximizing $MI$ during the training of $D$. 
An example of computation is presented in Appendix~\ref{mi-comp}. 
In this situation, the loss function of the discriminator can be decomposed as follows $L_{D} = L_{Acc} - \lambda_{D_{mi}} * MI$.

Note that the discriminator $D$ and the critic $D_{std}$ are composed of a set of parameters that are shared between both models, on top of which we added others parameters that are specific for each model task. This structure provides a beneficial relationship between both models. 
As a consequence, the training of both models $D$ and $D_{std}$ is carried out using the loss : $L_{D,D_{std}} =  L_{D} + \lambda_{D_{std_{gan}}} * L_{D_{std}}$, which correspond to the sum of the sensitive inference loss $L_{D}$ and the loss of the \wgan{} $L_{WGAN}$.

\subsubsection{Training of the transformation model $G_{w}$}
Hereafter, we detail the training process of the transformation model $G_{w}$ and how it achieves the identity, transformation, and protection properties.

\paragraph{Identity property.} To ensure and control the identity mapping of members of the privileged group $R_{priv}$, we rely on the minimization of the reconstruction loss $L_{Recons}$. 
This loss, instantiated with the $L_1$ norm (Equation~\ref{eq:recons}), constraints the model $G_{w}$ to learn the distribution of the privileged data.
\begin{equation}
    L_{Recons} = L_1(R_{priv}, G_{w}(R_{priv})) = \dfrac{\sum_{r_{i} \in R_{priv}} \mid r_{i} - G_{w}(r_{i}) \mid}{\| R_{priv} \|}.
    \label{eq:recons}
\end{equation}

We have observed in our early experiments that without the reconstruction constraint, the model $G_{w}$ is highly subject to the \textit{mode collapse} phenomenon, in which the generative model focuses its learning on a single or few modes of the data distribution. 
For instance, this would correspond to the situation in which the model outputs the same identical profile regardless of the inputted one. 
Thus, the reconstruction constraint enhances the diversity of the transformation procedure by learning the privileged distribution while improving the ability of $G_{w}$ to perform as the identity function for members of the privileged group.

\paragraph{Transformation property.} Recall that the privileged group is associated with the sensitive attribute value $s_{1}$.
The model $G_{w}$ has for main objective to transform a data point in such a way that it belongs to the privileged group. 
To realize this, the classifier $C$ is used to ensure this property by outputting the probability of a datapoint $r_{i}$ from the protected group to belong to the target group after its transformation, $C(G_{w}(r_{i}))^{1}$. 
Therefore, $G_{w}$ aims at maximizing the probability $C(G_{w}(r_{i}))^{1}$ with the output of the classifier $C$ being integrating within the training of $G_{w}$ through the loss $L_{C}$ to maximize:
\begin{equation}
    L_{C} = \dfrac{\sum_{r_{i} \in R_{prot}} C(G_{w}(r_{i}))^{1}}{\| R_{prot} \|}.
    \label{eq:pc}
\end{equation}
Note that maximizing Equation~\ref{eq:pc} correspond to minimizing $-L_{C}$.

The transformation is further improved through the use of the optimal transport, as explained with \wgan{} in Section~\ref{sec:opt-tranport}. 
More precisely, \wgan{} ensures that only the modifications necessary for the transformation are introduced during the transformation. 
As such, $G_{w}$ also minimizes the value function of \wgan{} (Equation~\ref{eq:em-kanton-dual}, with $f$ being $D_{std}$ and $p_{\theta}$ being $G_{w}$), which corresponds to the minimization of Equation~\ref{eq:d-std}.
As the first term of Equation~\ref{eq:d-std}, $D_{std}(R_{priv})$, does not involve $G_{w}$, its derivative with respect to $G_{w}$ parameters will be equal to zero. 
Thus, we can only minimize the term $ - D_{std}(G_{w}(R_{prot}))$ with respect to the parameters of the model $G_{w}$. 
As a consequence, the equation can be rewritten in the following way:
\begin{equation}
    L_{Gan} = - D_{std}(G_{w}(R_{prot})) = - \dfrac{\sum_{r_{j} \in R_{prot}} D_{std}(G_{w}(r_{j}))}{\| R_{prot} \|}.
    \label{eq:l-gan}
\end{equation}
 The objective of this minimization objective is to have the critic considering the privileged data $R_{priv}$ and the transformed protected $G_{w}(R_{prot})$ as equals.

\paragraph{Protection property.}
The (privacy) protection consists in the inability of inference of the sensitive attribute, from the privileged $R_{priv}$ and the transformed protected data $G_{w}(R_{prot})$. 
This property is introduced in the optimization of $G_{w}$ through the loss $L_{S}$, which can be instantiated with the $\ber{}$ (minimization of Equation~\ref{eq:g-max-ber})  or the accuracy (minimization of Equation~\ref{eq:g-min-acc}) to maximize the protection level. 
\begin{equation}
    \begin{split}
        L_{S} = &\dfrac{k-1}{k} -  \\
        &\dfrac{1}{k} \left(1 - \dfrac{\sum_{r_{i}^{1} \in R_{priv}} D(r_{i}^{1})^{1}}{\| R_{priv} \|} + \sum_{j=2}^{k}\left[1 - \dfrac{\sum_{r_{i}^{j} \in R_{prot} \mid S=s_{j}} D(G_{w}(r_{i}^{j}))^{j}}{\| S=s_{j} \|}\right]\right).
    \end{split}
     \label{eq:g-max-ber}
\end{equation}
More precisely, when $L_{S}$ is instantiated with the $Ber$ it becomes:
\begin{equation}
    \begin{split}
        L_{S} = &\dfrac{-1}{k} +  \\
        &\dfrac{1}{k} \left(\dfrac{\sum_{r_{i}^{1} \in R_{priv}} D(r_{i}^{1})^{1}}{\| R_{priv} \|} + \sum_{j=2}^{k}\dfrac{\sum_{r_{i}^{j} \in R_{prot} \mid S=s_{j}} D(G_{w}(r_{i}^{j}))^{j}}{\| S=s_{j} \|}\right).
    \end{split}
    \label{eq:g-max-ber-simp}
\end{equation}
Similarly, when $L_{S}$ is instantiated with the accuracy, it results in the following:
\begin{equation}
     L_{S} = \dfrac{\sum_{r_{i}^{j} \in \{R_{priv}, G_{w}(R_{prot})\}} D(r_{i}^{j})^{j}}{\| R_{priv} \|}.
     \label{eq:g-min-acc}
\end{equation}

Note that for a binary attribute, minimizing the accuracy does not protect the inference of the sensitive attribute, as having an accuracy of zero simply means that the model predictions are the reversed of the groundtruth (\emph{i.e.}, the $MI$ would still have the same value as the $MI$ obtained with the correct predictions). 
In this situation, the \ber{} is more suitable.

When dealing with multiple protected groups, the minimization of the mutual information is particularly useful as there might exist some situation in which the accuracy is minimal or the \ber{} are maximal for the protection, but the amount of information preserved in the transformed data point about the sensitive attribute is still significant. 
For instances, the discriminator could predict every member of the $ceo$ group as $student$, every $student$ as $nurse$ and every $nurse$ as $ceo$. 
In such case, the prediction accuracy is equal to zero but the mutual information is still high as the group labels are simply permuted. 
In this situation, it is necessary to minimize the mutual information between the predicted sensitive groups and the real groups through the following loss:
\begin{equation}
     L_{S} + \lambda_{G_{mi}} * MI.
\end{equation}
The mutual information also enables the use of the accuracy in the loss function in case of a single binary sensitive attribute, as it is impossible to achieve an accuracy of $0$ with a mutual information equal to $0$.

\paragraph{Global optimization of $G_{w}$.}
Now that each component of the optimization of $G_{w}$ have been defined, we can introduce the global loss function that the transforming model $G_{w}$ has to minimize:
\begin{equation}
    L_{G_{w}} = \lambda_{rec} L_{Recons} - \lambda_{c} L_{C} + \lambda_{gan} L_{Gan} +  \lambda_{d} L_{S}.
    \label{eq:gw}
\end{equation}

In Equation~\ref{eq:gw}, each term of the loss is weighted with a coefficient $\lambda$. 
These coefficients are used to assign the relative importance of each term and are used to better fine tune the approach, depending on several factors such as the dataset considered, the number of groups, etc. The higher the value of the coefficient, the higher the importance of the property associated.

As our model $G_{w}$ does not take the sensitive attribute as input, it can be used safely on any data point of the dataset to transport it towards the privileged distribution, with the insurance that the member of the privileged group will not be modified. 
As a consequence, a data practitioner only needs to identify some members of the privileged distribution to build the mapping function $G_{w}$, and once the model is trained he can transform the rest of the dataset into the privileged distribution. 
This procedure can be applied even in situation in which the access to the value of the sensitive attribute is difficult or impossible.


\subsubsection{Convergence analysis}
Let's assume that at each training iteration for $G_{w}$, we have access to the optimal discriminator $D^{*}$ and the optimal critic $D_{std}^{*}$. 

The optimal discriminator $D^{*}$ is able to differentiate every sensitive group (i.e, the privileged from the transformed protected groups, and the transformed protected groups from each other).
The optimal critic $D_{std}^{*}$ is the one that maximizes the loss $L_{D_{std}}$ (i.e, the difference between $D_{std}(R_{priv})$ and $D_{std}(R_{prot})$).
These assumptions are realistic in the sense that both models are neural networks that can approximate any functions~\cite{leshno1993multilayer}, and since they are convex functions, there exists an optimal (minimum) point. For each training step, the generator $G_{w}$ can be fixed while $D_{std}$ and $D$ are trained until they reach their respective optimal state. 

$G_{w}$ minimize the loss function presented in equation~\ref{eq:gw}. As we are dealing with the sum, the minimum of $L_{G_{w}}$ is achieved only when $L_{Recons}$, $-L_{C}$, $L_{Gan}$ and $L_{S}$ reach their respective minimum values. 

With the previous assumptions on $D^{*}$ and $D_{std}^{*}$:

\paragraph{Identity} $L_{recons}$ is minimum if and only if $G_{w}(r_{i}) = r_{i} \: ; \: \forall \: r_{i} \in R_{priv}$. $G_{w}$ is thus forced to learn the correct representation of the privileged data distribution, ensuring that the approach can be safely applied to the dataset without the need to specify members of the privileged group. As $L_{recons}$ is a convex function, the minimum is reachable.

\paragraph{Transformation} As $C$ outputs probabilities (between $0$ and $1$), the minimum possible value of $-L_{C}$ is $-1$ ($L_{C}$ is maximum at $1$). This minimum value is reached only if $C(G_{w}(r_{i}))^{1} = 1 \: \forall \: r_{i} \in R_{prot}$, meaning that the probability of belonging to the target group (as predicted with $C$) for any member of the transformed protected data is $1$. In other words, all members of the transformed protected group belong to the same decision frontier as the privileged data, for a classifier trained to distinguish the original privileged data $R_{priv}$ from the original protected groups data $R_{prot}$. This property ensures that the transformed dataset is usable for existing trained models. The minimum value for this term is achievable, as $L_{C}$ is convex. 
Nonetheless, achieving the optimal (minimum) value for this property does not ensure that the least amount of modifications is introduced with the transformation.

\paragraph{Optimal transport} As $D_{std}^{*}$ is the optimal solution that maximizes $L_{D_{std}}$ for any possible function $G_{w}$ (equation~\ref{eq:d-std}), $G_{w}$ can only minimize $L_{D_{std}}$ by reaching the state $R_{priv} = G_{w}(R_{prot})$. At this state, $G_{w}$ would correspond to the transport plan of $R_{prot}$ to $R_priv$ that requires the least amount of \textit{efforts} or modifications in our case, as per the Kantorovich-Rubinstein Duality shown with \wgan{} (section~\ref{sec:opt-tranport}). 

\paragraph{Protection} Finally, let's assume that $L_{S}$ is instantiated with equation~\ref{eq:g-max-ber-simp}. For ease of writing, let's pose $\dfrac{\sum_{r_{i}^{1} \in R_{priv}} D(r_{i}^{1})^{1}}{ \| R_{priv} \|}$ as $D(R_{priv})^{1}$ and $\dfrac{\sum_{r_{i}^{j} \in R_{prot} \mid S=s_{j}} D(G_{w}(r_{i}^{j}))^{j}}{\| S=s_{j} \|}$ as $D(G_w(R_{prot}\mid S=s_{j}))^{j}$. $L_S$ is therefore written as :
\begin{equation*}
    L_{S} = \dfrac{-1}{k} + \dfrac{1}{k} \left(D(R_{priv})^{1} + \sum_{j=2}^{k}D(G_w(R_{prot}\mid S=s_{j}))^{j}\right)
\end{equation*}
$L_{S}$ is minimal if the term $D(R_{priv})^{1} + \sum_{j=2}^{k}D(G_w(R_{prot}\mid S=s_{j}))^{j}$ is equal to $1$ (recall that the \ber{} is always positive, and the optimal value is achieved with the value $\dfrac{k-1}{k}$, thus having $L_{S} = 0$) this means either :
\begin{enumerate}
    \item $D(R_{priv})^{1} = 1$ and $ \sum_{j=2}^{k}D(G_w(R_{prot}\mid S=s_{j}))^{j} = 0$
    \item $D(R_{priv})^{1} = 0$ and $\sum_{j=2, j \neq t}^{k}D(G_w(R_{prot}\mid S=s_{j}))^{j} = 0$ and $D(G_w(R_{prot}\mid S=s_{t}))^{t} = 1$
    \item or $D$ makes partial errors in each group such that the total sum of correct prediction equals $1$ : $D(R_{priv})^{1} = \zeta$ and $\sum_{j=2}^{k}D(G_w(R_{prot}\mid S=s_{j}))^{j} = 1-\zeta$.
\end{enumerate}
As we assumed $D=D^{*}$, the discriminator cannot make partial mistakes in predictions, thus $D(R_{priv})^{1}=D^{*}(R_{priv})^{1}$ would always be equal to $1$ ($D^{*}$ is always able to distinguish members of the privileged group from the rest). Cases (2) and (3) are therefore not possible. $G_{w}$ would therefore have to modify the data such that $\sum_{j=2}^{k}D(G_w(R_{prot}\mid S=s_{j}))^{j} = 0$. 

This solution can be achieved with $G_{w}$ changing the data such that either (a) $D(G_w(R_{prot}\mid S=s_{j}))^{j+1} = 1$ (meaning that $D$ predict all $student$ as $nurses$ and all $nurse$ as $student$), (b) $D$ behave as a random classifier on the modified data, or (c) $G_{w}$ modifies $R_{prot}$ such that $D$ predict all modified protected data as belonging to the privileged group. 

In the first case, the mutual information is still maximal, as the joint probability between the original $S$ and its predicted value still exists (by knowing the original value (e.g: $student$) we can determine the predicted value of the modified data (i.e: $nurse$)) even though the prediction error is maximal. The minimization of the mutual information introduced helps remove the joint distribution and ensures that only the only reachable solutions are (b) or (c) with the help of $D_{std}$. The same demonstration holds for equation~\ref{eq:g-min-acc}. As both equations are convex functions ($L_{S}$ is convex), a minimum exists for the protection equation. 

With this analysis, we can observe that the combination of our different objectives offers a plausible solution that is achievable. In other words, a minimum exists for $L_{G_{w}}$ and is achievable, and the solution that minimizes $L_{G_{w}}$ will satisfy our \textit{identity}, \textit{transformation} and \textit{protection} properties (our properties does not hinder each other, but rather improve each other). This, assuming that we have access to the optimal $D^{*}$ and $D_{std}^{*}$ at each training step of $G_{w}$.

\section{Extension of state-of-the-art approaches}
\label{sec:soaupdate}
The structure of our approach \fairmapping{} closely ressemble some of the known state of the art approaches. This resemblance is mainly handled through the hyper-parameters $\lambda_{rec}$, $\lambda_{c}$, $\lambda_{gan}$, $\lambda_{d}$ that control our different objectives. 

For instance, we will show how \wgan{}, \attgan{} and \gansan{} can be obtained with our approach.
\begin{itemize}
    \item \wgan{} is easily obtained by setting $\lambda_{rec}$, $\lambda_{c}$ and $\lambda_{d}$ to $0$. This leave only the \textit{gan} transformation on which our approach is based on. Thus, $L_{G_{w}} = \lambda_{gan}L_{Gan}$. 

    \item \attgan{}, as described in \ref{sec:attgan}, share similar characteristics to \fairmapping{}. We can modify our approach to obtain \attgan{} by setting the reconstruction to all datapoints of the dataset, instead of only the target group, and maximising the prediction of $S$ with $D$ instead of minimising it as done in our setting. The obtained loss function would be similar to $L_{G_{w}} = \lambda_{rec} \boldsymbol{L_{Recons_{all}}} - \lambda_{c} L_{C} + \lambda_{gan} L_{Gan} \boldsymbol{-} \lambda_{d} L_{S}$

    \item \gansan{} is obtained by also setting the reconstruction constraint to the whole dataset instead of only the privileged group and ignoring the classification and the gan transformation constraint by setting the coefficient $\lambda_{c}$ and $\lambda_{gan}$ to zero. The original \gansan{} approach requires that the discriminator $D$ predict $S$ from the reconstructed data from the privileged group and the modified protected data, instead of predicting from the original privileged and modified protected data as in \fairmapping{}. However, we can approximate \gansan{} using the \fairmapping{} prediction methodology (training the discriminator with original data from the privileged group) and use the reconstructed data as well as the transformed protected data at test time.   
\end{itemize}

As our approach objective is to preserve the data semantic meaning by imposing a constraint on the distribution on which the data will be mapped onto, we proposed some modifications of existing framework in order to achieve this semantic meaning preservation. We proposed $\gansan{}-OM$ as the $\gansan{}$ framework with the final distribution fixed.
$\gansan{}-OM$ is obtained from \gansan{} by only changing the discriminator and the reconstruction loss inputs. Thus, during the training procedure, the discriminator predict the sensitive attribute using the original privileged data and the modified protected group (as in equation~\ref{eq:d-acc-loss} or ~\ref{eq:d-min-ber-simp}), and the auto-encoder is trained only to reconstruct data from the privileged group: $L_{recons} = L1(R_{prot}, G_{w}(R_{prot}))$. Since we are maximising the protection against the original data from the privileged group, we expect the modified protected data to be as close as possible to the privileged group, thus maximising the transformation property.
The same protocol can be applied to the approach \dirm{}, leading to its extended version dubbed as $\dirm{}-OM$.

In \gansan{}~\cite{aivodji2021local}, the authors investigated the protection achieved if only a part of the dataset is transformed through their approach. 
In other words, they analysed the sensitive attribute protection in cases in which only a part of the dataset is transformed to protect the sensitive attribute, which would corresponds to the situation in which some users decide to transform their data while others do not. 
They have observed empirically that in such cases, the sensitive attribute is protected as long as at least half of the data of the dataset (including both the sensitive and the privileged) is transformed. 
If the data points belonging to the protected or the privileged group are the only one transformed, the sensitive attribute cannot be protected.

Our approach \gansan{}-OM differs from their analysis by the fact that our protocol optimise the protection with respect to the original data of the privileged group, while in their, the protection is optimised for the intermediate distribution on which both the protected and the privileged group are mapped onto.


\section{Experiments}
\label{sec:experiments}
In this section, we describe the experimental setup use to evaluate our approach.
First, we will present the datasets used, followed by the state-of-the-art approaches to which we compare our approach. Afterwards, we will discuss the evaluation metrics, and we review the set of external classifiers used throughout our evaluation.
Then, we will discuss methodology used for comparison and the transformation processes that can be carried out with Fair Mapping.
Finally, we will review the different use cases in which our approach can be deployed.

We evaluate the approach on three datasets from the literature: \textit{Adult Census Income}, \textit{German Credit} and \textit{Lipton}.

\textit{Adult Census Income~\footnote{\url{https://archive.ics.uci.edu/ml/datasets/adult}}} is a dataset extracted from the US census database of \textit{1994}. It is composed of $45222$ individuals and $14$ attributes which describe the social and economic status of each individual (e.g: \textit{Occupation}, \textit{Income}, \textit{Native Country}, etc.). The task in the dataset is the prediction of the income level of individuals, whether the given individual would have an income greater than $50K$. The sensitive attribute is the binary attribute \textit{sex}, containing either \textit{Male} or \textit{Female}. For the multi-sensitive attribute case, we will also use the attribute Race with values $Whites$ and $Non-White$ in addition to the \textit{sex}. Thus, we will obtain a single attribute named as $group$ which will contain the combination of each binary attribute ($White-Male$, $Non-White-Female$, etc.).

\textit{German Credit~\footnote{\url{https://archive.ics.uci.edu/ml/datasets/statlog+(german+credit+data)}}} is available on the UCI repository.
It contains the profiles of $1000$ applicants to a credit loan, each profile being described by $21$ attributes.
Previous work~\cite{Kamiran2009} have found that using the \textit{age} as the sensitive attribute by splitting its values at the threshold of $25$ years (differentiating between old and young) yields the maximum discrimination based on Disparate Impact.
In this dataset, the decision attribute is the quality of the customer with respect to their credit score (\emph{i.e.}, good or bad).

\textit{Lipton} is a synthetic dataset created by Lipton, Chouldechova and Auley~\cite{lipton2018does} to investigate the impact of fairness in-processing algorithms.
This dataset consists of three attributes: hair length and work experience and the decision attribute indicating if a given person should be hired.
Both the hair length and work experience are correlated with the gender (the sensitive attribute), while the decision is only based on the work experience.
This dataset is relevant to our study as it was used in~\cite{black2020fliptest} to investigate the bias of this dataset by using an approximation of the optimal transport based on GANs.

Table~\ref{tab:datasets-composition} summarizes the composition of each dataset, as well as the proportions of different groups.

\begin{table}[h!]
    \caption{Information on the datasets used in our experiments.}
    \resizebox{\columnwidth}{!}{%
        \begin{tabular}{cccccccccc}
            \toprule
            Dataset & size  & sensitive features                      & Nb. sens. groups & Privileged Group Prop.  & Maj. Group Prop. & $Pr(Y = 1)$ & $Pr(Y = 1 \mid S=s_1)$ & Optimal $\ber{}$ & Optimal $\sac{}$ \\
            \midrule
            Adult  & 45222  & sex                                  & 2                & 0.63                 & 0.63                 & 0.2478             &  0.3124  &  $\dfrac{1}{2}$ & 0.63                     \\
            Adult2  & 45222  & sex, race                                  & 4                & 0.597                 & 0.597                 & 0.2478             & 0.3239  &  $\dfrac{1}{2}$ & 0.597                         \\
            German  & 1000  & Age in years                            & 2                & 0.81                & 0.81                & 0.7                & 0.59  &  $\dfrac{1}{2}$ & 0.81                         \\
            Lipton  & 2000  & gender                                  & 2                & 0.5                 & 0.5                 & 0.3425             & 0.27  &  $\dfrac{1}{2}$ & 0.5                         \\
            
            \bottomrule
        \end{tabular}
    }
    \label{tab:datasets-composition}
\end{table}

The performances of \fairmapping{} are evaluated in two steps : the comparison step and the fairness step. In these experiments, we considered that the decision attribute is not modified with our approach (we used the original dataset decision). We discuss the rationale behind the use of a modified decision in section~\ref{sec:decision-trans}.

\subsection{State-of-the-art comparison}
In the comparison step, we evaluate the performances obtained with our approach and compare them against those obtained with several approaches of the state of the art, namely \wgan{}~\cite{arjovsky2017wasserstein}, \attgan{}~\cite{he2019attgan}, \gansan{}~\cite{aivodji2021local} and \dirm{}~\cite{feldman2015certifying}.
We re-implemented these approaches (using the same procedure as in the original paper) in order to apply them in our context, except the \dirm{} which we took from the AIF360 framework~\cite{bellamy2018ai}. These approaches have been described in section~\ref{sec:related_work}. In addition to these approaches, we also compare the performances obtained with $\gansan{}-OM$.

For each of these approaches, we compute the metrics $fidelity$, $classification$, \textit{Balanced Error Rate} and \textit{Accuracy}. Each of these metrics echoed the different objective of \fairmapping{} described in section~\ref{sec:fairmapping}. 

\begin{itemize}
    \item $fidelity$ ($\fid{}$) represents the closeness of the modified data to their original counterpart. It is obtained through equation~\ref{eq:fidelity}, and the perfect fidelity (data are identical) has the value of $1$.
    \begin{equation}
        \fid{} = \dfrac{1}{\| X \|} (X - G_{w}(X))^{2} = \dfrac{\sum_{r_{i} \in X} (r_{i} - G_{w}(r_{i}))^{2}}{\| X \|}
        \label{eq:fidelity}
    \end{equation}
    Throughout our analysis, we computed the \fid{} in three fashion : at the whole dataset level ($X=R$, $\fid{} = \fid{}_{all}$), only at the privileged group data ($X=R_{priv}$, $\fid{}_{priv}$), or only with the protected data ($X=R_{prot}$, $\fid{}_{prot}$).

    \item the $classification$ or $\classification{}$ measures the proportion of transformed data of a given group that belongs to another group of the dataset, from the point of view of the original dataset. The proportion is given by a classifier trained on the original version of the dataset. In our experiments, we measure the proportions of individuals in the transformed dataset that have been predicted by external classifiers (described in the following paragraphs) to be in the privileged group. The $classification$ is driven by the equation~\ref{eq:classification}
    \begin{equation}
        \classification{} = P(f(G_{w}(X)) = s_{1}) = \dfrac{\sum_{r_i \in X} f(G_{w}(r_{i})) == s_{1}}{\| X \|}
        \label{eq:classification}
    \end{equation}
    Just as with $\fid{}$, we consider that the classification is computed either for the whole dataset ($X=R$, $\classification{}=\classification{}_{all}$) or only for the protected group level ($X=R_{prot}$, $\classification{}=\classification{}_{prot}$).

    \item the \textit{Balanced Error Rate} ($\ber{}$) and \textit{Accuracy} ($\sac{}$) are external classifier performance measures that we apply to the prediction of the sensitive attribute. As \fairmapping{} both protects the sensitive attribute using the original privileged data (recall that the discriminator $D$ is trained using the transformed protected and the original privileged data) and reconstructs the data from the privileged group, we can measure the protection of the sensitive attribute with respect to either the original privileged data or its reconstructed version. In the former case, we will append the subscript $og_{prv}$ ($\ber{}_{og_{prv}}$), and the latter case will be identified with $rc_{prv}$ ($\ber{}_{rc_{prv}}$).
\end{itemize}

\subsubsection{Optimization and validation}
The optimization of our hyperparameters was carried out using the ray-tune~\cite{liaw2018tune} optimization framework, with optuna~\cite{akiba2019optuna} as the underlying algorithm. Each dataset is divided into a training and a validation set, and for each approach, we tested $100$ combinations of different hyperparameters. 
Our experiments were carried out on a maximum of $6$ CPUs and $2$ GPU with $4$ Go of memory. To simplify our hyperparameters search, we first find the model structure that maximizes the fidelity of the reconstructed data. Such structure will serve as the structure of all models involved in our experimentation, and will not change throughout our experimentations. The model tuning will thus correspond to the search of the appropriate values of the $\lambda_{i}$ parameters.

We train each approach to optimize their respective metrics :
\begin{itemize}
    \item \fairmapping{} maximizes $\fid{}_{priv}$ for the identity, $\classification{}_{prot}$ for the transformation and $\ber{}_{og_{prv}}$ for the protection (we also minimize $\sac{}_{og_{prv}}$ instead of $\ber{}_{og_{prv}}$).
    
    \item \wgan{} only transforms the data onto the privileged group. Thus, we the approach is trained only to maximize the transformation metric measured over all the dataset rows $\classification{}_{all}$.
    
    \item \gansan{} protects the sensitive attribute by finding the minimum amount of perturbation to introduce in all the datapoints of the dataset. Thus, \gansan{} maximize the fidelity over all the dataset $\fid{}_{all}$ and the protection with the reconstructed privileged group and modified protected data $\ber{}_{rc_{prv}}$. For the multivalued sensitive attribute, we exploit the $\ber{}$ to extend the $\gansan{}$ approach.
    
    \item \attgan{} as described in section~\ref{sec:related_work} mapped the privileged data to the protected group distribution, and the protected data onto the privileged distribution, while ensuring that the data mapped onto their original distribution are reconstructed. Thus, it maximizes $\fid{}_{all}$, and also maximizes the classification of transformed datapoints to different groups based on a random permutation of the sensitive attribute $\classification{}_{shuf}$. We will only report the classification metric computed with respect to the target group $\classification{}_{all}$, instead of the random permutation.
    
    \item \dirm{} does not particularly maximize any metric. The protection of the sensitive attribute is controlled with a single parameter $\lambda$. As such, we decided to maximize the metric $\fid{}_{all}$ and the protection based on the reconstructed data $\ber{}_{rc_{prv}}$, since the approach modify all groups in the dataset. The approach is specifically designed to handle a single binary sensitive attribute. Thus, it can not be extended to the multivalued case.

    \item \gansan{}-OM correspond to \gansan{} applied in the context of \fairmapping{}, thus, it maximizes the reconstruction of the protected group $\fid{}_{prot}$ and the protection with the privileged group $\ber{}_{rc_{prv}}$.

    \item \dirm{}-OM is similar to \gansan{}-OM.
\end{itemize}

To achieve a fair comparison between all approaches, we rely on a set of external classifiers to compute the metrics to optimize, such as $classification$ and $protection$. Indeed, most approaches we compare ourselves to have their own discriminator model adapted to their specific objective. Moreover, some approaches, such as the $\dirm{}$, do not include a discriminator to measure the protection of the sensitive attribute. We thus rely on the external classifiers to have a similar basis of comparison for all.

Once, each approach has optimized its respective set of metrics, we compute the Pareto front that exhibits the different trade-offs we can obtain between the different metrics (e.g., the cost on $\fid{}_{priv}$ to provide better protection $\ber{}_{og_{prv}}$ and $\classification{}_{prot}$ for \fairmapping{}, the cost on $\fid{}_{all}$ to achieve a higher $\ber{}_{rc_{prv}}$ for $\gansan{}$). As each approach has a different set of metrics, we provide a fair comparison by building a second Pareto front on top of the first one. The first Pareto front is specific to the approach set of metrics, while the second Pareto front displays the trade-offs obtained in the context of another approach.
The underlying idea is that the best of all approaches should outperform others in every perspective, which translates into having the highest Pareto front (for all metrics we maximize). 

For instance, to compare \fairmapping{} and \gansan{}, we optimize each approach with their respective set of metrics described above and build their respective Pareto front (i.e, $\fid{}_{priv}$, $\ber{}_{og_{prv}}$ and $\classification{}_{prot}$ for \fairmapping{}; $\fid{}_{all}$, $\ber{}_{rc_{prv}}$ for \gansan{}). This represents a situation where each approach is used regardless of the other. Afterwards, each \gansan{} model on its Pareto front is evaluated with the set of metrics of \fairmapping{}, thus enabling the building of a second Pareto front for \gansan{} representing its application in the \fairmapping{} perspective: for each trade-off on the \gansan{} Pareto-front, we compute the metrics of \fairmapping{} ($\fid{}_{priv}$, $\ber{}_{og_{prv}}$ and $\classification{}_{prot}$), and we filter the results in order to build a new Pareto front based on the set of metrics of \fairmapping{}.
Similarly, we evaluate \fairmapping{} with the set of metrics of \gansan{} and building another Pareto front for \fairmapping{}. 

In the main part of our article, we will only present results with the \fairmapping{} perspective (which also correspond to the \gansan{}-OM and \dirm{}-OM perspective). 

All of these approaches are compared to the baseline, which correspond to the original data without any modifications.

\paragraph{External classifiers}
The computation of some metrics rely on the prediction made by some classifiers. We compute those metrics with a set of classifiers dubbed as external classifiers, as they are independent to the frameworks of protection of the sensitive attribute. 
For the state-of-the-art comparison, the set of external classifiers used is composed of \textit{Gradient Boosting Classifier} (GBC) and Support Vector Machine (SVM). For each metric, we only report the best result obtained among all external classifiers, which corresponds to the worst value for the $classification$ among all classifiers (the worst possible case); the lowest protection achievable (lowest $\ber{}$ and highest $\sac{}$).

All classifiers were trained using the \textit{scikit-learn} Library\footnote{https://scikit-learn.org/stable/}.
In Table~\ref{tab:exthprms}, we report the set of hyperparameters used.

Despite the fact that we have tried a diverse set of classifiers, we acknowledged that there might still exist a classifier (or a set of hyperparameters that would lead to a classifier) with a higher predictive power for inferring the sensitive attribute from our transformation.

\subsubsection{State of the art : Results}
In this section, we will only present the results obtained when using transported data from the protected group and the reconstructed data from the privileged one. This usage correspond to one where the model $G_{w}$ is used without the knowledge of the sensitive attribute, thus, we cannot identify the privileged group to decide whether we should apply the transformation process on the data. The identity property should ensure that the privileged group data are unmodified. However, as the models are not perfect, $G_{w}$ would only be able to approximate the identity function, outputting data from the privileged distribution as close as possible to the original distribution. 
In section~\ref{sec:results-orig-priv}, we present the results obtained when the privileged data is known and kept unmodified while the protected group data are transformed. We propose an explanation to the significant difference observed between the use of the original privileged data and their reconstructed version to protect the sensitive attribute in section~\ref{sec:rec-vs-protection}
We refer the reader to the table~\ref{tab:datasets-composition} for the optimal protection metrics values. 

\paragraph{Lipton results}
\begin{figure}[ht]
        \includegraphics[width=1\linewidth, height=0.45\textheight]{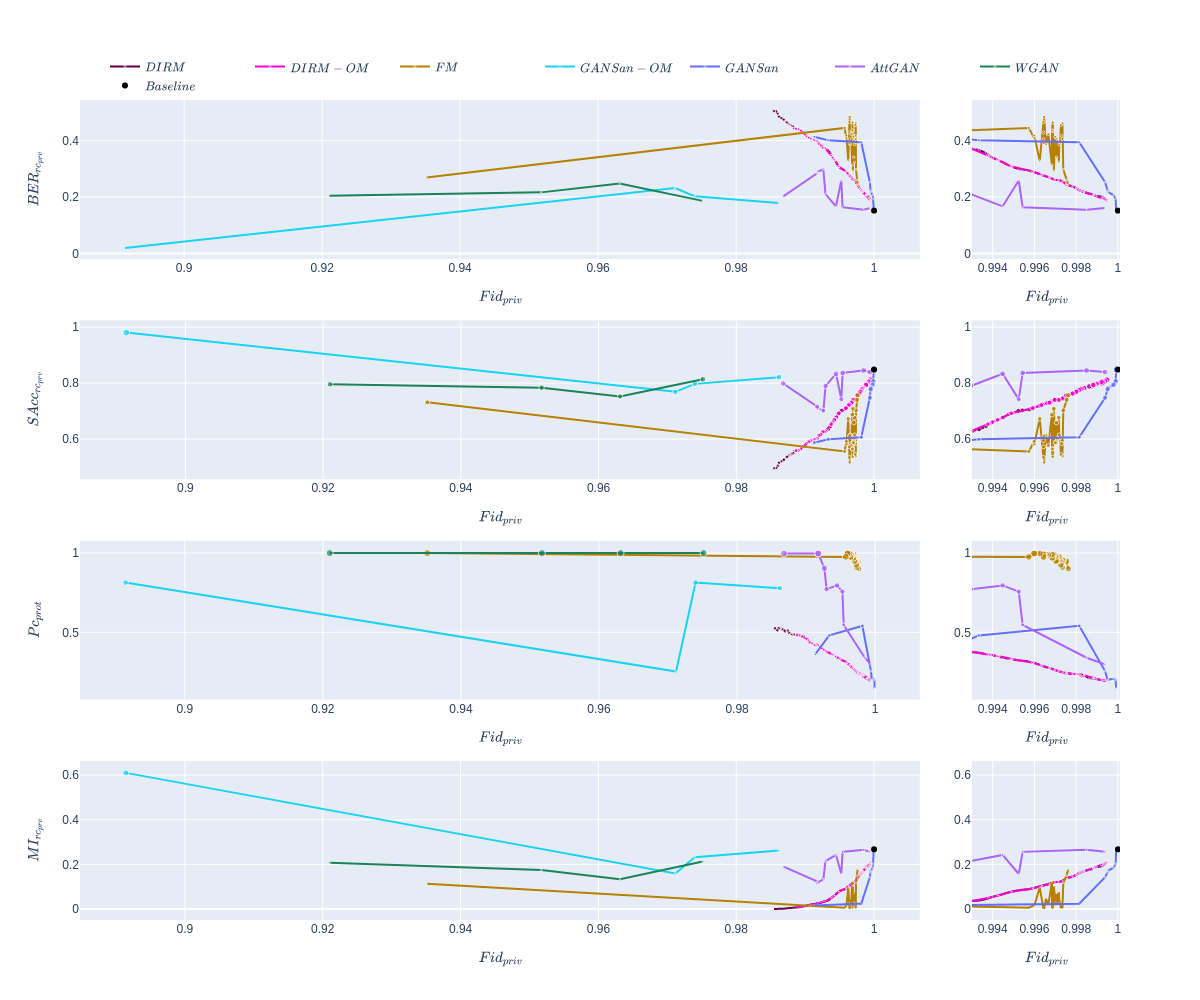}
    \caption{Pareto fronts in the \textit{Fair Mapping} perspective for approaches investigated on Lipton. The right column present all points on the fronts, while the right columns present the same results but on the range [$0.985$ - $1$], for a better visualization. Top to bottom: protection $\ber{}_{rc_{prv}}$, protection $\sac_{rc_{prv}}$, accuracy of $S$, $\classification_{prot}$, mutual information $MI_{rc_{prv}}$.
    }
    \label{fig:results-front-lipton}
\end{figure}

In Figure~\ref{fig:results-front-lipton} we present the results obtained on the \textit{Lipton} dataset. The dataset is balanced, and the optimal protection is achieved with a \ber of $0.5$, and with the accuracy of $0.5$. As we can observe on the figure, the worst trade-offs between the protection $\ber{}_{rc_{prv}}$ and the fidelity $\fid{}_{priv}$ is achieved with \wgan{}, followed by $\gansan{}-OM$. Nonetheless, $\wgan{}$ achieve the perfect transformation $\classification{}_{prot}$ in contrast to $\gansan-OM$, which, as we could have expect, has a $\classification{}_{prot}$ of at most $81\%$. While the lower protection results of $\gansan{}-OM$ are somewhat unexpected, its performances on the classification can be explained by the reconstruction of the protected group, thus limiting the transformation towards the privileged group. We can ground this explanation by observing the results of $\gansan{}$ and $\dirm$, where there is no transformation towards the privileged group: $\classification{}_{prot}$ is close to a random guess.

\attgan{} successfully transforms the data towards the privileged group, at a slight cost on the fidelity. However, its protection is not among the best achievable. 
$\dirm{}$ and $\dirm{}-OM$ achieve the highest protection among all approaches, but the lowest transformation. $\dirm{}-OM$ behave similarly to $\dirm{}$, even though they optimize different objectives. This can be explained by the fact that they rely on the same underlying procedure of building the median cumulative distribution function (and mapping the data towards such median), on top of which $\dirm{}-OM$ replace the modified privileged data by its original counterparts.
$\gansan{}$ dominates all approaches for the highest level of fidelity ($\gansan{}$ provides the highest $\ber{}_{rc_{prv}}$ value for fidelity values $\fid{}_{priv}$ above $0.998$), but fails to transform the data toward the privileged distribution, leaving the data onto an uncontrolled median distribution. 

Our approach $\fairmapping{}$ provide the best trade-off on all metrics. The transformation $\classification_{prot}$ is close to $1$, while the protection is well above other approaches when considering the fidelity range of $[0.995-0.998]$. We can also observe that the mutual information is close to zero for most of the results on the Pareto front. 
We 

\paragraph{German Credit results}
\begin{figure}[ht]
        \includegraphics[width=1\linewidth, height=0.45\textheight]{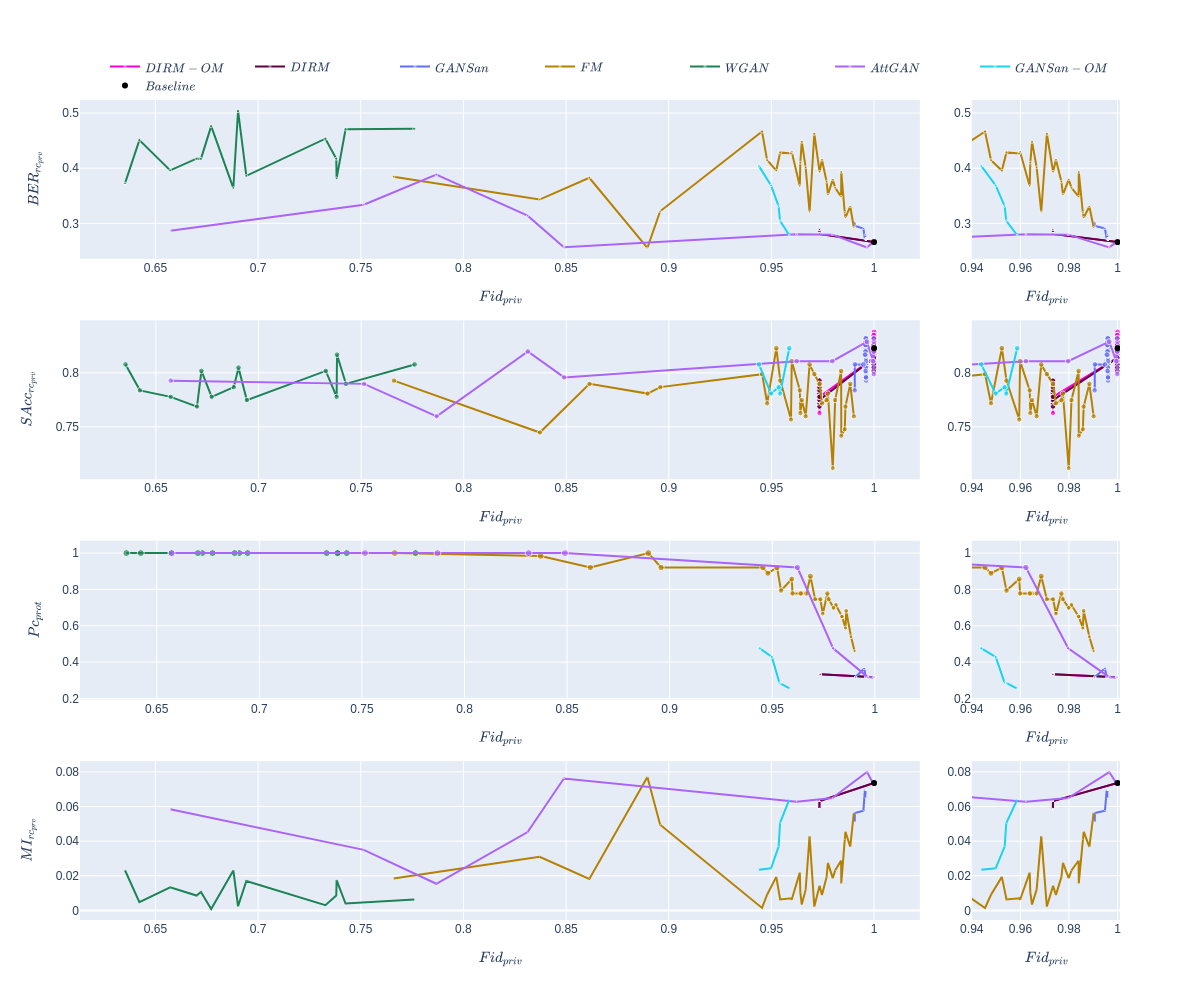}
    \caption{Pareto fronts in the \textit{Fair Mapping} perspective for approaches investigated on German Credit. The right column present all points on the fronts, while the right columns present the same results but on the range [$0.94$ - $1$], for a better visualization. Top to bottom: protection $\ber{}_{rc_{prv}}$, protection $\sac_{rc_{prv}}$, accuracy of $S$, $\classification_{prot}$, mutual information $MI_{rc_{prv}}$.
    }
    \label{fig:results-front-german}
\end{figure}

On German credit (Figure~\ref{fig:results-front-german}), the protected group only represents $19\%$ of the dataset. As such, the $\ber$ of predicting the sensitive attribute is already close to $0.3$ on the original data (the baseline), which makes the protection of the sensitive attribute harder. As a matter of fact, $\dirm{}$, and $\gansan{}$ which have some of the best performances on Lipton are not performing well above the baseline of $\ber{}_{rc_{prv}}=0.3$. $\attgan{}$ does not produce better results either. 
$\wgan{}$ performs better than most approaches for the protection, but at a significant cost on the fidelity $\fid{}_{priv}$ (the highest fidelity achievable with the approach is less than $0.8$).
As we could expect from $\wgan{}$ and $\attgan{}$ on the transformation properties, these approaches almost perfectly transform the data of the protected group such that they are predicted as part of the privileged distribution.
$\fairmapping{}$ outperforms all approaches on nearly all metrics, and for metrics where the approach is not the best, its results are close to the best achievable results. We can explain the results of $\fairmapping{}$ with two factors: first the mutual information and secondly the mapping towards the privileged distribution. 
On one hand, minimizing the mutual information forces the approach to better protect the sensitive attribute, by further removing correlation due to the group size. On the other hands, mapping the protected group towards the privileged one offer the advantage of reducing the complexity of finding the ideal intermediate distribution that would protect the sensitive information, in addition to not modifying the largely present group in the dataset (since the privileged group is largely present in the dataset, the models only have to find the suitable modification of the protected groups).
\gansan{}-OM enhances the protection but still fails to transform the data. \dirm{}-OM behave identically to \dirm{}.

\paragraph{Adult Census results}
\begin{figure}[ht]
        \includegraphics[width=1\linewidth, height=0.45\textheight]{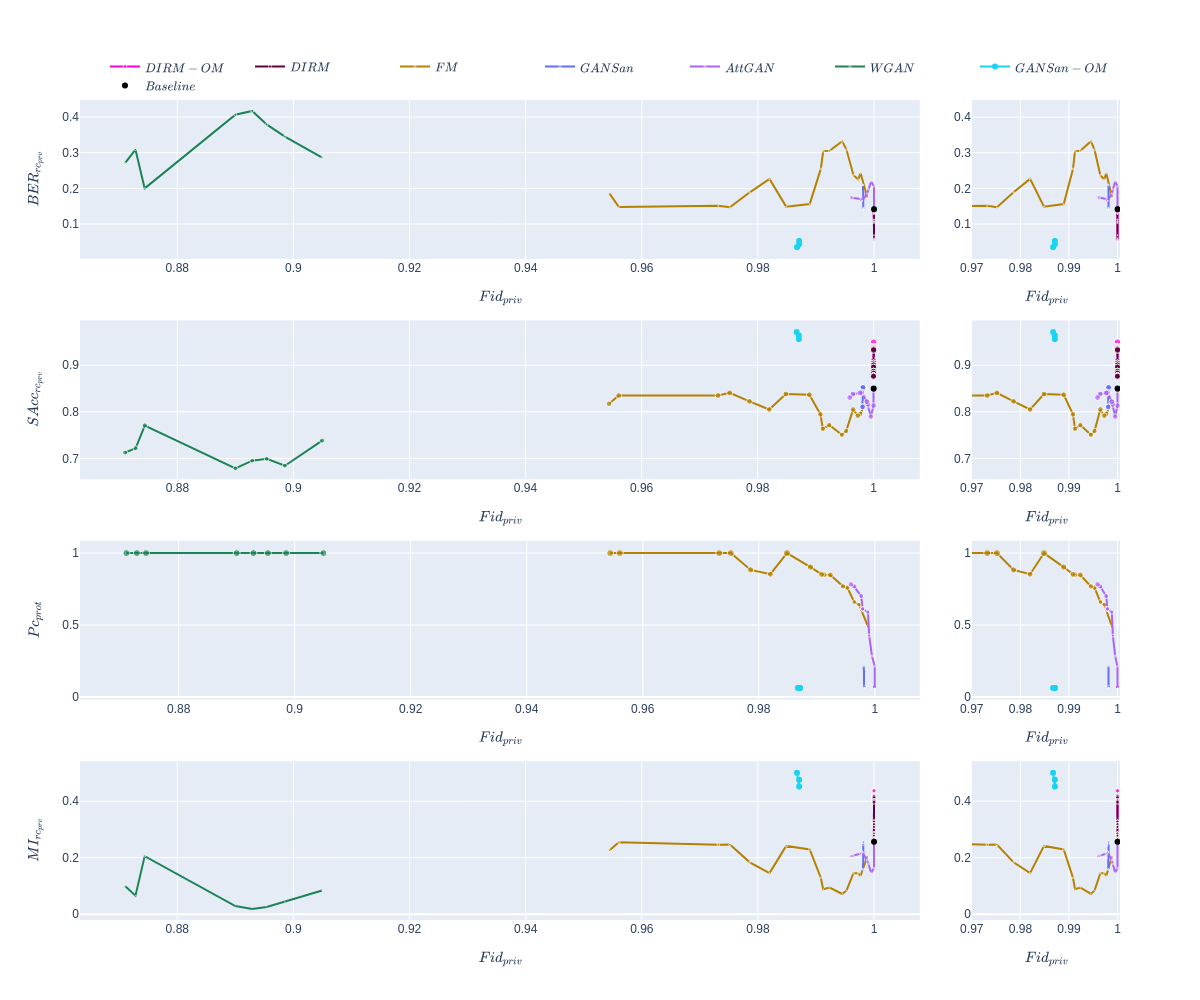}
    \caption{Pareto fronts in the \textit{Fair Mapping} perspective for approaches investigated on Adult Census. The right column present all points on the fronts, while the right columns present the same results but on the range [$0.99$ - $1$], for a better visualisation.
    }
    \label{fig:results-front-adult}
\end{figure}

The results on Adult Census Income (Figure~\ref{fig:results-front-adult}) show that the protection of the sensitive attribute can be difficult to achieve on complex distributions. In fact, simpler approaches such as \dirm{} and \dirm{}-OM do not provide a protection measured with either $\ber{}_{rc_{prv}}$ or $\sac{}_{rc_{prv}}$ better than the baseline of $0.16$ and $0.85$ respectively. In some cases, those approaches modifications even worsen the protection. \gansan{}-OM also worsen the results. These observations suggest that the reconstruction of the protected group (recall that \gansan{}-OM has to protect the sensitive attribute by modifying only the protected data, while limiting the amount of modification introduce with the reconstruction of the protected group) hinder the quality of the protection, and prevent the appropriate modification of the data in order to protect the sensitive attribute. 

$\gansan{}$ and $\attgan{}$ achieve some of the highest fidelities, with \attgan{} slightly improving the protection for the highest fidelity, and outperforming $\gansan{}$ on the transformation. Our approach $\fairmapping{}$ dominates all other approaches except $\wgan{}$ by providing the highest protection for the highest level of fidelities. The maximum protection with $\ber$ ($0.33$) is almost the double of the value on the original dataset ($0.16$). At that point, the transformation $\classification_{prot}$ is around $0.77$.
$\wgan{}$ has the highest protection achievable among all approaches. This level of protection comes with a significant cost on the fidelity. Nonetheless, $\wgan{}$ is the only approach that has the perfect transformation ($\classification_{prot}$) for all results on its Pareto front. 

\paragraph{Adult Census with 2 sensitive attributes results}
\begin{figure}[ht]
        \includegraphics[width=1\linewidth, height=0.45\textheight]{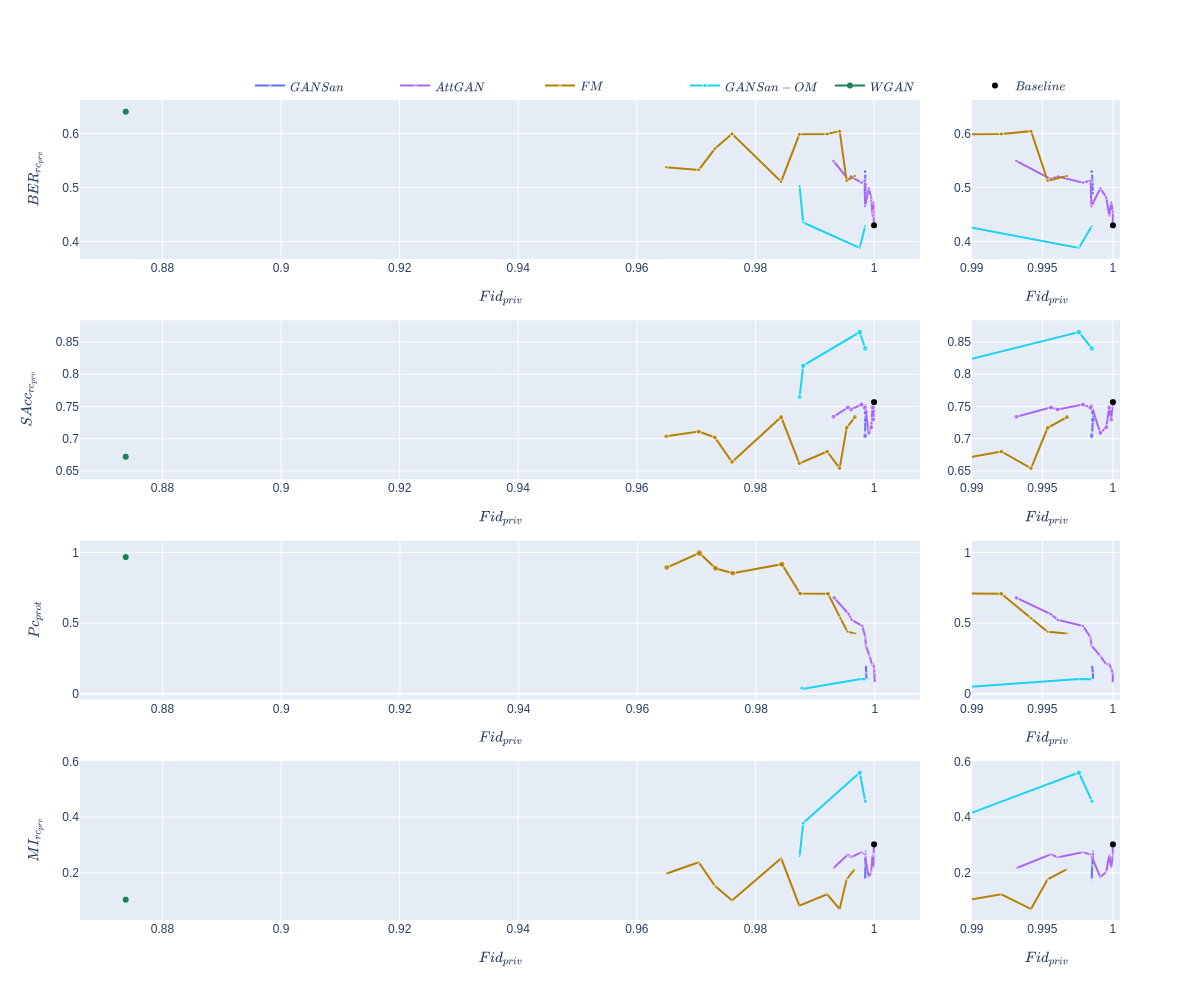}
    \caption{Pareto fronts in the \textit{Fair Mapping} perspective for approaches investigated on Adult Census with 2 attributes. The right column present all points on the fronts, while the right columns present the same results but on the range [$0.99$ - $1$], for a better visualization. \dirm{} and \dirm{}-OM are not represented since these approaches could not be adapted to the multiattribute case.
    }
    \label{fig:results-front-adult2}
\end{figure}
We showcase the results of the protection of more than 2 groups on Adult census with 2 attributes combined as a single one. we dubbed this dataset as \textit{Adult2}. Results are presented on Figure~\ref{fig:results-front-adult2}. Note that the best protection is achieved at a $\ber$ value of $\dfrac{3}{4}$, a $\sac{}$ of $0.6$.

The protection with more than is single attribute is harder to achieve with all approaches. Only FairMapping is able to protect the sensitive attribute while improving the transformation of the protected group. We can observe that only $\wgan{}$ and $\fairmapping{}$ are able to provide a classification $\classification{}_{prot}$ greater than $0.5$. Similarly, both approaches achieve a similar level of mutual information, significantly lower than the baseline. Just as observed on other datasets, $\gansan{}-OM$ is unable to improve any metrics above the baseline.

The highest protection of $\fairmapping{}$ measured with $\ber{}_{rc_{prv}}$ is achieved at the value of $0.632$. At that point, the $\sac{}_{rc_{prv}}$ measured is $0.67$, the fidelity is $0.9946$ and the transformation has the value of $\classification{}_{prot} = 0.662$. Closely located to this point, the highest protection measured with the $\sac{}_{rc_{prv}}$ has a better transformation result but at a slight cost on the fidelity: $\ber{}_{rc_{prv}} = 0.63$, $\sac{}_{rc_{prv}} = 0.65$, $\classification{}_{prot} = 0.72$, $\fid{}_{prv} = 0.9930$. At these trade-offs, we relinquish a bit of the fidelity and protection for a significant improvement on the transformation quality of the protected group.

We can also explain our results by the fact that we used the model structure that maximizes the fidelity. As such, it might be possible to further improve the results by relaxing the fidelity and using another model not pre-optimized for this metric. However, this relaxation will increase the space of hyperparameters and might result in an increase of the training time. 

\paragraph{Divergences}
As FairMapping transports the protected data onto the privileged distribution, we measure the proximity between the transformed protected distribution and the known privileged group. We rely on the \textit{Sinkhorn} divergence~\cite{cuturi2013sinkhorn} to compute the divergence between the original privileged group and the transformed protected data ($R_{priv} \sim G_{w}(R_{prot})$), between the reconstructed privileged group (obtained through $G_{w}$) and the transformed protected group ($G_{w}(R_{priv}) \sim G_{w}(R_{prot})$) and finally between the original protected data and their transformed version ($R_{prot} \sim G_{w}(R_{prot})$). The objective here is to obtain a smaller $R_{priv} \sim G_{w}(R_{prot})$ and $G_{w}(R_{priv}) \sim G_{w}(R_{prot})$ than $R_{prot} \sim G_{w}(R_{prot})$. Ideally, we wish to obtain the value of $0$ for $R_{priv} \sim G_{w}(R_{prot})$ and $G_{w}(R_{priv}) \sim G_{w}(R_{prot})$. 

\begin{figure}[ht]
        \includegraphics[width=1\linewidth, height=0.3\textheight]{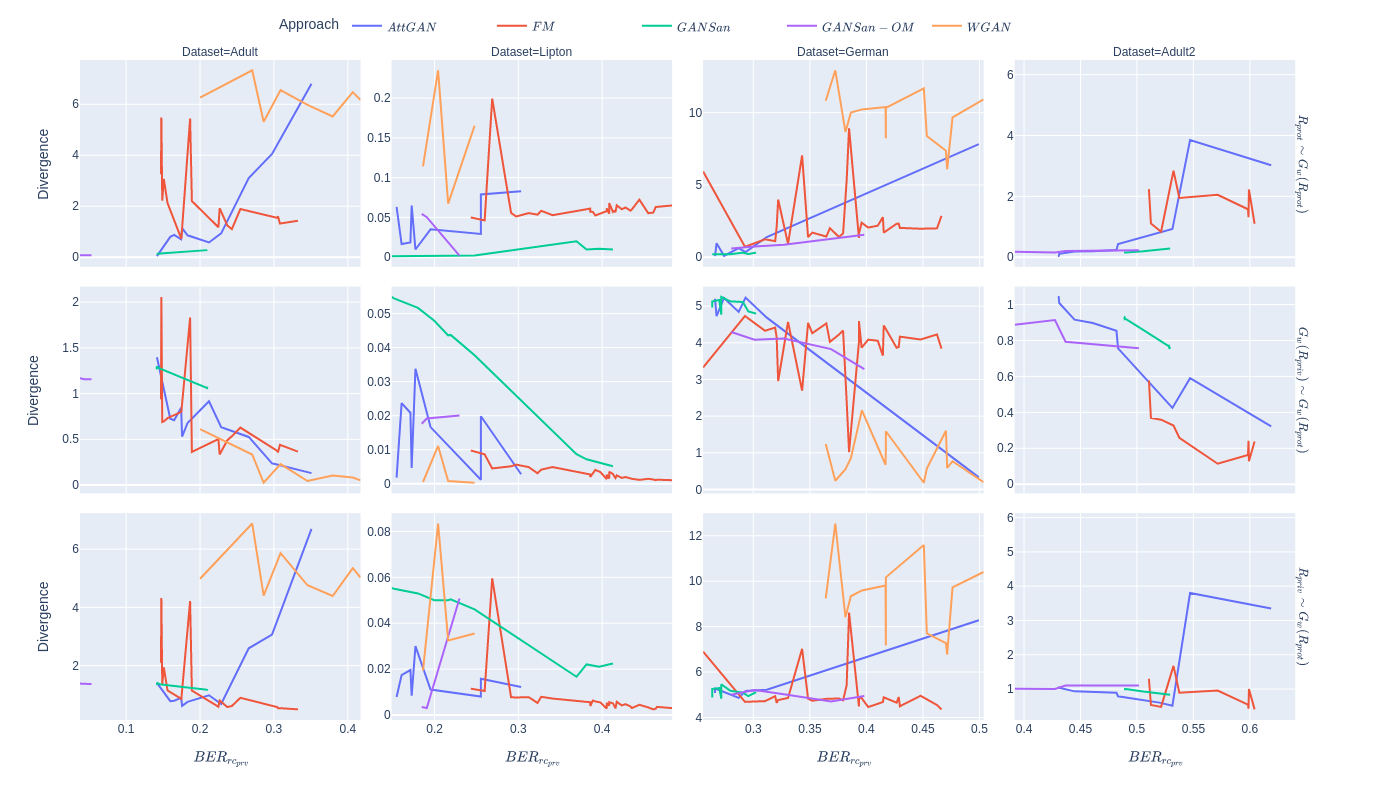}
    \caption{Divergences computed on German, Adult, Adult2 and Lipton. Each column represent a dataset while each row represent the divergence between the transformed protected data and respectively the original protected data ($R_{prot} \sim G_{w}(R_{prot})$), the reconstructed privileged group ($G_{w}(R_{priv}) \sim G_{w}(R_{prot})$), and the original privileged group ($R_{priv} \sim G_{w}(R_{prot})$). The divergences are represented with respect to the protection $\ber{}_{rc_{prv}}$.
    }
    \label{fig:divergence-ber-r-m}
\end{figure}

We present in figures~\ref{fig:divergence-ber-r-m} \ref{fig:divergence-ber-o-m} (appendix~\ref{sec:div-org}) the computed divergences with respect to the fidelity on the privileged group $\fid{}_{priv}$, and in figure~\ref{fig:divergence-fid-prv} the divergence with respect to the protection obtained with the same models on the Pareto fronts (\ref{fig:results-front-lipton}, \ref{fig:results-front-adult}, \ref{fig:results-front-german}, \ref{fig:results-front-adult2}) 

We can observe on the divergences that the higher the protection, the lower the divergences value of $G_{w}(R_{priv}) \sim G_{w}(R_{prot})$ and the higher $R_{prot} \sim G_{w}(R_{prot})$. Our approach is able to achieve the lowest divergence results with higher protection values. We can also observe that the \textit{Sinkhorn} divergence between the original protected data and its transformed version ($R_{prot} \sim G_{w}(R_{prot})$) decreases as the fidelity increases. Nevertheless, its values are still higher than the divergences computed with the privileged group ($R_{priv} \sim G_{w}(R_{prot})$ and $G_{w}(R_{priv}) \sim G_{w}(R_{prot})$). This suggests that as the fidelity increases, our model in $\fairmapping{}$ able to learn the structure of the data and output a dataset with less random values.

We can also observe that the intermediate distribution obtained with other preprocessing approaches ($\gansan{}$, $\attgan{}$) are closer to the protected distribution, but far from the privileged one. These results are surprising, especially on the German credit dataset, where the privileged group is largely present. The models produce data in which one does not know what to expect as outcomes.
With FairMapping, the transformed distribution is closer to the privileged group distribution. As a consequence, a classifier trained on the original data would easily consider the transformed datapoint obtained with FairMapping as belonging to the same decision frontier as the privileged group. 
It is also worth noting the protection property, notably the accuracy $sac{}_{rc_{prv}}$ can be used as a surrogate of the $\mathcal{A}-distance$~\cite{ben2006analysis} (the $proxy-\mathcal{A}-distance$~\cite{aubreville2021quantifying}), which is a divergence metric between domains.

\begin{figure}[ht]
        \includegraphics[width=1\linewidth, height=0.45\textheight]{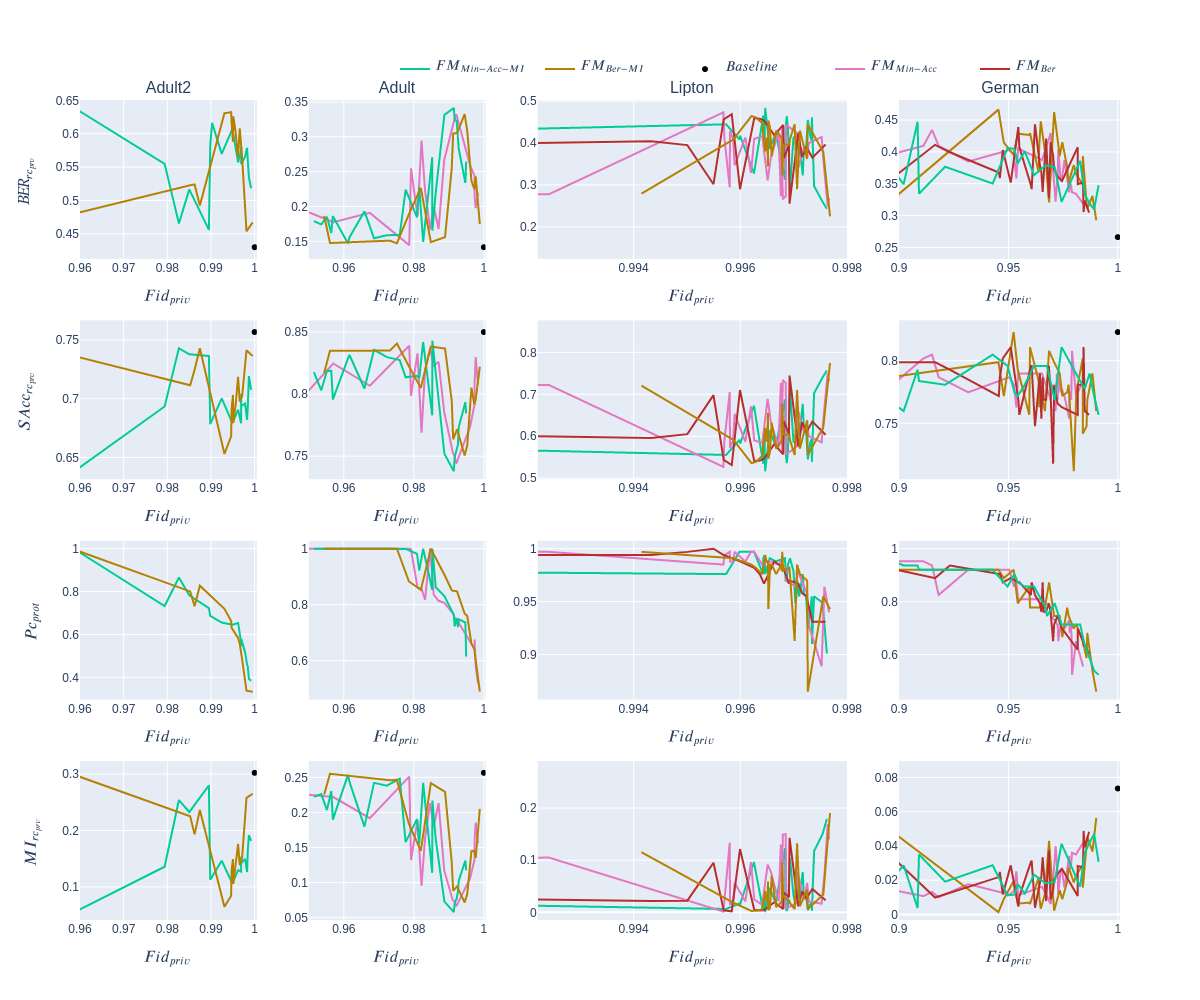}
    \caption{Different \textit{Fair Mapping} instances. We can observe that the $\fairmapping{}$ instances behave almost similarly for all chosen loss function for protecting the sensitive attribute. Each column represent the performances on a single dataset. On Adult2, we always include the mutual information regularisation.
    }
    \label{fig:results-all-fm-rec}
\end{figure}

\paragraph{FairMapping instances}
As mentioned in section~\ref{sec:fairmapping}, our approach can be instantiated with either the $\ber{}$ or the accuracy, each with and without the mutual information regularization. The results of the different instances are presented in Figure~\ref{fig:results-all-fm-rec}.

We can observe that the different instances of \fairmapping{} behave almost similarly on all dataset we have experimented with, the mutual information slightly improves the results where it is introduced: 
when the number of groups is greater than $2$ ($k > 2$), we always add the mutual information regularization.
On the German dataset, we can observe that the mutual information regularization helps improve the $\ber{}$ (all $\fairmapping{}$ that uses the $MI$ are slightly better than those without.). On Lipton, the dataset is balanced. The mutual information does not have a significant impact as observed on German. This suggests that in situation where the dataset in highly unbalanced, introducing the mutual information regularization would improve the quality of results.

\subsection{Fairness evaluation}
\label{sec:fairnes-results}
Once we computed the Pareto-front displaying the achievable trade-offs between the \textit{identity}, the \textit{protection} and the \textit{classification}, we select the best trade-off on which we further compute the transformed dataset utility and other fairness metrics. 

The best trade-off is obtained by selecting the set of points on the Pareto front that minimizes equation~\ref{eq:selection-equation}.
\begin{equation}
    best = \alpha * (\ber{}_{rc_{prot}} - \dfrac{k - 1}{k})^{2} + \beta * MI^{2} + \gamma * (\classification{}_{prot} - 1)^{2} + \delta * (\fid_{priv} - 1)^{2}
    \label{eq:selection-equation}
\end{equation}
Equation~\ref{eq:selection-equation} computes a distance between the performances of a model and the ideal results we would like to obtain, namely a protection $\ber{}_{rc_{prot}}$ equal to $\dfrac{k - 1}{k}$, a mutual information of $0$, a classification $\classification{}_{prot}$ of $1$ and a the perfect fidelity of $1$.  We empirically set the values of coefficients ($\alpha, \beta, \gamma, \delta$) to respectively ($1, 0, 0.2, 1$) when there are only two groups (the mutual information is less important), and to ($1, 1.7, 0.2, 1$), when there are more than $2$ groups in the dataset. The classification is the easiest to achieve, thus we lowered its importance relatively to other parameters of the equation.


With the model selected we extracted its training hyperparameters, and we evaluate the generalization capabilities of such model by conducting a cross-validation with $3-folds$, and average the metrics over the folds.

We evaluate the data utility and the fairness obtained.

\emph{Dataset utility.} As our approach belongs to the fairness pre-processing family, it generates a transformed dataset that could be used for several subsequent analysis tasks.
We quantify the utility preserved by our transformation by computing the amount of perturbations introduced into the categorical attributes $\catdam{}$, which, in our setup, is only computed on the privileged group ($\catdam{}=\catdam{}_{priv}$). In addition, we also computed the $fidelity$ previously defined.

The accuracy $\accuracy{}_{y}$ of external classifiers when trained to predict the decision attribute using the transformed dataset allows us to measure the amount of information preserved on a specific task.

Finally, we also compute a \emph{diversity} metric introduced originally in~\cite{aivodji2021local}.
\begin{equation}
    \diversity{}  = \dfrac{\sum_{i=1}^{N}\sum_{j=1}^{N}\sqrt{\sum_{k=1}^{d} (r_{i,k} - r_{j,k})^2}}{N \times (N-1) \times \sqrt{d}}
    \label{eq:divers}
\end{equation}
The objective of this metric is to measure the spread of the transformed data points. 
Thus, it facilitates the detection of “poor transformations'' (\emph{e.g.}, all data points are mapped onto a single data point in a specific cluster of the target distribution, \emph{cf.} Section~\ref{subsec:fair:overview}).
This diversity is computed on both the original and transformed datasets, with the difference between these two quantities indicating the loss of diversity incurred by the transformation.
Indeed, a lower diversity of the transformed dataset corresponds to a decrease of the combinations of attributes-values.

\emph{Fairness.} To quantify group fairness, we rely on the metrics described in Section~\ref{sec:background}, namely demographic parity $\demoPar{}$ and $\eqgap{}$. 
Recall that the demographic parity is satisfied if the difference of positive rates $Pos.Rate$ across groups is less than a threshold $\epsilon$. 
The $\eqgap{}$ with $y=1$ (respectively $y=0$) corresponds to the difference of true positive rates $Tp.Rate$ (respectively false positive rates $Fp.Rate$) across groups.

To compute the metrics on this step, we rely on a different set of external classifiers composed of Multi Layer Perceptron (MLP), Decision Tree (DT) and logistic regression (LR). These classifiers parameters are described in Table~\ref{tab:exthprms}. Just as carried out at the comparison step, for the classification and the protection metrics, we report the best metric obtained among all external classifiers. We report only the MLP classifier results for the task accuracy ($\accuracy{}_{y}$) since it provides the best task accuracy among all external classifier on the original data. 

At the fairness step, the use of a different set of classifiers  than those at the comparison step is to ensure the independence of the metric computation to the optimization process, to ensure that the optimization procedure does not produce results that only best fit the known classifiers.

\paragraph{Use cases}
We identify 3 use cases through which we can evaluate the fairness of our approach : the \textit{data publishing}, \textit{fair classification} and \textit{local sanitization}. These use cases have been originally proposed by~\cite{aivodji2021local}.
We will only present results for the \textit{data publishing} and \textit{local sanitization} use cases. Results for the \textit{fair classification} are presented in appendix~\ref{sec:fair-res}.

\begin{itemize}
\item \emph{Data publishing.} In this scenario, the data is transformed for publishing or sharing purposes and can be used afterwards for a variety of subsequent analysis tasks. 
This scenario is motivated by the fact that the data curator (\emph{e.g.}, a statistical agency) might be interested in releasing a version of its data which could be used by interested parties for various data analysis objectives. 
In this scenario, a classifier is trained on the output of $\fairmapping{}$ to predict a specific task, and the test set is also composed of only the dataset transformed with \fairmapping{}.

\item \emph{Local sanitization.} To detect whether a system makes a decision based on the sensitive attribute, one could compute the decision using the original data, and compare it to a decision obtained using our transformed profile, which is not correlated to the sensitive attribute. 
Similarly, an individual who does not trust a decision system and wish to maintain his sensitive attribute private, could rely on our approach to transform his data locally (\emph{e.g.}, on his smartphone) before releasing the transformed version of his data. 
The local sanitization scenario assumes the existence of a classifier trained on the original data (with the existing biases).

Our approach, in contrast to most of the prior works, also offer the advantage that an individual who had locally transformed his data would benefit from the same advantages as the privileged group, from the point of view of the classifier trained on the original data. 
This advantage is a direct consequence of our \textit{transformation} property, in which we constrain the models to transform data points such that a trained classifier would predict them as part of the target group. 
Another benefit of our approach is that it does not require the classifier to be retrained from scratch on the transformed dataset, thus limiting the operational complexity (\emph{i.e.}, training time, hyperparameters selection, deployment of the best model, etc.) of fairness improvement.

In this scenario, we suppose that a classifier trained on the original unmodified data exists and cannot be modified. Thus, we test the prediction of the classifier using the data obtained with \fairmapping{}.

\end{itemize}
Table~\ref{tab:scenarios} provides an overview of the composition of the dataset for each of these use cases.

\begin{table}[h!]
\caption{Use cases envisioned for the evaluation of Fair Mapping. 
Each dataset is composed of either the original attributes or their transformed versions. 
$Transformed$ refers to the transformed dataset, which corresponds to the case where only the protected group is transformed while the privileged group is kept original ($\protectedmapped{}$), and the case where the dataset is composed of the reconstructed privileged group and the transformed protected group data ($\allmapped{}$). 
For all use cases except the baseline, the results are computed using only the original decisions}
    \resizebox{\columnwidth}{!}{%
        \begin{tabular}{ccccc}

            \toprule
            \multirow{2}{*}{\textbf{Scenario}} & \multicolumn{2}{c}{\textbf{Training set composition}} & \multicolumn{2}{c}{\textbf{Test set composition}} \\
            \cmidrule{2-5}
            & \emph{\textbf{A}} & \emph{\textbf{Y}} & \emph{\textbf{A}} & \emph{\textbf{Y}} \\

            \midrule
            \textbf{Baseline}            & $Original$          & $Original$         & $Original$          & $Original$            \\
            \textbf{Data publishing}     & $Transformed$       & $Original$         & $Transformed$       & $Original$         \\
            \textbf{Fair classification} & $Transformed$       & $Original$            & $Original$       & $Original$            \\
            \textbf{Local sanitization}  & $Original$          & $Original$            & $Transformed$       & $Original$            \\
            \bottomrule
        \end{tabular}
    }
    \label{tab:scenarios}
\end{table}

\subsubsection{Fairness evaluation: Results}
\begin{table}[]
\caption{Performances on the selected trade-offs for which we will conduct the cross-validation and the fairness analysis}
\label{tab:selected-perfs}
\resizebox{\columnwidth}{!}{%
    \begin{tabular}{ccccc}
        \hline
        Dataset                      & Adult    & Adult2   & German   & Lipton   \\ \hline
        $\fid{}_{priv}$            & $0.9945$ & $0.9760$ & $0.9454$ & $0.9964$ \\
        $\ber{}_{rc_{prv}}$     & $0.3324$ & $0.5991$ & $0.4664$ & $0.4834$ \\
        $\sac{}_{rc_{prv}}$     & $0.7506$ & $0.6637$ & $0.5165$ & $0.7987$ \\
        $MI_{rc_{prv}}$          & $0.0717$ & $0.1001$ & $0.0013$ & $0.0005$ \\
        $\classification{}_{prot}$ & $0.7686$ & $0.8534$ & $0.9759$ & $0.9206$
    \end{tabular}
}
\end{table}

We present in Table~\ref{tab:selected-perfs} the metrics for the trade-offs selected by the equation~\ref{eq:selection-equation}. On Table~\ref{tab:fairness-protection-results-mean}, we present the final results of the cross validation using the selected trade-offs hyperparameters. The standard deviations are available in section~\ref{sec:std-fairness-results}

\begin{table}[]
\caption{Cross validation protection results for all experimented datasets.
We can observe that our approach produces stable results for most datasets, as measures with a different set of external classifiers.
Refers to Table~\ref{tab:fairness-protection-results-std} for the standard deviations}. The baseline of the fidelity is obtained by measuring the distance between the original dataset and a random permutation of its columns.
\label{tab:fairness-protection-results-mean}
\resizebox{\columnwidth}{!}{%
\begin{tabular}{cccccccccc}
\cline{1-10}
\multicolumn{2}{c}{}                  & \multicolumn{2}{c}{Adult}                  & \multicolumn{2}{c}{Adult2}                 & \multicolumn{2}{c}{German}                 & \multicolumn{2}{c}{Lipton}                 \\ \cline{1-10} 
                              &       & Baseline & \fairmapping{} & Baseline & \fairmapping{} & Baseline & \fairmapping{} & Baseline & \fairmapping{} \\ \cline{2-10} 
\multirow{8}{*}{Metrics}      
& $\ber{}_{rc_{prv}}$ &  $0.1395$ &  $0.2604$  &  $0.4382$  &  $0.5201$  & $0.2797$ &  $0.3727$ &  $0.1870$  &  $0.4090$ \\
& $\ber{}_{og_{prv}}$ &  $0.1395$ &  $0.0528$  &  $0.4382$  &  $0.3270$  & $0.2797$ &  $0.1934$ &  $0.1870$  &  $0.3230$ \\
& $\sac{}_{rc_{prv}}$ &  $0.8506$ &  $0.7872$  &  $0.7548$  &  $0.7234$  & $0.8100$ &  $0.7840$ &  $0.8130$  &  $0.5910$ \\
& $\sac{}_{og_{prv}}$ &  $0.8506$ &  $0.9627$  &  $0.7548$  &  $0.8992$  & $0.8100$ &  $0.8880$ &  $0.8130$  &  $0.6770$ \\
& $MI_{rc_{prv}}$ &  $0.2607$ &  $0.1223$  &  $0.2941$  &  $0.2394$  & $0.0664$ &  $0.0232$ &  $0.2154$  &  $0.0187$ \\
& $MI_{og_{prv}}$ &  $0.2607$ &  $0.4785$  &  $0.2941$  &  $0.7350$  & $0.0664$ &  $0.1584$ &  $0.2154$  &  $0.0656$ \\
& $\classification{}c_{prot}$ &  $-$ &  $0.6857$  &  $-$  &  $0.3947$  & $-$ &  $0.7945$ &  $-$  &  $0.8949$ \\
& $Fid_{priv}$ &  $0.9049$ &  $0.9962$  &  $0.9050$  &  $0.9550$  & $0.7367$ &  $0.9346$ &  $0.6536$  &  $0.9975$ \\
\end{tabular}
}
\end{table}

We can observe that, overall, our approach provides a protection that is around the double of the baseline. All metrics are reduced when carrying out the cross-validation, especially the classification $\classification{}_{prot}$. This observation suggests that the selected hyperparameters does not generalize very well the transformation property. We believe that a larger exploration of the hyperparameters space will produce models with better performances. 
Moreover, we can observe on Table~\ref{tab:fairness-protection-results-std} that the approach is not very stable across the experimentation. We can see that the transformation metric $\classification{}_{prot}$ has a very large deviation, especially on Adult and Adult2 datasets. As some protected group sizes are less than $10\%$ in Adult2 dataset, the approach did not correctly learn the representation of those distributions in order to correctly carry out the transport. In fact, as the classifier $C$ is trained on the original data until convergence prior its use by the mapping models, we have observed that the best performances achievable with that model is around $75\%$, which might not be enough for the model $C$ to correctly guides $G_{w}$ during the mapping procedure.

\begin{table}[]
    \caption{Mean of diversity computed on all 4 datasets}
    \label{tab:diversity}
    \resizebox{\columnwidth}{!}{%
    \begin{tabular}{ccccccc}
        \hline
        \multirow{2}{*}{} & \multicolumn{2}{c}{Whole dataset} & \multicolumn{2}{c}{Protected group} & \multicolumn{2}{c}{Privileged group} \\ \cline{2-7} 
                          & Baseline       & fairmapping      & Baseline        & Fairmapping       & Baseline        & Fairmapping        \\ \hline
        Adult2 & 0.2755 & 0.1795 & 0.2786 & 0.1422 & 0.2622 & 0.1918  \\
        Adult & 0.2772 & 0.2605 & 0.2777 & 0.2621 & 0.2660 & 0.2588  \\
        Lipton & 0.2474 & 0.1593 & 0.2314 & 0.1622 & 0.1756 & 0.1543  \\
        German & 0.5040 & 0.4493 & 0.4785 & 0.4304 & 0.5056 & 0.4516  \\   
        \end{tabular}
    }
\end{table}

\begin{figure}[ht]
        \includegraphics[width=1\linewidth, height=0.2\textheight]{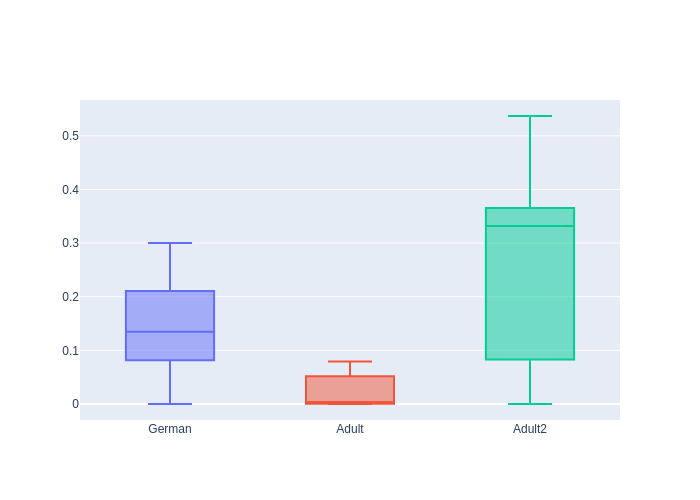}
    \caption{Categorical damage observed on datasets German Credit and Adult Census. Lipton does not have any categorical columns, thus, it is not represented here.
    }
    \label{fig:cat-damage}
\end{figure}

Concerning the datasets utility, we present in Table~\ref{tab:diversity} the diversity obtained with the cross-validation, and Figure~\ref{fig:cat-damage} present the damage on categorical columns, which correspond to the number of modified rows of a dataset that are not numeric.
With $2$ sensitive attributes, the damage on Adult2 is quite important, even though the fidelity is close to the one observed on Adult. The respective median values are $0.1345$, $0.0.0036$, $0.3317$ for German credit, Adult and Adult2. This suggests that for German credit, half of the categorical columns have a damage less than $13\%$, while the other half is located between $13\%$ and $30\%$. In other words, $13$ over $100$ individuals from the privileged group would have a different value on the reconstructed data than on the original one. For Adult2, ensuring the identity appears to be highly difficult. This difficulty is also translated into the poorer diversity results of the privileged group on the same set. 

\paragraph{Dataset Publishing}
\begin{figure}[ht]
        \includegraphics[width=1\linewidth, height=0.15\textheight]{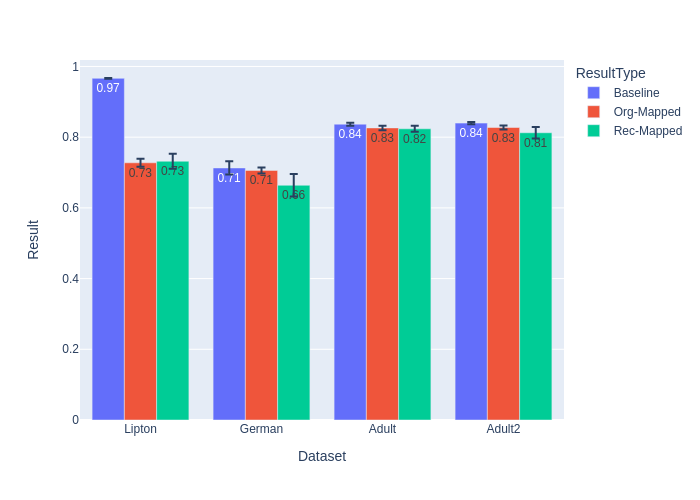}
    \caption{Accuracies achieved with the MLP classifier on the Data publishing scenario. The black vertical bar indicates the standard deviation across all computed folds. \textit{Org-Mapped} indicates the results computed with the original privileged group and the transformed protected data, whereas \textit{Rec-Mapped} uses the reconstructed privileged group and the transformed protected group data. The baseline is computed on the original unmodified dataset.
    }
    \label{fig:results-acc-data-pub}
\end{figure}
In this scenario, our approach can achieve the accuracy presented in Figure~\ref{fig:results-acc-data-pub}. We can observe that $\fairmapping{}$ significantly affect the prediction accuracy, especially on the Lipton dataset. On any other dataset, the accuracy remains stable and near the baseline. By looking at the prediction in each group (Figure~\ref{fig:results-gr-acc-data-pub}), we can observe that the model makes nearly the same prediction rate as in the baseline. This suggests that our approach is still able to preserve the utility of the dataset in terms of task prediction.

\begin{figure}[ht]
        \includegraphics[width=1\linewidth, height=0.15\textheight]{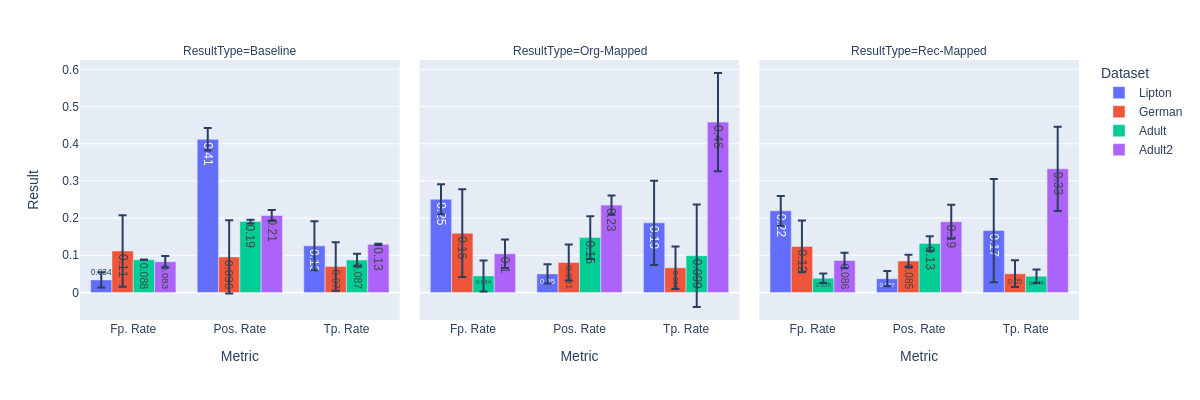}
    \caption{Fairness results computed on the data publishing scenario. Each column represent a different type of dataset usage. From left to right, we have the baseline (the original unmodified dataset), the dataset with only the protected group modified ($Org-Mapped$), and the dataset with the protected group transformed while the privileged group is reconstructed ($Rec-Mapped$).
    Each metric represented in this figure correspond to the maximum difference of the metric, observed between the privileged and the protected group.
    }
    \label{fig:results-fair-data-pub}
\end{figure}

Figure~\ref{fig:results-fair-data-pub} presents the different measures of fairness obtained with our experiments, namely the $\demoPar{}$ and $\eqgap{}$. Note that the demographic parity requires the computation of the amount of decisions predicted as positives $Pos. Rate$. The equalized odds as well as the equality of opportunities require the computation of the true positives rates $Tp. Rates$ and false positives rates $Fp. Rates$. In Figure~\ref{fig:results-fair-data-pub} we present the maximum difference between $Pos. Rate$, $Tp. Rate$ and $Fp. Rate$ across all groups. We can observe that \fairmapping{} is able to reduce the metric gap in nearly all datasets, except $Adult2$. With the use of the original privileged data, the difference in predictions becomes more apparent. By looking at Figure~\ref{fig:results-fair-gr-data-pub} which presents the metric computed in each group, we can have a deeper understanding of our approach. 
On Lipton and German dataset, we can observe that the metrics computed within each group are reversed. The false positive rate becomes higher in the protected group, while it becomes closer to zero in the privileged one.Thus, it suggests that the approach has increased the amount of positive decision attributed in the protected group, as we can observe with the metric $Pos. Rate$. The decrease of $Pos. Rate$ in the privileged group might be due to a poorer reconstruction of that group. With these observations, we can conclude that \fairmapping{} has transferred the properties of the privileged group onto the protected one. 
On Adult, as the transformation is not high enough, the metric follows nearly the same trend as in the baseline.
Adult2 is highly unstable, as the standard deviations are very large for each metric computed in the different dataset groups.

\begin{figure}[ht]
        \includegraphics[width=1\linewidth, height=0.15\textheight]{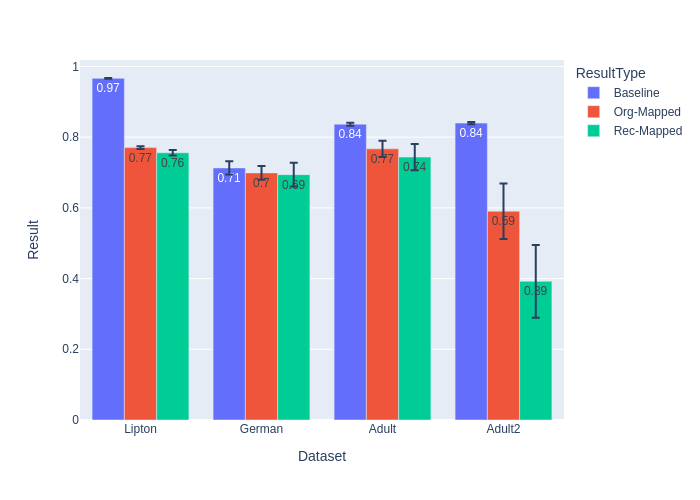}
    \caption{Accuracies achieved with the MLP classifier on the Local sanitisation scenario. The black vertical bar indicates the standard deviation across all computed folds.
    }
    \label{fig:results-acc-local-san}
\end{figure}

\paragraph{Local Sanitization}
We present in Figure~\ref{fig:results-acc-local-san} the local sanitization scenario, where $\fairmapping{}$ modifies the data before their usage on a classifier trained with the original data to predict the decision.
We can observe that the results on the local sanitization are quite similar to those in the data publishing scenario, for dataset that has a high level of protection and transformation ($Lipton$ and \textit{German}). Interestingly, on Lipton, German and Adult datasets, even though the accuracy is the same as in the data publishing scenario, the group accuracies are completely different (Figure~\ref{fig:results-gr-acc-local-san}). We can observe that the group accuracy on the privileged data closely follows the baseline, but on the protected group, the accuracy significantly drops. This is mainly due to the fact that the transformed protected group closely resemble the target distribution, thus as the original decision are strongly biased, the correlation with the original decision are removed. The classifier predict the transformed protected group as if they are the privileged group. Recall that the classification $\classification{}_{prot}$ metric is greater than $70\%$ for these datasets.

\begin{figure}[ht]
        \includegraphics[width=1\linewidth, height=0.15\textheight]{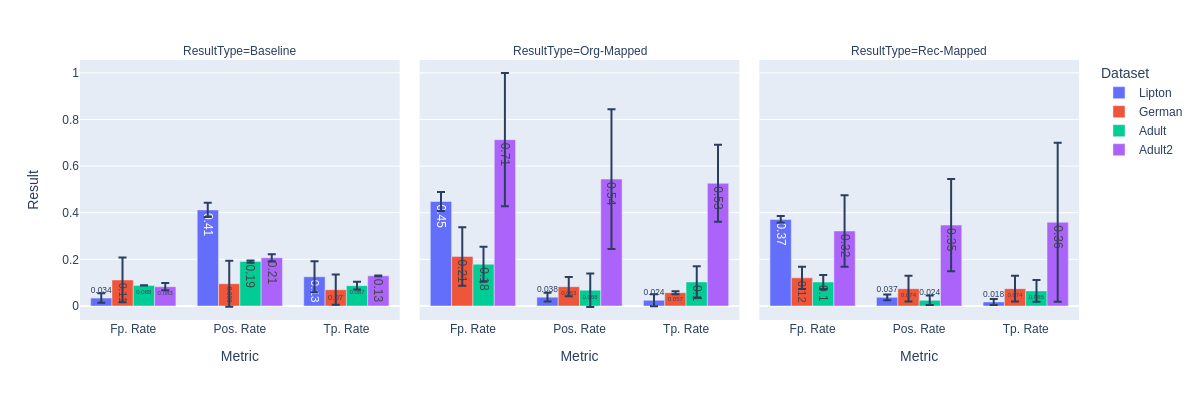}
    \caption{Fairness results computed on the Local Sanitisation scenario. Each column represent a different type of dataset usage. From left to right we have the baseline (the original unmodified dataset), the dataset with only the protected group modified ($Org-Mapped$) and the dataset the protected group transformed and the privileged group reconstructed ($Rec-Mapped$).
    }
    \label{fig:results-fair-local-san}
\end{figure}

\begin{figure}[ht]
        \includegraphics[width=1\linewidth, height=0.15\textheight]{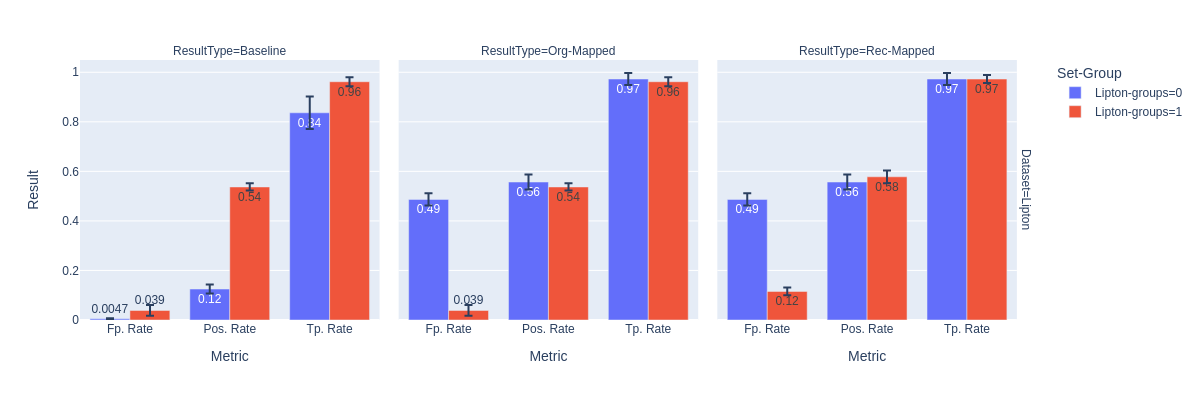}
        \includegraphics[width=1\linewidth, height=0.15\textheight]{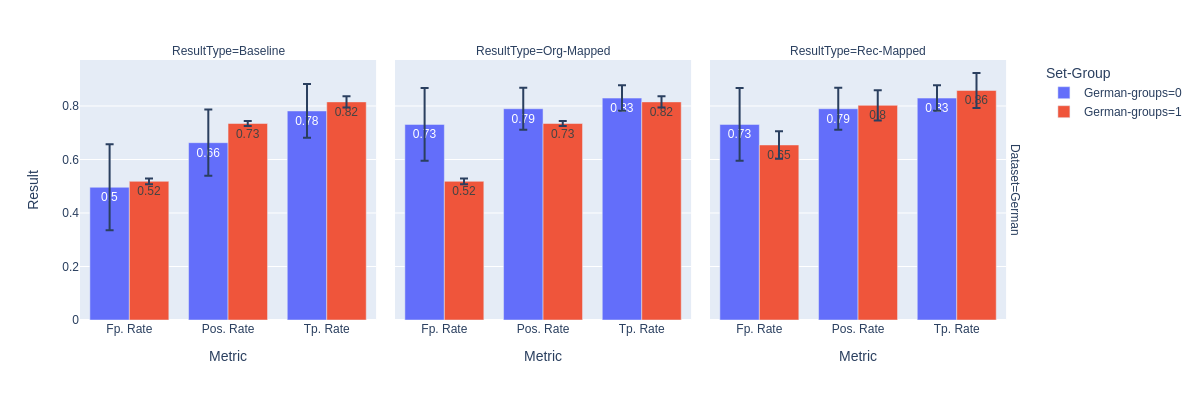}
        \includegraphics[width=1\linewidth, height=0.15\textheight]{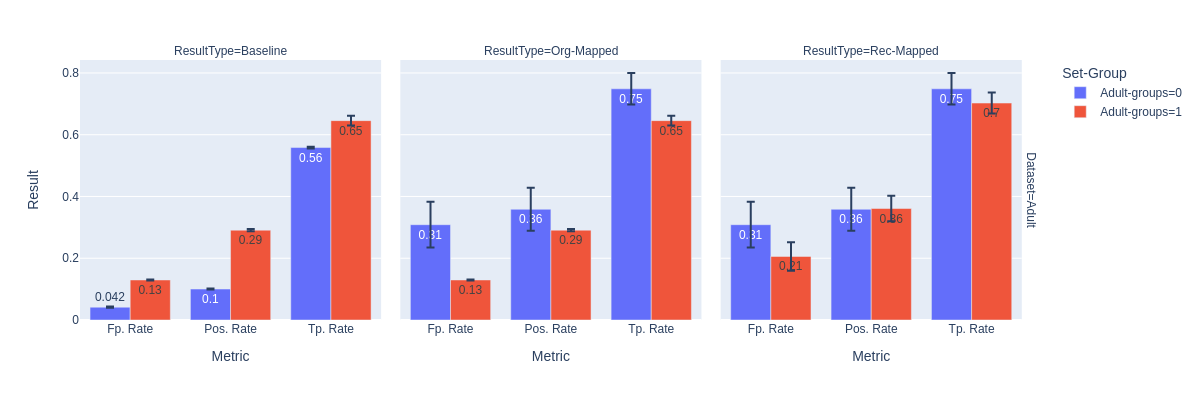}
        \includegraphics[width=1\linewidth, height=0.15\textheight]{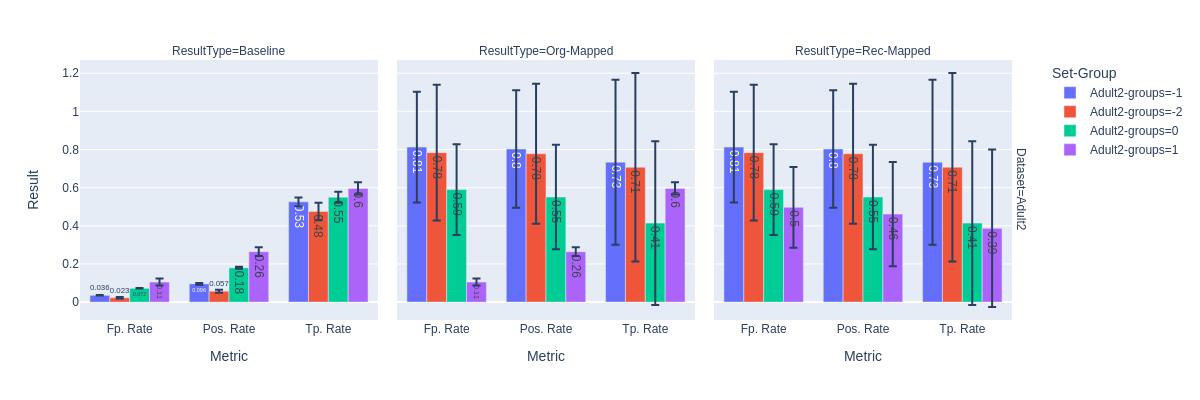}
    \caption{Accuracy of the task decision at each group level for the scenario data publishing.
    }
    \label{fig:results-fair-gr-local-san}
\end{figure}

As the predictions made with the transformed protected group resemble those of the privileged one, the false positive rate of the protected group will be higher. This is why we can observe the results on figure~\ref{fig:results-fair-local-san} (which represent the difference of fairness metrics computed within each group). By looking at the detailed metrics within each group (Figure~\ref{fig:results-fair-gr-local-san}), we can confirm that the false positive rate of the protected group has significantly increased, while it has not changed for the privileged one. The demographic parity as well as the equality of opportunity are both significantly reduced on all dataset except Adult with 2 attributes. 

These results suggest that our approach \fairmapping{} is able to appropriately transfer the property of the privileged group onto the protected one, while preventing the sensitive attribute inference. More experiments as well as a better and larger hyperparameters exploration are necessary when dealing with multiple attributes, to improve the quality of results as well as the stability across different folds of the cross validation.

\subsection{Execution times}
We present in Table~\ref{tab:time_requirements} the mean execution time for each approach to complete one epoch (one loop over the entire training dataset) of computation on the Lipton dataset. $\dirm{}$ takes less than $10$ seconds for the approach to complete, therefore it is not represented in the table. 
\begin{table}[h!]
    \caption{Mean execution time computed on the Lipton dataset, computed with a single GPU}
    \label{tab:time_requirements}
    \resizebox{\columnwidth}{!}{%
        \begin{tabular}{ccccccc}
            \hline
            Approach                        & \fairmapping{} & \fairmapping{} with MI & \gansan{} & \gansan{}-OM & \attgan{} & \wgan{} \\ \hline
            Time per epoch (seconds)                  & $0.224337$                      & $0.236976$                              & $0.126364$                 & $0.140211$                    & $0.141715$                 & $0.183009$               \\
            Time for 1000 epochs \textit{(mm:ss)} & $0:03:44$                       & $0:03:56$                               & $02:06$                  & $02:20$                     & $02:21$                  & $03:03$                \\ \hline
        \end{tabular}
    }
\end{table}
Our approach $\fairmapping{}$ takes the longest time to complete, while $\gansan{}$ is the fastest (excluding \dirm{}). The is due to the fact that our approach has the largest number of models involved in the computation. The execution time is improved by the fact that the classifier $C$ can be trained independently of the core models of our approach ($G_{w}$, $D$ and $D_{std}$). 

In our experiments with the Adult Census dataset, found out that on a GPU with $4Go$ of memory, we can fit up to $3$ different running instances of $\fairmapping{}$ with some residual memory left, $4$ instances of $\gansan{}$, $\gansan{}-OM$ and $\wgan{}$, and $2$ instances of $\attgan{}$. As a consequence, the hyperparameters exploration can be a bit faster for $\fairmapping{}$.
\section{Conclusion}
\label{sec:conclusion}

In this paper, we proposed a novel approach Fair Mapping that transforms a given data point onto a chosen target distribution while preventing the possibility of inferring the sensitive attribute.
In particular, our approach preserves the realistic aspect of the transformed dataset while providing the transformed data points with the same benefit as the privileged group from the point of view of a trained classifier. 
As a consequence, a classifier already trained does not need to be re-trained to use the transformed dataset, which is usually required with other pre-processing approaches. \ros{Nevertheless, our approach is not suitable for training a fair classifier that will be used on the unmodified original data. Indeed, as our procedure transport all distribution unto the target one, the resulting dataset is only composed of a single distribution. A classifier trained on the transformed dataset would not be suitable for deploiement where the data are used without first being processed trough our approach, since the classifier has not been trained on other distributions than the target one.}

Our experimental results shows that our approach is able to protect the sensitive attribute and outperforms some approach of the state of the art. The mutual information further enforce the sensitive attribute protection, especially in case of more than two sensitive groups. We have also observed, by comparison with the state of the art, that the protection of the sensitive attribute can be better carried out when some degree of freedom is given to a protected distribution, thus increasing the space of possible representations. In addition, our approach also improves some other fairness metrics used in the literature, as we have demonstrated in our experiments.

Nonetheless, some improvement can be carried out in our approach, especially on the generalisation and the stability of our framework during the cross-validation. A larger exploration of the hyperparameters space would further enhance the quality of our results. 

A future direction of research would consist in the verification of such property, as well as the ability to work on groups that were not seen during the training of models (\emph{i.e.}, \textit{Black-Female}), but can be obtained by combining known sensitive values (\emph{i.e.}, \textit{Black-Male} and \textit{White-Female}).

\section{Relationship between the group Fairness and prevention of inference of the sensitive attribute}
\label{sec:group-fair-priv}
In this section, we discuss the relationship between the group fairness and the prevention of inference of the sensitive attribute.
For the demonstration, let's assume that $A, Y$  are independent to $S$. Thus $P(A=a, Y=y, S=s) = P(A=a, Y=y)P(S=s)$. Furthermore, let's assume that we have access to a predictor $f$, that infer $y$ from $A$ : $\hat{Y} = f(A)$. Since $A, Y$ and $S$ are independent, so will be $f(A)$, since $f$ only uses $A, Y$. The demographic parity states:
\begin{equation}
    \begin{cases}
            DP &= P(\hat{Y}=y / S=s_{i}) - P(\hat{Y}=y / S=s_{j}) < \epsilon \\
            &= P(f(A)=y / S=s_{i}) - P(f(A)=y / S=s_{j}) < \epsilon \\
            &\forall \: \{s_{i}, s_{j}\} \in S\\
            
            EO &= P(\hat{Y}=y / S=s_{i}, Y=y) - P(\hat{Y}=y / S=s_{j}, Y=y)  < \epsilon \\
            &= P(f(A)=y / S=s_{i}, Y=y) - P(f(A)=y / S=s_{j}, Y=y) < \epsilon \\
            &\forall \: \{s_{i}, s_{j}\} \in S
    \end{cases}
    \label{eq:rel-pirv-fair}
\end{equation}
with $y=1$. 

The Independence assumption, allow us to write $P(f(A)=y / S=s_{i}) = P(f(A)=y)$ and $P(f(A)=y / Y=y, S=s_{i}) = P(f(A)=y) / Y=y$. By plug-in it in equation~\ref{eq:rel-pirv-fair} we obtain
\begin{equation*}
    \begin{cases}
        DP &= P(f(A)=y / S=s_{i}) - P(f(A)=y / S=s_{j}) \\
        &=  P(f(A)=y) -  P(f(A)=y) = 0 \\
        EO &= P(f(A)=y / S=s_{i}, Y=y) - P(f(A)=y / S=s_{j}, Y=y) \\
        &= P(f(A)=y) / Y=y - P(f(A)=y) / Y=y) = 0
    \end{cases}
    \label{eq:rel-pirv-fair}
\end{equation*}
Thus, if $A, Y$ and $S$ are independent, metrics such as the accuracy and the equalized odds would be satisfied. 
Fairness preprocessing approaches that works by preventing the inference of the sensitive attribute aim to achieve the independence between $S$ and $A$ (most of these approaches assume that $Y$ is independent to $S$) by removing the correlations between the sensitive attribute and the other variable. As a consequence, they should also improve the demographic parity and the equalized odds.

\section{Transformation of the data decision attribute}
\label{sec:decision-trans}
Throughout this paper, we only considered the case where the decision attribute is not modified with our approach. Nonetheless, it is important to remark that our approach can also transform the decision attribute, further enhancing the overall fairness as the approach would find the most appropriate decision for each transformed individual on the privileged distribution. As a matter of fact, if the protected and the privileged distribution significantly differs in terms of decision rates due to any form of discrimination or favouritism, the approach would apply the same treatment to all datapoints. Thus, leading to the correct decision had the individuals belonging to the privileged group on which data are mapped onto. 
Unfortunately, by transforming the decision attribute, we lose the groundtruth for the decision prediction, as the modified decision is an altered one, that might not correspond to any realistic record.

\section{Optimal \ber{} value for preventing the sensitive attribute inference}
\label{ber-demo}
Suppose that a dataset is composed of $k$ different sensitive groups, with $P(S=s_{x})$ the probability or proportion of each group, and $\hat{S}$ the predicted sensitive attribute. The best protection is achieved when the classifier behaves as a random guess of each class. As such, we can consider the classifier as a random sampling of the sensitive attribute with replacement. Let's consider that the proportion of each class $s_{x}$ obtained through the prediction or sampling of the random classifier predict is $P(\hat{S}=s_x)$.
As the probability of sampling successfully a member of class $s_{x}$ is driven with $P(S=s_{x})$, for each class $s_{x} \: ; \: x \in \{1, \ldots, k\}$, the law of large number state that $P(\hat{S}=s_x)$ will become closer to $P(S=s_{x})$. Meaning that in the long run, the random classifier will predict each class $s_{x}$ with probability $P(S=s_{x})$. 

As we assumed the classifier behaving as a random guessing, the prediction of $S$ is independent of the real value $S$. No information on $\hat{S}$ can be obtained by observing $S$. Thus, $P(\hat{S}=s_x \mid S=s_x) = P(\hat{S}=s_x) = P(S=s_x)$ as previously shown.

As a consequence, $\ber{} = 1 - \dfrac{1}{k}\left(\sum_{i=1}^{k}P(\hat{S}=s_x \mid S=s_{x})\right)$ (cf. equation~\ref{eq:ber}), becomes $\ber{} = 1 - \dfrac{1}{k}\left(\sum_{i=1}^{k}P(S=s_x)\right) = 1 - \dfrac{1}{k}\left(1\right) = 1 - \dfrac{1}{k} = \dfrac{k-1}{k}$

Maximizing the protection with the $\ber{}$ thus correspond to minimizing $\dfrac{k-1}{k} - \ber{} = \dfrac{-1}{k} + \dfrac{1}{k}\left(\sum_{i=1}^{k}P(\hat{S}=s_x \mid S=s_{x})\right)$.

\section{Mutual information computation}
\label{mi-comp}
Given the sensitive attribute $S$ with values $s_i\:;\: i \in \{1, \ldots, k\}$, and its prediction  $\hat{S}$ made by a classifier $f$, the computation of the mutual information between $S$ and $\hat{S}$ requires access to the joint distribution $P(S, \hat{S})$, and the marginals $P(S)$ and $P(\hat{S})$. Since we are dealing with the discrete set $S$, the joint distribution can be computed by counting the number of elements in each set ($S=s_i \cap \hat{S}=s_j$, $i,\: j \: \in \{1, \ldots, k\}$), divided by the number of total elements $\mid S \mid=k$ : $P(S=s_i \cap \hat{S}=s_j) = \dfrac{\mid S=s_i \cap \hat{S}=s_j\mid}{k}$. Similarly, the joint probability can be obtained through $P(S=s_i \cap \hat{S}=s_j) = P(\hat{S}=s_j \mid S=s_i) * P(S=s_i)$

Assume that the predictor $f$ outputs the probability of each datapoint to belong to each group $s_i$. Thus, for each individual $t$ (in a dataset $R$ of $N$ individuals) with data $r_t$, $f$ outputs $\langle p_{t}^{1}, \ldots, p_{t}^{k}\rangle$, with $p_{t}^{j} = f(r_{t})^{j}$ the probability that $r_t$ has the sensitive value of $s_{j}$. The group predicted by $f$ can be obtained by selecting the group with the highest probability ($argmax$). However, the $argmax$ function is not differentiable.

The joint distribution for \fairmapping{} is computed by using the probabilities obtained through $f$ ($p_{t}^{i}$), instead of counting the exact group obtained through $f$. $P(S=s_i \cap \hat{S}=s_j)$ is obtained with equation~\ref{eq:mi-joint-dist}.
\begin{equation}
    \begin{split}
        P(S=s_i \cap \hat{S}=s_j) &= P(\hat{S}=s_j \mid S=s_i) * P(S=s_i) \\
        &= \dfrac{\sum_{r_{t} \in R/S=s_i}f(r_{t})^{j}}{\mid r_{t} \in R/s_i \mid} * P(S=s_i)
    \end{split}
\label{eq:mi-joint-dist}
\end{equation},
with $r_{t} \in R/S=s_i$ the set of datapoints in the dataset with the sensitive attribute $s_i$, $f(r_{t})^{j}$ the probability that $r_{t}$ has the sensitive attribute $s_{j}$ as given with $f$, and $P(S=s_i)$ the proportion of the data with sensitive value $s_i$.
The marginal $P(\hat{S}=s_j)$ can be obtained with a similar procedure, by averaging over the predicted probabilities of $f$ (equation~\ref{eq:mi-marg}).
\begin{equation}
    \begin{split}
        P(\hat{S}=s_j) = \dfrac{\sum_{t=1}^{N}f(r_{t})^{j}}{N}
    \end{split}
\label{eq:mi-marg}
\end{equation}

\section{Hyperparameters for external classifiers}
\label{sec:external-clfs}

In Table~\ref{tab:exthprms} we present the external classifiers and their respective hyperparameters. 
\begin{table}[h!]
    \centering
    \caption{Hyperparameters used for all external classifiers. Parameters not mentioned are used with default values}
    \label{tab:exthprms}
    \resizebox{.9\linewidth}{!}{
        \begin{tabular}{|c|c|c|}
            \hline
            Classifier & \multicolumn{2}{|c|}{Parameters} \\ \hline

            \multirow{3}{*}{Gradient Boosting Classifier (GBC)} & \textit{Estimators}    & $100$              \\
            & \textit{Learning rate} & $0.1$              \\
            & \textit{Maximum depth} & $3$                \\\hline 



            \multirow{4}{*}{Multi-layer Perceptron Classifier (MLPC)} & 
            \multicolumn{2}{|c|}{\textit{Default parameters}} \\
            & \textit{hidden layer sizes} & $100$ \\
            & \textit{activation} & $relu$  \\
            & \textit{solver} & $adam$ \\\hline
            
            \multirow{4}{*}{Decision Tree (DTC)} & 
            \multicolumn{2}{|c|}{\textit{Default parameters}} \\
            & \textit{criterion} & $gini$ \\
            & \textit{splitter} & $best$  \\
            & \textit{max depth} & $None$ \\\hline
            
        \end{tabular}
    }
\end{table}

\section{Experiment results with original privileged data}
\label{sec:results-orig-priv}
In this section, we present the results of all approaches investigated in this paper when using the original data from the privileged group. These results are, for most approaches, obtained by replacing the reconstructed values of the privileged group by their original counterpart. We refer the reader to the Table~\ref{tab:datasets-composition} for the optimal protection metrics values. 

\subsection{Lipton results}
\begin{figure}[ht]
        \includegraphics[width=1\linewidth, height=0.45\textheight]{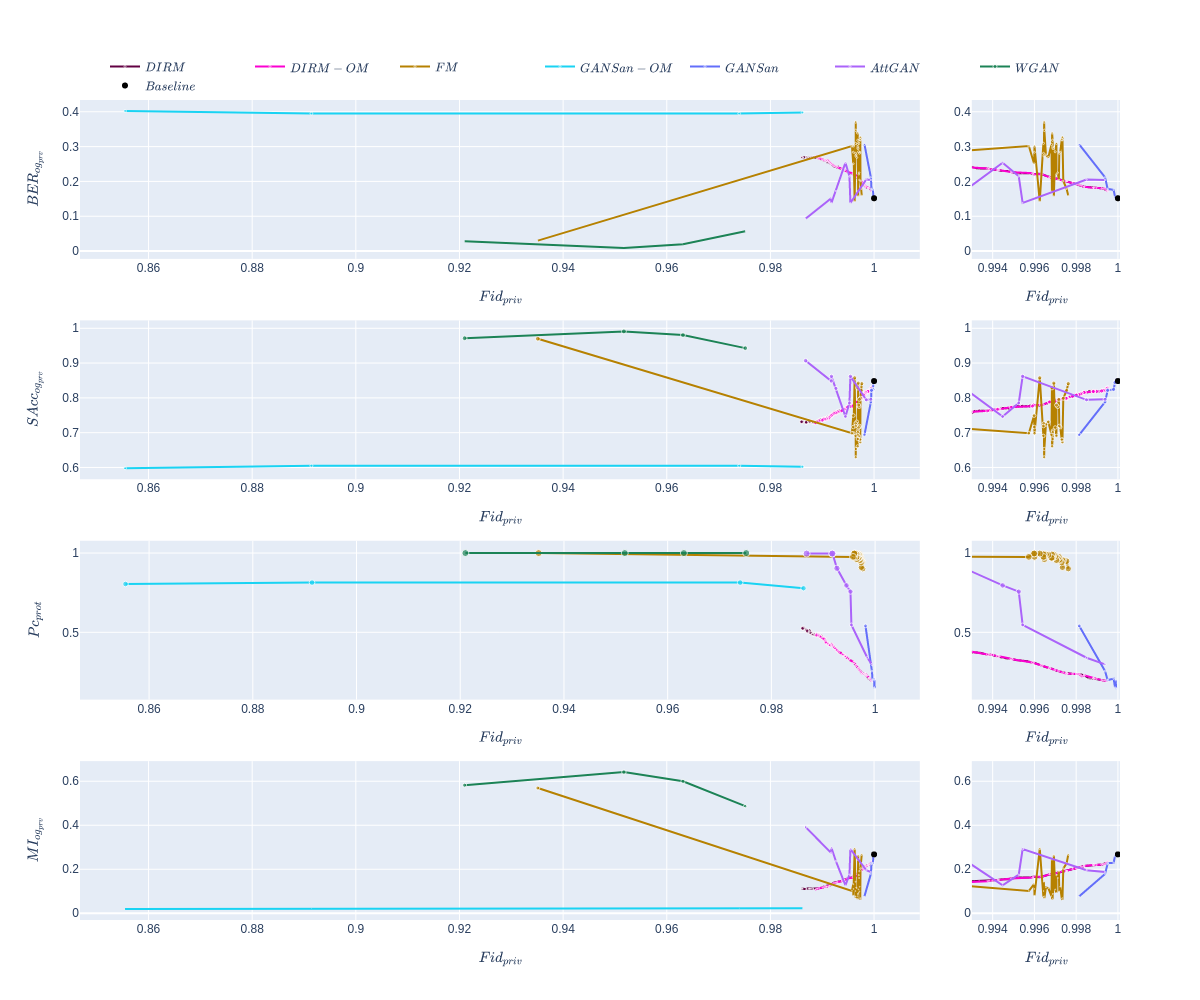}
    \caption{Pareto fronts in the \textit{Fair Mapping} perspective for approaches investigated on Lipton. The right column present all points on the fronts, while the right columns present the same results but on the range [$0.99$ - $1$], for a better visualization. Top to bottom: protection $\ber{}_{og_{prv}}$, protection $\sac_{og_{prv}}$, accuracy of $S$, $Pc_{prot}$, mutual information $MI_{og_{prv}}$.
    }
    \label{fig:results-front-lipton-org}
\end{figure}
On the Lipton dataset (Figure~\ref{fig:results-front-lipton-org}), we can observe that all approaches exhibit a similar behaviour as observed with the reconstructed privileged data. However, the protection achieved by all approaches are lower than the one achieved with the reconstructed privileged group. This observation is particularly true for $\gansan{}-OM$, where the model displays the highest protection among all approaches, even though the fidelity of the privileged group is lower. 
\wgan{} transforms the data such that it resemble the privileged group, but still fail to protect the sensitive attribute.

\subsection{German Credit results}
\begin{figure}[ht]
        \includegraphics[width=1\linewidth, height=0.45\textheight]{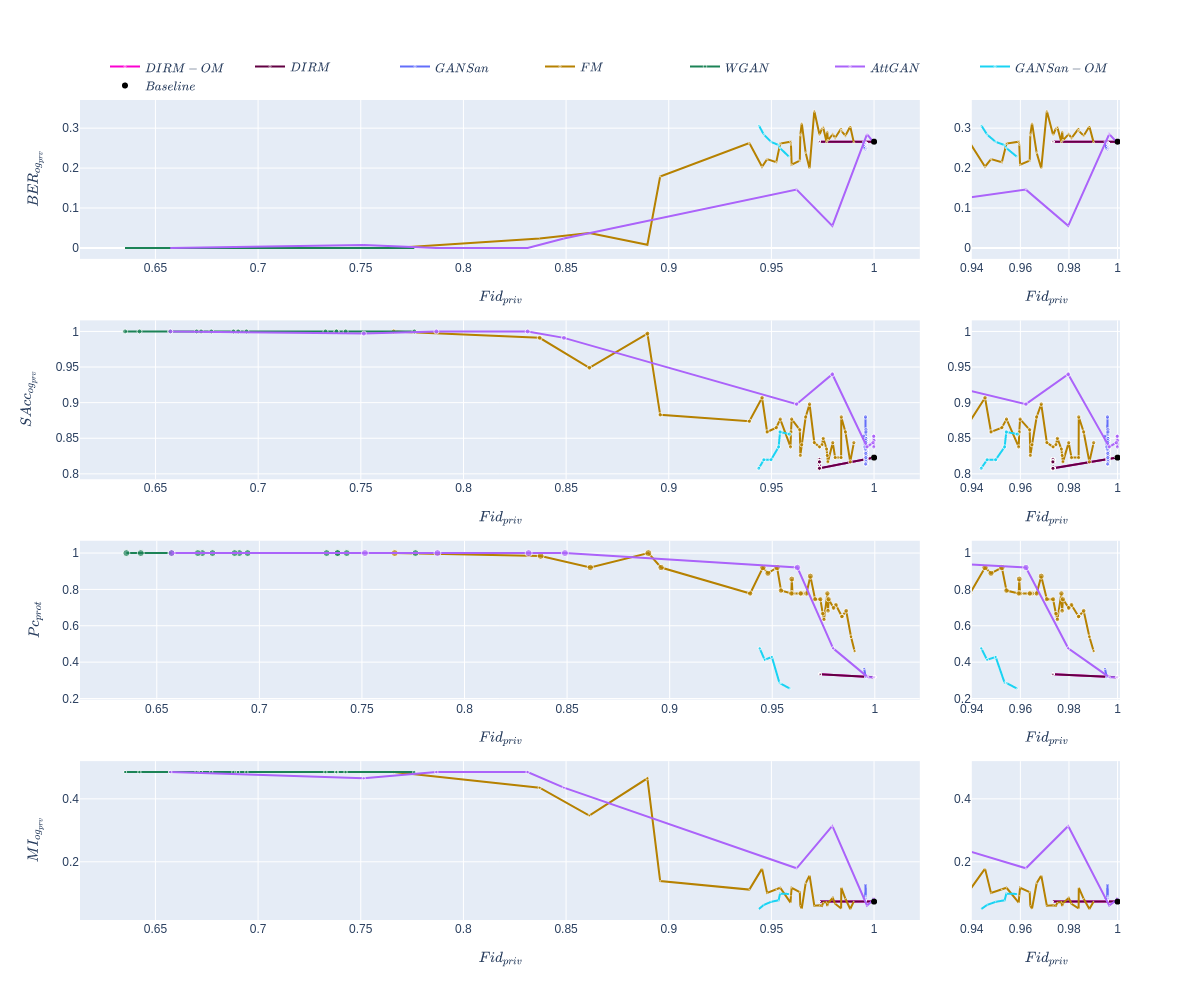}
    \caption{Pareto fronts in the \textit{Fair Mapping} perspective for approaches investigated on German. The right column present all points on the fronts, while the right columns present the same results but on the range [$0.94$ - $1$], for a better visualisation.
    }
    \label{fig:results-front-german-org}
\end{figure}

On German credit (Figure~\ref{fig:results-front-german-org}), the use of the original privileged data only slightly improves the protection of the sensitive attribute. Only $\fairmapping{}$ and $\gansan{}-OM$ slightly improves the baseline result. All other approaches do not. the transformation is largely dominated by $\attgan{}$, closely followed by $\fairmapping{}$. Unfortunately, $\dirm{}-OM$ neither improves the protection nor the transformation ($Pc_{prot}$).

\subsection{Adult Census results}
\begin{figure}[ht]
        \includegraphics[width=1\linewidth, height=0.2\textheight]{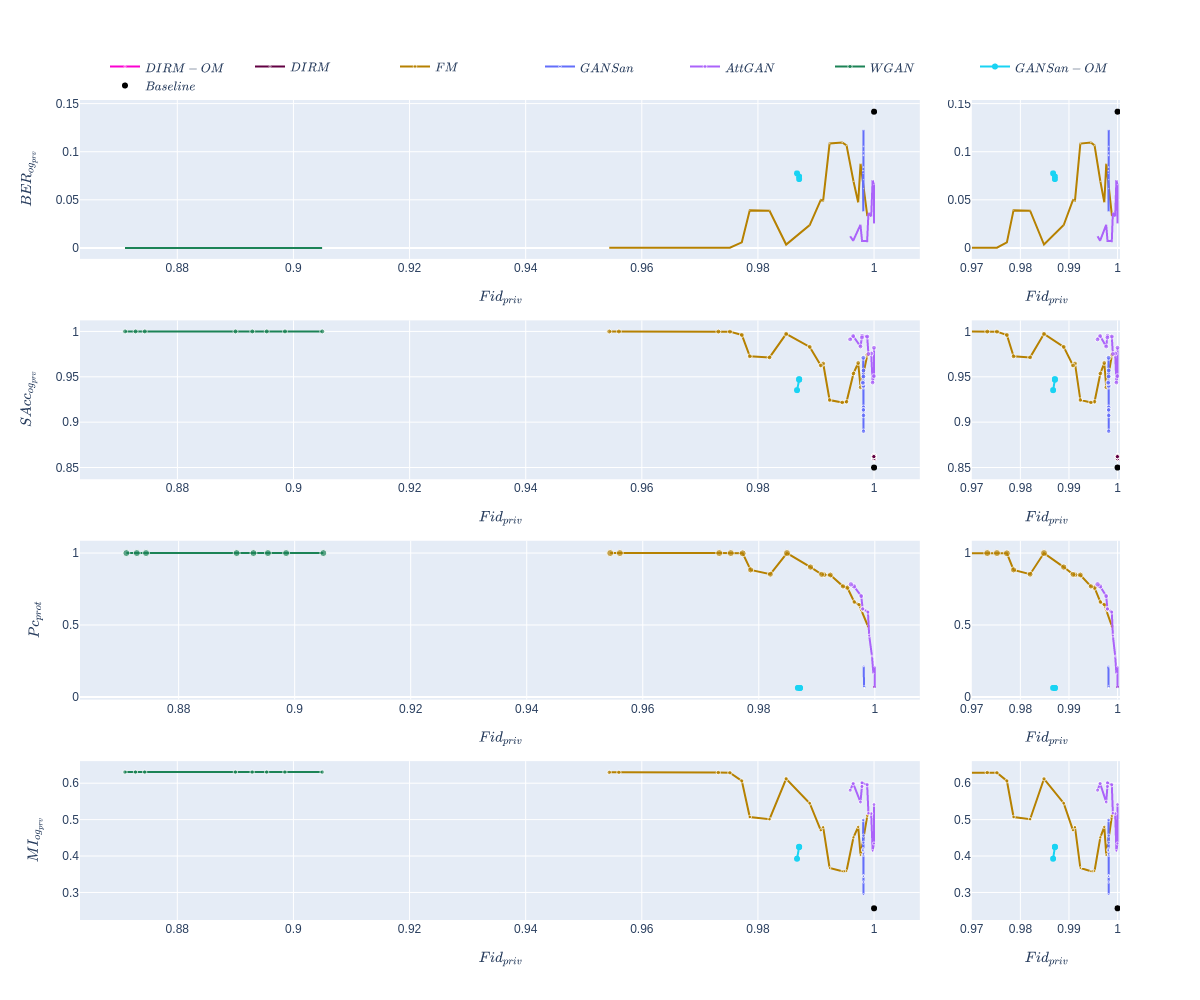}
    \caption{Pareto fronts in the \textit{Fair Mapping} perspective for approaches investigated on Adult census. The right column present all points on the fronts, while the right columns present the same results but on the range [$0.97$ - $1$], for a better visualisation.
    }
    \label{fig:results-front-adult-org}
\end{figure}

On Adult Census (Figure~\ref{fig:results-front-adult-org}), the protection of the sensitive attribute with the original data from the privileged group is nearly impossible for all approaches. No approaches successfully provide a protection better than the baseline as measured with either the $\ber{}$ or the $\sac{}$, even though some approaches such as $\wgan{}$ or $\fairmapping{}$ can successfully maximize the transformation metric $Pc_{prot}$. This suggests that a minimum level of modification in the privileged group is necessary to protect the sensitive attribute. 

\subsection{Adult2 Census with results}
\label{sec:adult2-org-results}
\begin{figure}[ht]
        \includegraphics[width=1\linewidth, height=0.45\textheight]{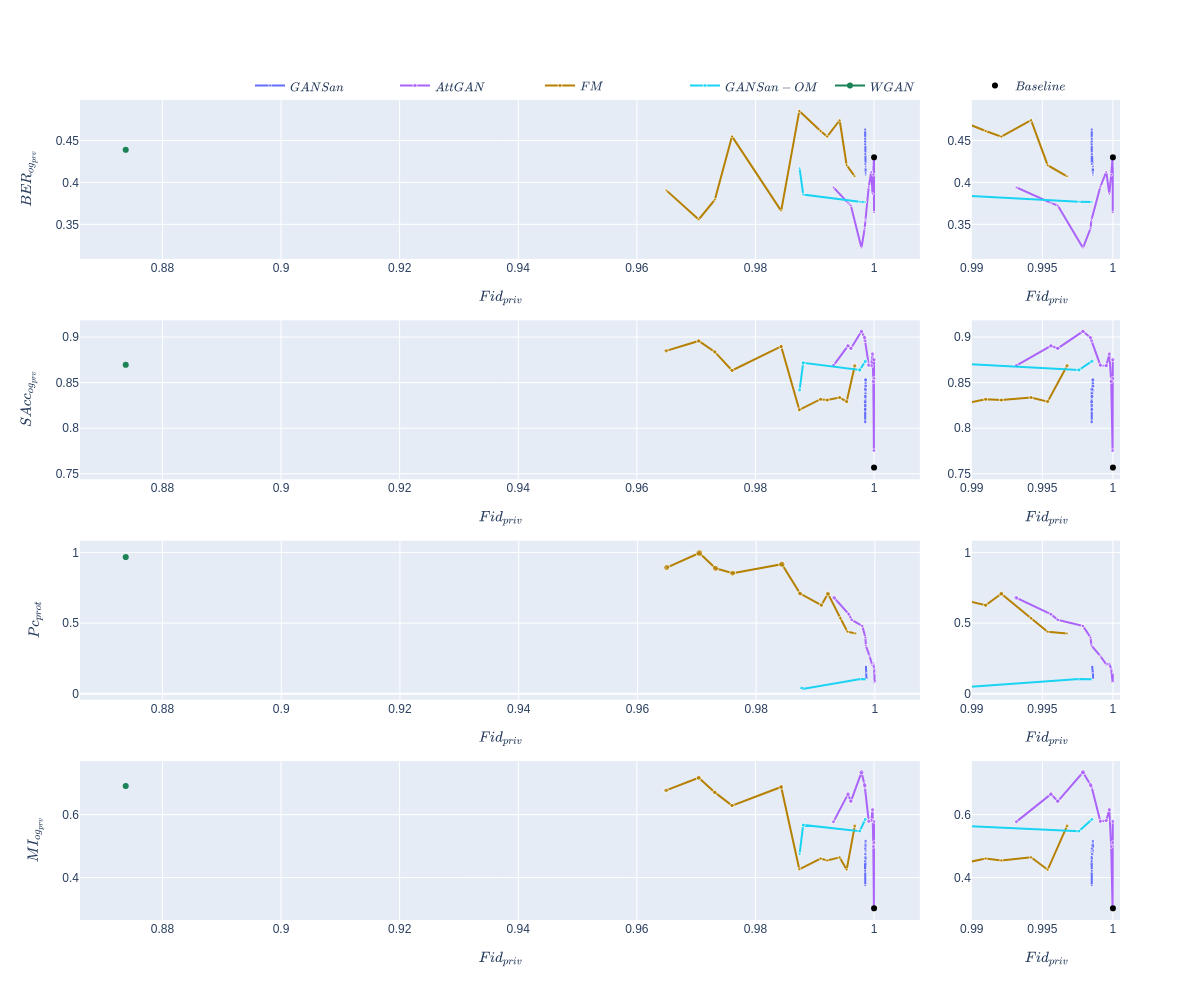}
    \caption{Pareto fronts in the \textit{Fair Mapping} perspective for approaches investigated on Adult census with 2 attributes. The right column present all points on the fronts, while the right columns present the same results but on the range [$0.99$ - $1$], for a better visualization.
    }
    \label{fig:results-front-adult2-org}
\end{figure}

Just as observed with Adult on a single binary attribute, we can observe (Figure~\ref{fig:results-front-adult2-org}) that the protection of two attributes without the modification of the privileged group is highly difficult. Even though we observe that we can achieve a higher \ber{} value, the accuracy is still higher than the baseline computed with the original data. The mutual information is still significantly higher than measured on the original data. No approaches provide an appropriate level of protection. This observation suggest that in case of multiple attribute, the $\ber{}$ can not be the used as the only protection measure, and certain amount of modification of the privileged data is required in order to improve the data protection.

\subsection{\fairmapping{} instantiations}
\begin{figure}[ht]
        \includegraphics[width=1\linewidth, height=0.3\textheight]{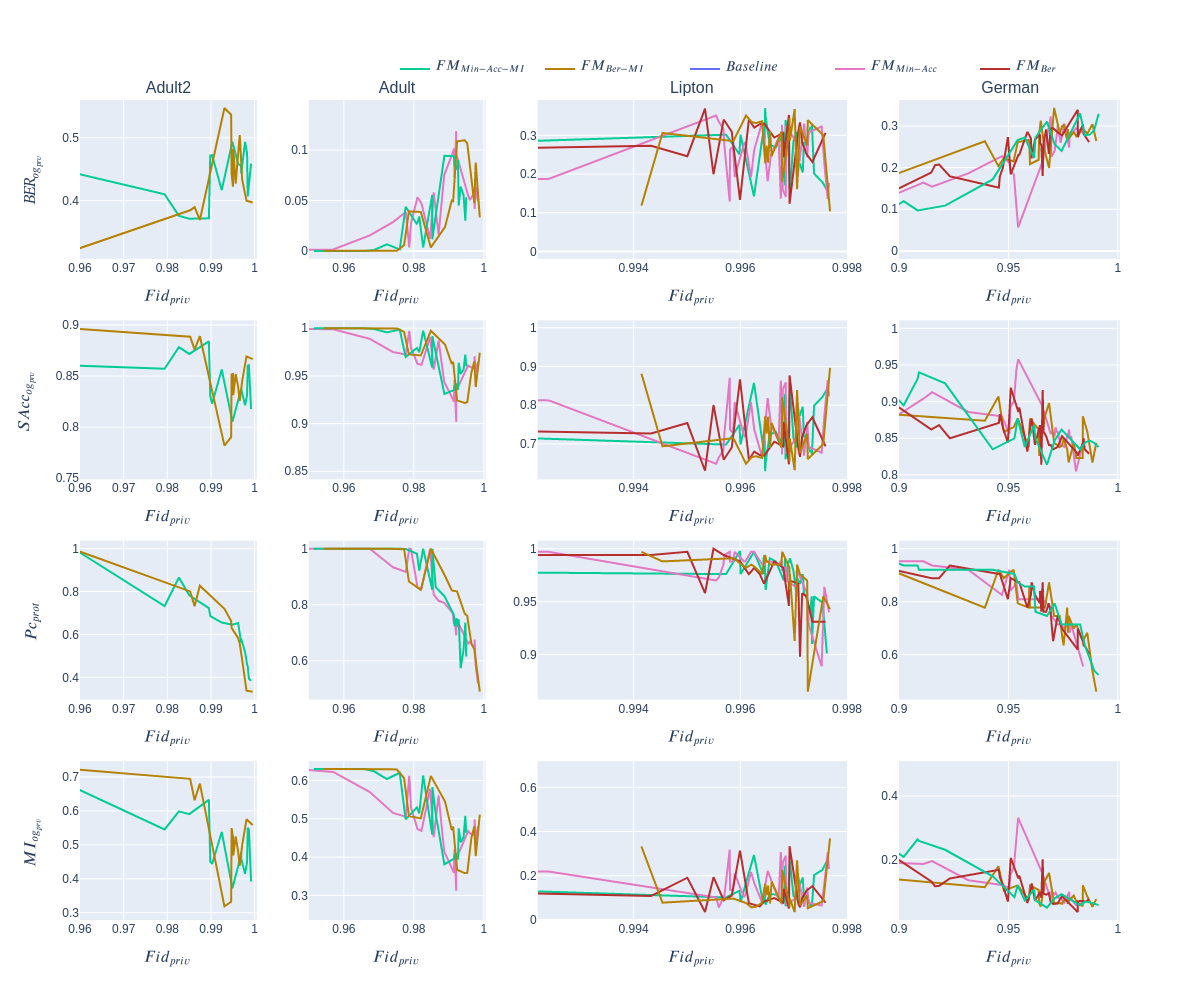}
    \caption{Different \textit{Fair Mapping} instances. We can observe that the different $\fairmapping{}$ instances behave almost similarly for all chosen loss function for protecting the sensitive attribute. Each column represent the performances on a single dataset. On Adult2, we always include the mutual information regularisation.
    }
    \label{fig:results-all-fm-org}
\end{figure}

We present in Figure~\ref{fig:results-all-fm-org} the different instances of \fairmapping{}. We can observe that all of those approaches have an almost similar range of values. The figure also demonstrates the stability of our approach.

\section{Divergences with original data}
\label{sec:div-org}
\begin{figure}[ht]
        \includegraphics[width=1\linewidth, height=0.2\textheight]{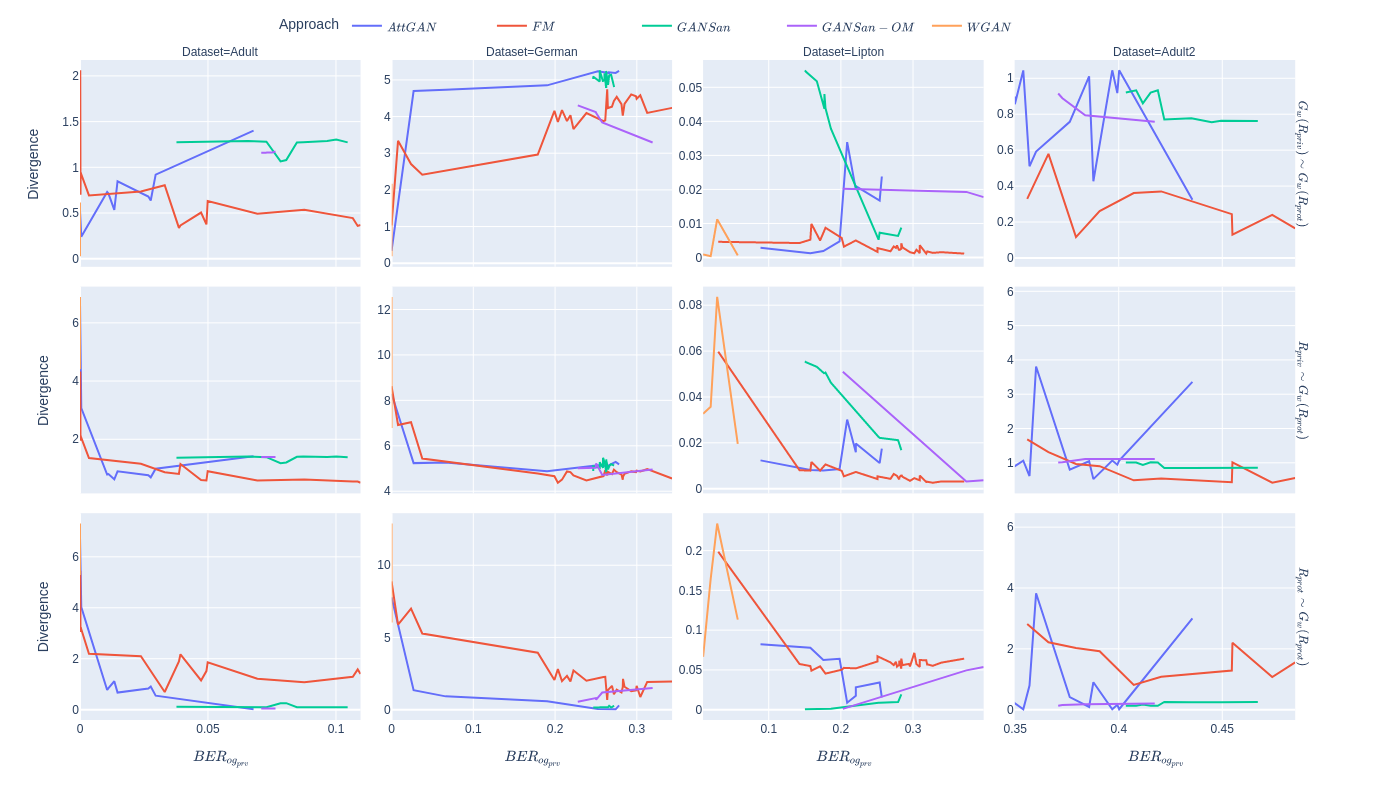}
    \caption{Divergences computed on German, Adult, Adult2 and Lipton. Each column represent a dataset while each row represent the divergence between the transformed protected data and respectively the original protected data ($R_{prot} \sim G_{w}(R_{prot})$), the reconstructed privileged group ($G_{w}(R_{priv}) \sim G_{w}(R_{prot})$), and the original privileged group ($R_{priv} \sim G_{w}(R_{prot})$). The divergences are represented with respect to the protection $\ber{}_{og_{prv}}$.
    }
    \label{fig:divergence-ber-o-m}
\end{figure}

\begin{figure}[ht]
        \includegraphics[width=1\linewidth, height=0.2\textheight]{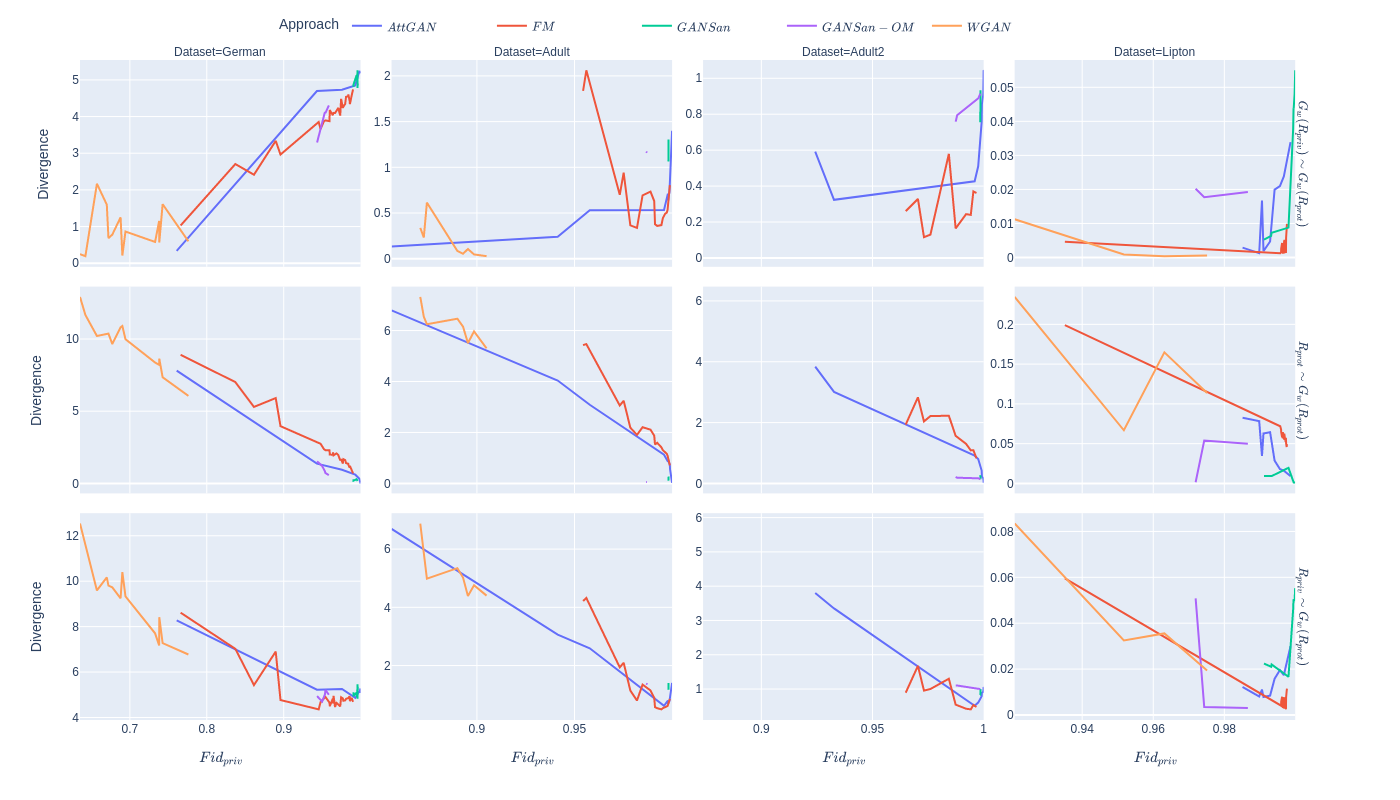}
    \caption{Divergences computed on German, Adult, Adult2 and Lipton. Each column represent a dataset while each row represent the divergence between the transformed protected data and respectively the original protected data ($R_{prot} \sim G_{w}(R_{prot})$), the reconstructed privileged group ($G_{w}(R_{priv}) \sim G_{w}(R_{prot})$), and the original privileged group ($R_{priv} \sim G_{w}(R_{prot})$). The divergences are represented with respect to the fidelity $\fid{}_{priv}$.
    }
    \label{fig:divergence-fid-prv}
\end{figure}

In Figure~\ref{fig:divergence-ber-o-m} we present the \textit{Sinkhorn} divergences results with respect to the protection computed with the original privileged group $\ber{}_{og_{prv}}$, while Figure~\ref{fig:divergence-fid-prv} present the same results with respect to the fidelity of the privileged data.

\section{AutoEncoder fidelity and protection}
\label{sec:rec-vs-protection}
In this section, we discuss the difference of results between the protection obtained with the original privileged data (presented in section~\ref{sec:experiments}) and the reconstructed privileged data (section~\ref{sec:results-orig-priv}). From those results, we can observe that a minimum amount of modification of the privileged data is required in order to enhance the sensitive attribute protection, while any method working the original privileged data cannot improve the results significantly above the baseline. In order to evaluate whether such important difference can be explained with other factors than our training procedure, we trained an AutoEncoder to reconstruct the privileged group data. For different fidelity values, we trained a classifier to distinguish the reconstructed privileged group from their original values. The underlying idea is that if the trained classifier are always able to distinguish the reconstructed data from their original version, the AutoEncoder introduce some form of watermark within the reconstructed data that render it always distinguishable from their original values. As such, we cannot expect the transforming model to be able to transport the protected distribution unto the privileged one without such watermark. The ideal values to achieve are an accuracy of $0.5$ and a $\ber{}$ of $0.5$ for fidelity values different from $1$.

\begin{figure}[ht]
        \includegraphics[width=1\linewidth, height=0.2\textheight]{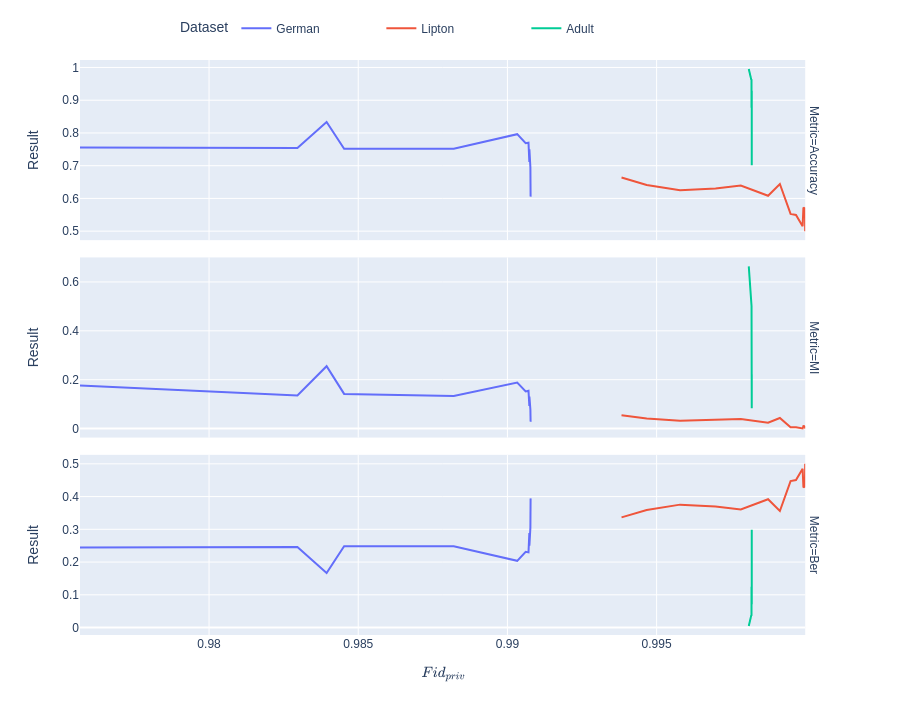}
    \caption{Accuracy, BER and MI of classifiers trained to differentiate the original privileged group from their reconstructed version with a vanilla (no other constraint except the reconstruction) AutoEncoder.
    }
    \label{fig:ae_fid_vs_prot}
\end{figure}

The results are presented in Figure~\ref{fig:ae_fid_vs_prot}. We can observe that for dataset Lipton and German, unless the fidelity obtained is well above $0.999489$ and $0.9907$ respectively, the trained classifier have a $\ber{}$ lower than $0.4$ and an accuracy greater than $0.64$ and $0.77$ respectively. In our results, achieving such fidelity values is highly difficult for any of the chosen approaches. On Lipton, the accuracy and the $\ber{}$ are well below $0.66$ and above $0.36$ respectively, for any fidelity values. This suggest that our transformation mechanism could transform the protected group into the privileged data such that protection measured reach similar values, which, unfortunately, is not exactly the case. 
On Adult on the other hand, the lowest accuracy is $0.7559$ and the highest $\ber$ is $0.2440$. The lowest fidelity is $0.9979$ and the highest is $0.9981$. In our results, the highest fidelity with our approach are $0.9967$ and $0.9988$ on Adult2 and Adult. As a consequence, we cannot expect the transformation to produce data indistinguishable from the original privileged distribution.


\section{Results on other approaches perspective}
\label{sec:res-other-perspective}
In this section, we will observe the performances of different approaches through different perspectives. 

\subsection{Lipton results}
\begin{figure}[ht]
        \includegraphics[width=1\linewidth, height=0.35\textheight]{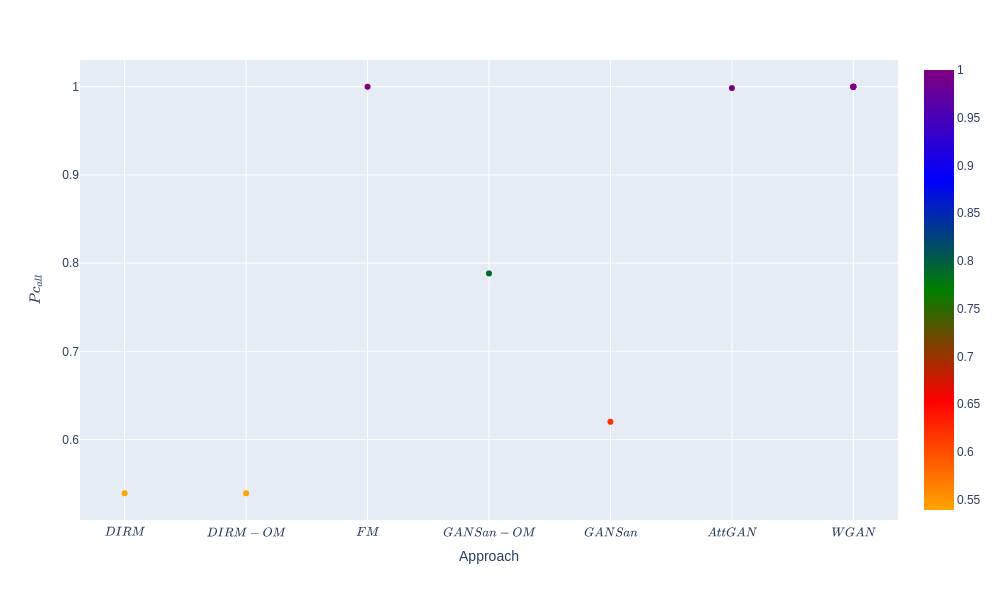}
    \caption{Results on the perspective of \wgan{} (lipton dataset), which is the transformation of data such that they resemble the privileged group. 
    }
    \label{fig:results-front-lipton-org-wgan}
\end{figure}

In the \wgan{} perspective (Figure~\ref{fig:results-front-lipton-org-wgan}) which is the transformation of data from the protected group, we can observe that nearly all approaches that have the privileged group as a chosen target distribution achieve a transformation performance above or nearly equal to $0.8$. The exception is with $\dirm{}-OM$. As mentioned in section~\ref{sec:experiments}, this result might be a consequence of the underlying repair scheme (procedure) or procedure of the $\dirm{}$ approach.

\begin{figure}[ht]
        \includegraphics[width=1\linewidth, height=0.35\textheight]{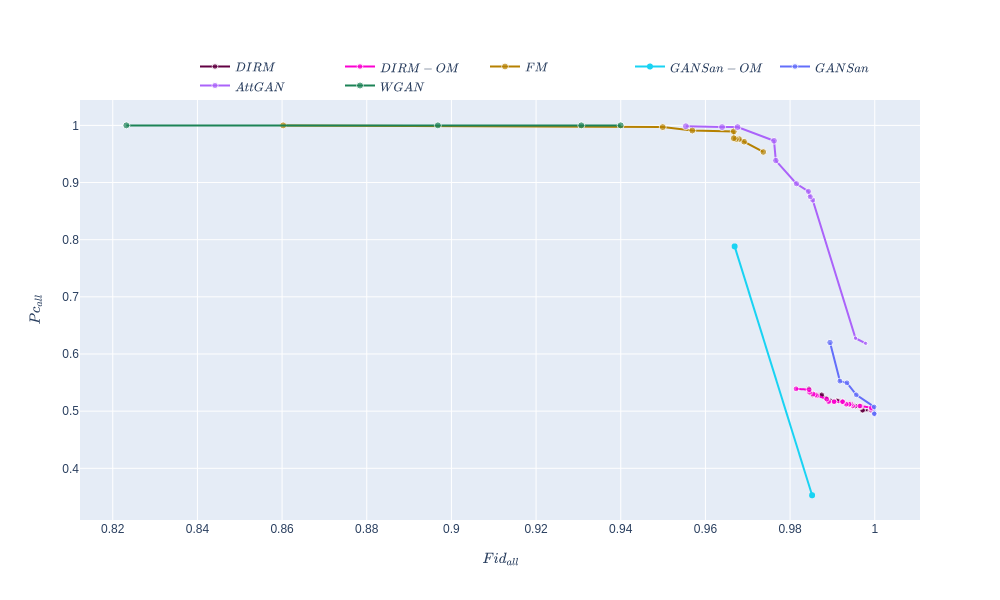}
    \caption{Results on the perspective of \attgan{}, (Lipton dataset)which is the transformation of data such that they resemble the privileged group. 
    }
    \label{fig:results-front-lipton-org-attgan}
\end{figure}

On the \attgan{} perspective (Figure~\ref{fig:results-front-lipton-org-attgan}), as we expect from $\gansan{}$ and $\dirm{}$ the transformation results are the lowest. $\attgan{}$ dominates for higher fidelities values, but are equal to $\fairmapping{}$ and $\wgan{}$ for lower values.

\begin{figure}[ht]
        \includegraphics[width=1\linewidth, height=0.45\textheight]{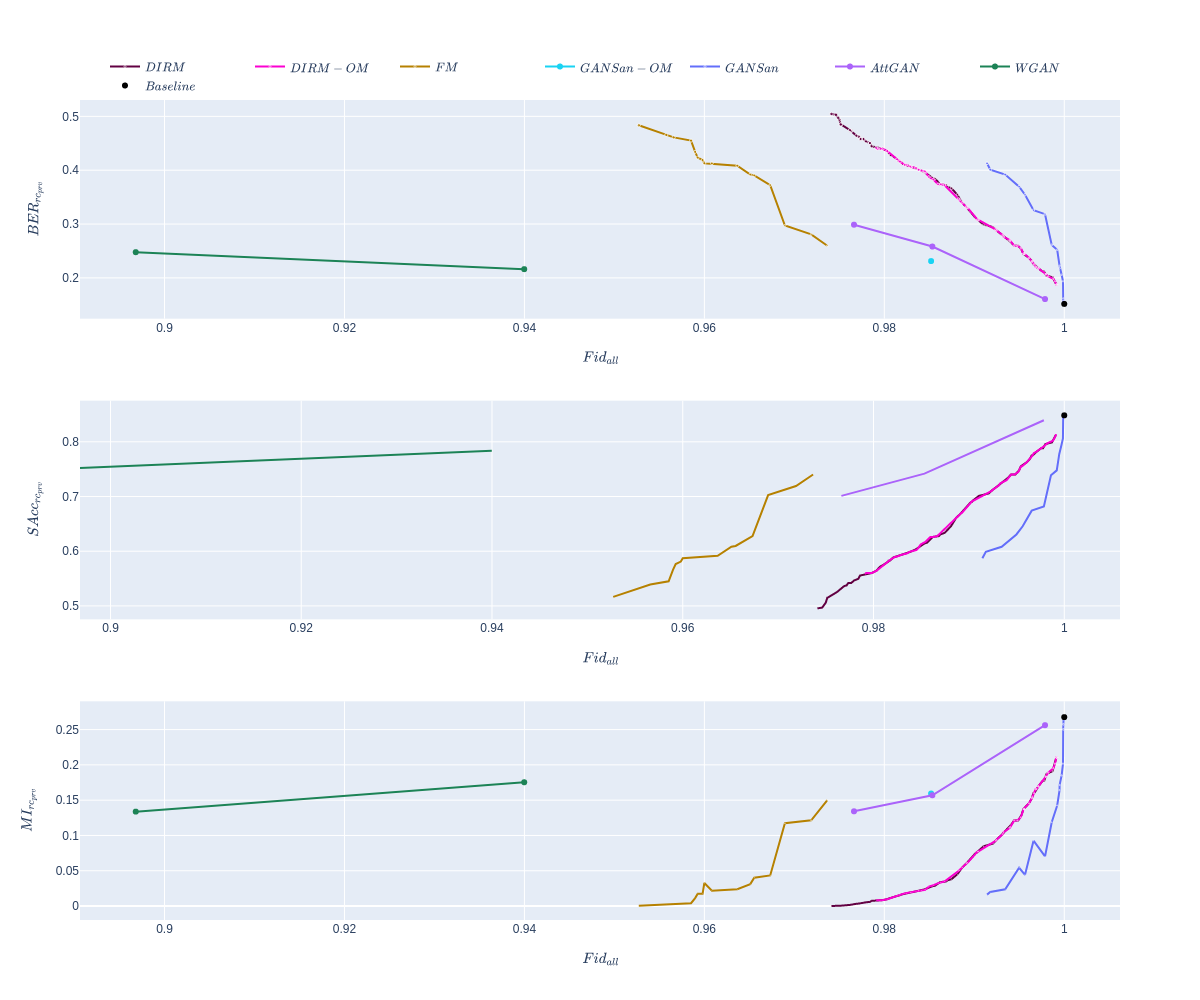}
    \caption{Results on the perspective of \gansan{} and \dirm{} (lipton dataset), which is the transformation of data such that they resemble the privileged group. 
    }
    \label{fig:results-front-lipton-org-gansan}
\end{figure}
\gansan{} and \dirm{} shares the same perspective (Figure~\ref{fig:results-front-lipton-org-gansan}). We can observe that \gansan{} provides the best trade-offs between protection and data reconstruction. $\fairmapping{}$ has a lower fidelity as we could expect, since there are no reconstruction constraint on the protected group, which we only want to render similar to the privileged group distribution. Nonetheless, it is interesting to observe that by only constraining the model to reconstruct the privileged group, it has enough information to significantly reduce the error in the  reconstruction of the protected data.

\subsection{German Credit results}

\begin{figure}[ht]
        \includegraphics[width=1\linewidth, height=0.35\textheight]{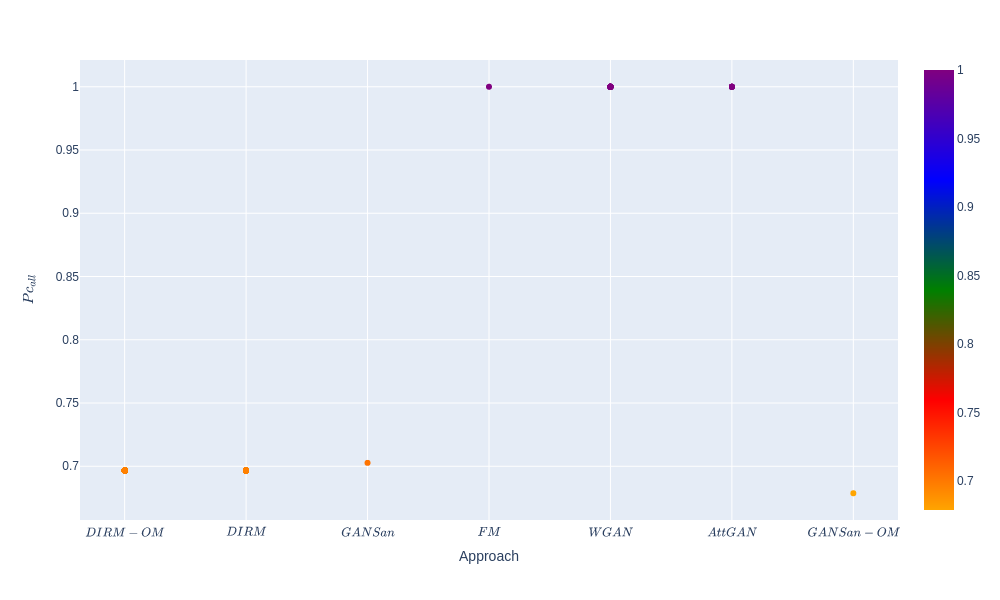}
    \caption{Results on the perspective of \wgan{} (German dataset), which is the transformation of data such that they resemble the privileged group. 
    }
    \label{fig:results-front-german-org-wgan}
\end{figure}

Just as observed with the Lipton dataset, the best results in the \wgan{} perspective (Figure~\ref{fig:results-front-german-org-wgan}) are achieved with $\wgan{}$, $\fairmapping{}$ and $\attgan{}$. All other approaches have performances close to $0.7$. Surprisingly, $\gansan{}$ outperforms $\dirm{}-OM$ and $\gansan{}-OM$ on the transformation metric.

\begin{figure}[ht]
        \includegraphics[width=1\linewidth, height=0.35\textheight]{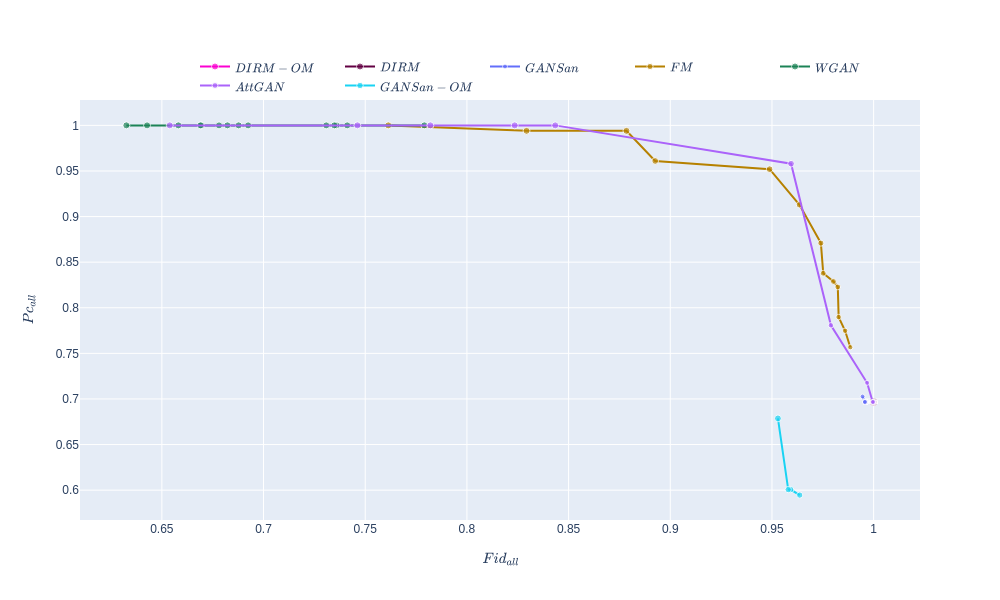}
    \caption{Results on the perspective of \attgan{} (German dataset), which is the transformation of data such that they resemble the privileged group. 
    }
    \label{fig:results-front-german-org-attgan}
\end{figure}

On the \attgan{} perspective (Figure~\ref{fig:results-front-german-org-attgan})
\attgan{} and \fairmapping{} are the most dominant approaches. 

\begin{figure}[ht]
        \includegraphics[width=1\linewidth, height=0.45\textheight]{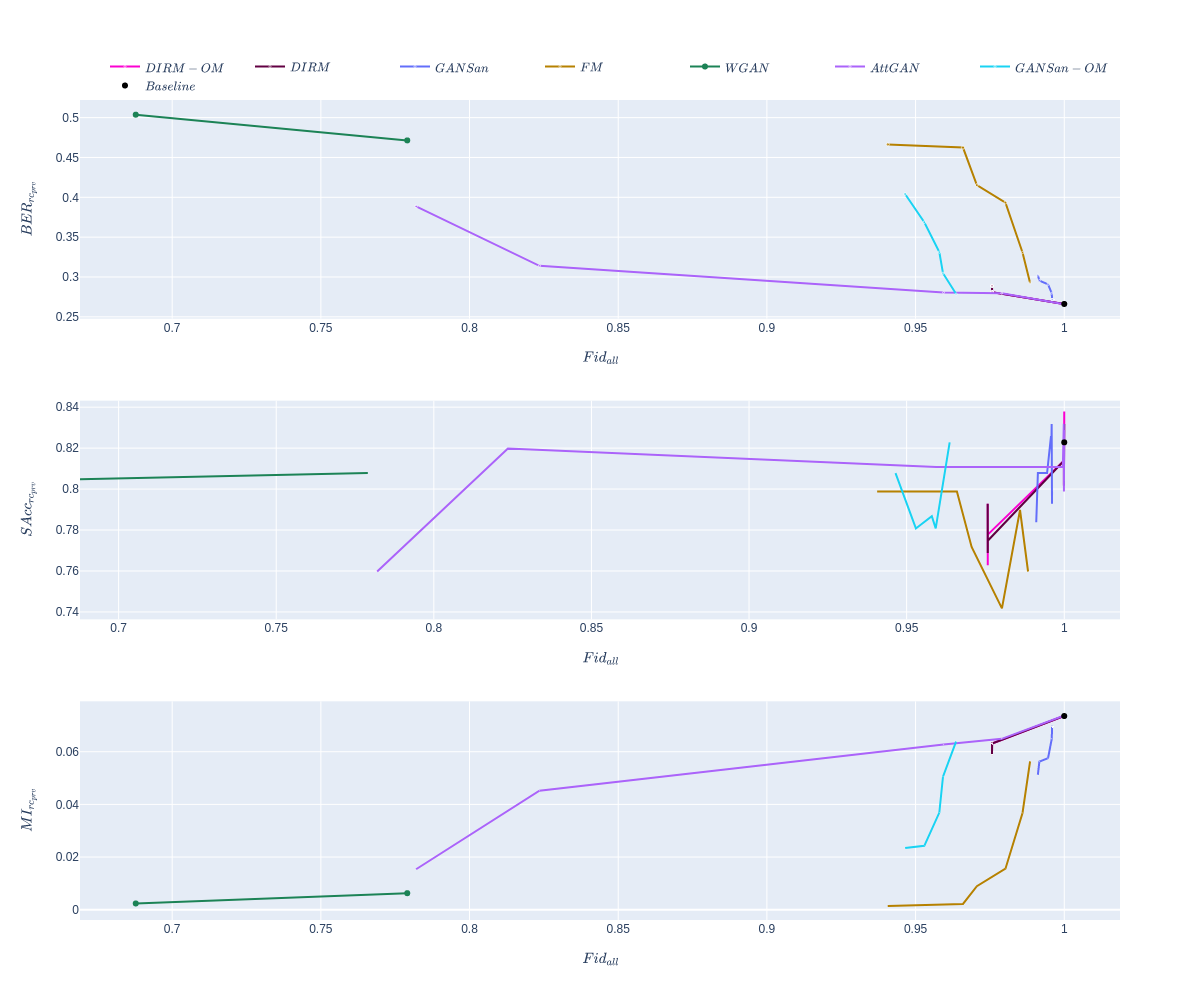}
    \caption{Results on the perspective of \gansan{} (German dataset), which is the transformation of data such that they resemble the privileged group. 
    }
    \label{fig:results-front-german-org-gansan}
\end{figure}
\fairmapping{} dominates all approaches on the \gansan{} and \dirm{} perspective. \gansan{} slightly improves the fidelity with a little enhancement of the protection, while \dirm{} produces results close to the baseline. Even if the protection is slightly enhanced with \gansan{}, we can observe that the mutual information is still high. A notable observation is that \gansan{} and \fairmapping{} completes each other in this perspective. In fact, where \fairmapping{} fall short in terms of fidelity on all the dataset datapoints, \gansan{} slightly improves such fidelity. Both Pareto-fronts nearly forms a single continuous one.

\subsection{Adult Census results}
\begin{figure}[ht]
        \includegraphics[width=1\linewidth, height=0.35\textheight]{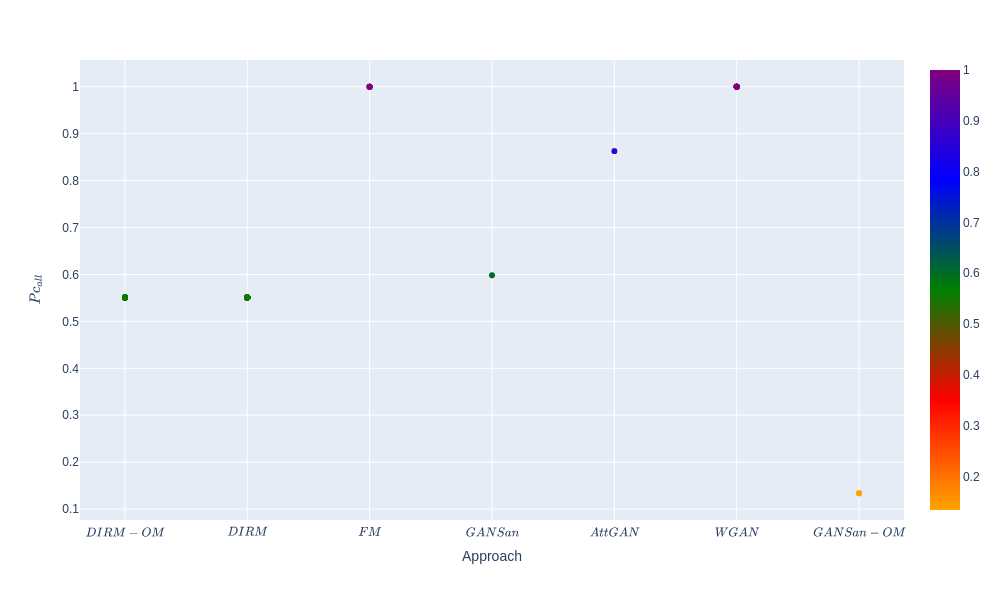}
    \caption{Results on the perspective of \wgan{} (Adult dataset), which is the transformation of data such that they resemble the privileged group. 
    }
    \label{fig:results-front-adult-org-wgan}
\end{figure}
On the \wgan{} perspective (Figure~\ref{fig:results-front-adult-org-wgan}), as observed on any other dataset, $\fairmapping{}$ and $\wgan{}$ are maximizing the transformation metric. Surprisingly, \attgan{} does not have a performance better than $90\%$, even though it also has the objective of maximizing the transformation metric.

\begin{figure}[ht]
        \includegraphics[width=1\linewidth, height=0.35\textheight]{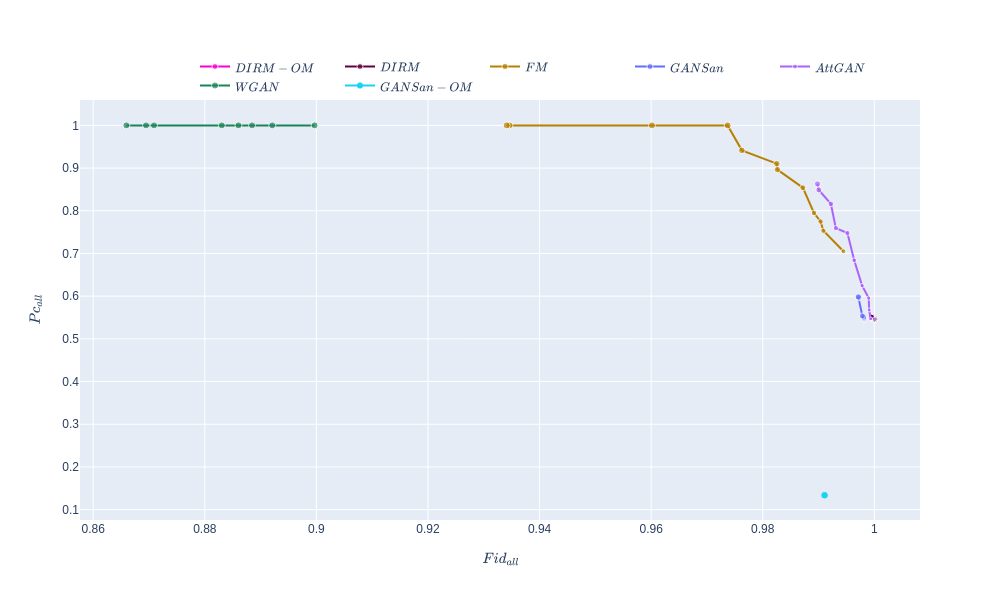}
    \caption{Results on the perspective of \attgan{} (Adult dataset), which is the transformation of data such that they resemble the privileged group. 
    }
    \label{fig:results-front-adult-org-attgan}
\end{figure}

In the perspective of \attgan{} in Figure~\ref{fig:results-front-adult-org-attgan}, \fairmapping{} and \attgan{} behave almost in a complementary fashion. \attgan{} slightly dominates \fairmapping{} on higher fidelity values. Surprisingly, \dirm{} and \dirm{}-OM dominates \gansan{}-OM.

\begin{figure}[ht]
        \includegraphics[width=1\linewidth, height=0.45\textheight]{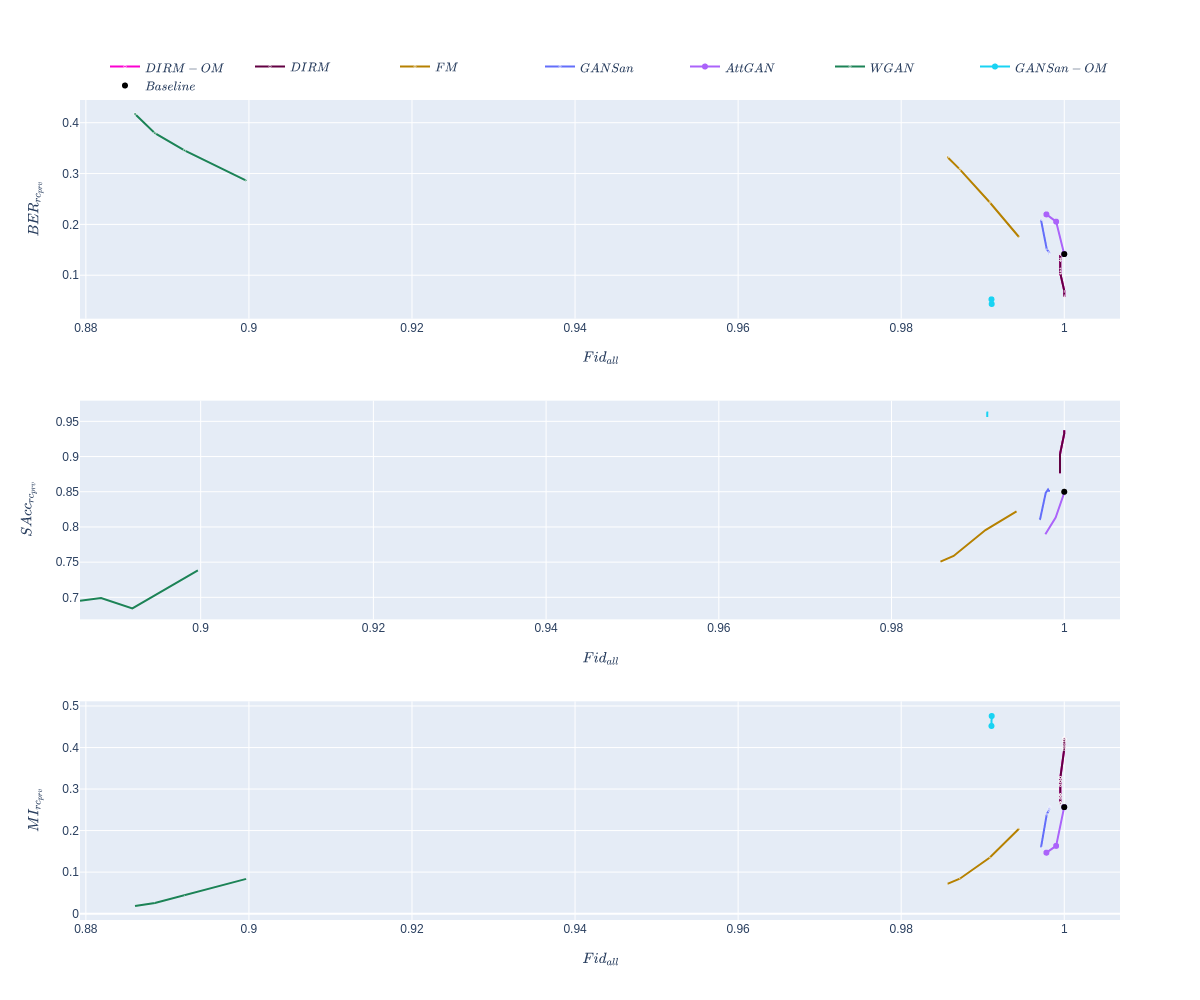}
    \caption{Results on the perspective of \gansan{} (Adult dataset), which is the transformation of data such that they resemble the privileged group. 
    }
    \label{fig:results-front-adult-org-gansan}
\end{figure}
On the \gansan{} perspective (Figure~\ref{fig:results-front-adult-org-gansan}), it is interesting to see that \fairmapping{} and \attgan{} still behave in a complementary fashion. \fairmapping{} achieves the highest protection and fidelity trade offs, \attgan{} provides the higher fidelity while $\wgan{}$ has the highest protection. 

\subsection{Adult2 Census with results}
\begin{figure}[ht]
        \includegraphics[width=1\linewidth, height=0.35\textheight]{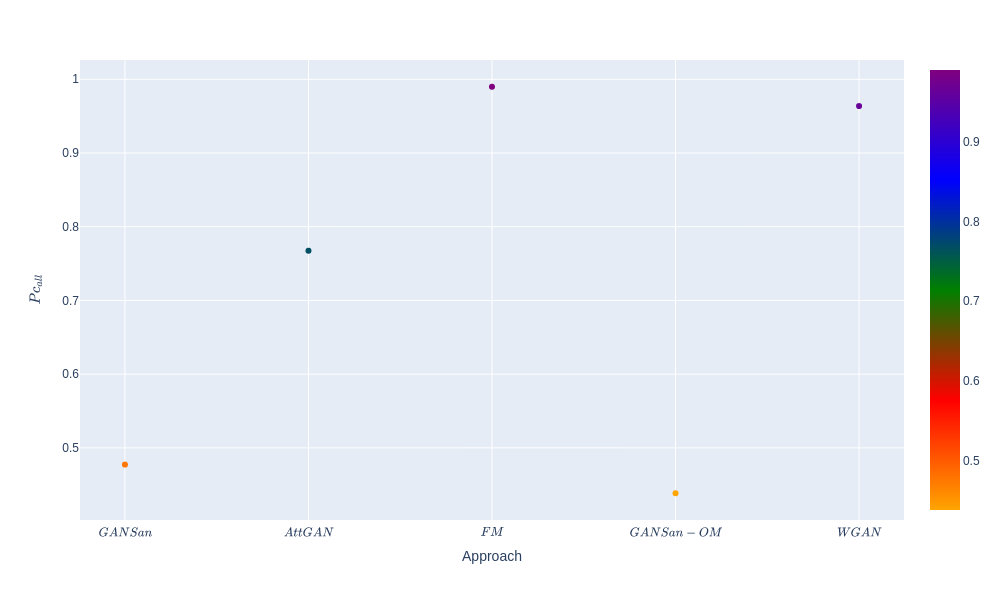}
    \caption{Results on the perspective of \wgan{} (Adult2 dataset), which is the transformation of data such that they resemble the privileged group. 
    }
    \label{fig:results-front-adult2-org-wgan}
\end{figure}
On the \wgan{} perspective (Figure~\ref{fig:results-front-adult2-org-wgan}), we can observe that among all the trade-offs achievable on the \fairmapping{} Pareto-front, there is at least one trade-off for which the highest transformation $\classification{}_{prot}$ is almost equal to $1$. For \gansan{} and \gansan{}-OM, the highest classification $\classification{}_{prot}$ among all trade-offs is less than $0.5$, while it is close to $0.97$ for $\wgan{}$. $\attgan{}$ can achieve the perfect score of $1$.

\begin{figure}[ht]
        \includegraphics[width=1\linewidth, height=0.35\textheight]{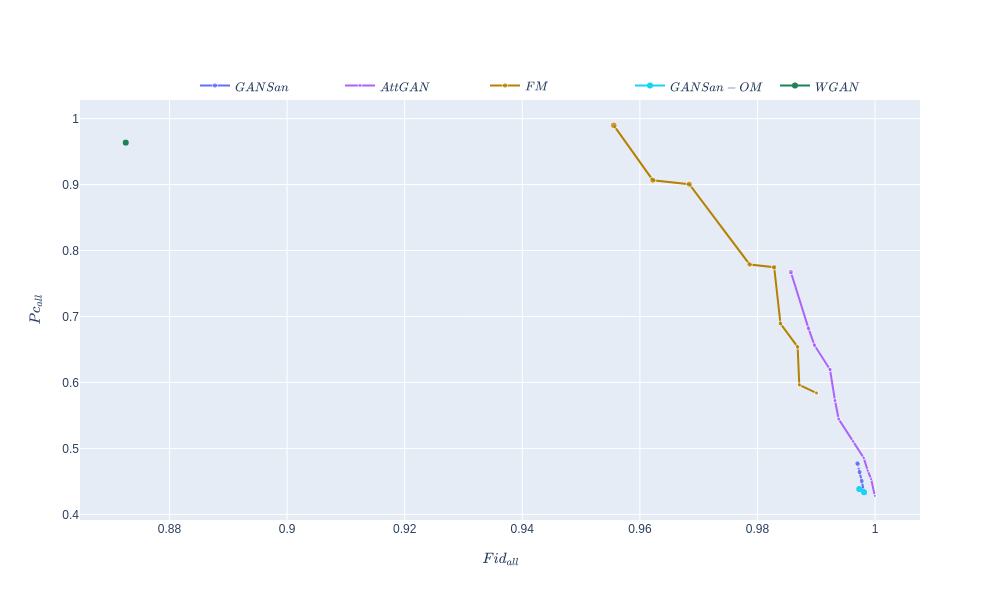}
    \caption{Results on the perspective of \attgan{} (Adult2 dataset), which is the transformation of data such that they resemble the privileged group. 
    }
    \label{fig:results-front-adult2-org-attgan}
\end{figure}

In the perspective of \attgan{} in Figure~\ref{fig:results-front-adult2-org-attgan}, we can observe that $\fairmapping{}$ exhibits an important negative slope with the increase of $\fid{}_{all}$. It is impossible for all approaches to map the data onto the privileged distribution while ensuring that the data are perfectly reconstructed. We can also note that $\fairmapping{}$ dominates $\attgan{}$ on fidelities lower than $0.989$. $\attgan{}$ is better for higher fidelities. Both approaches seem to form a continuous front, as we have observed in some other cases.

\begin{figure}[ht]
        \includegraphics[width=1\linewidth, height=0.45\textheight]{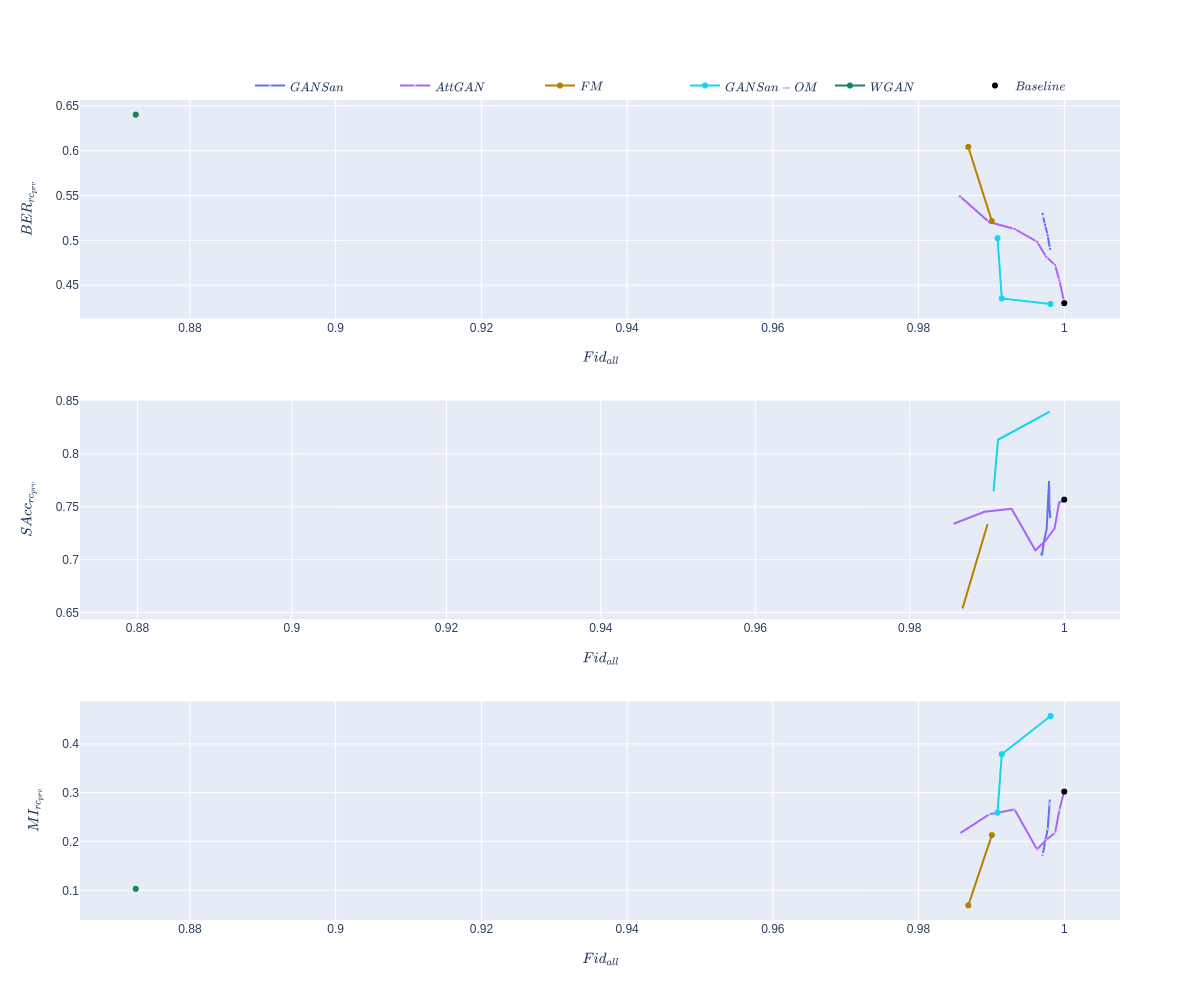}
    \caption{Results on the perspective of \gansan{} (Adult2 dataset), which is the transformation of data such that they resemble the privileged group. 
    }
    \label{fig:results-front-adult2-org-gansan}
\end{figure}
On the \gansan{} perspective (Figure~\ref{fig:results-front-adult2-org-gansan}), we can observe almost the same performances as with the single sensitive attribute of \gansan{}. 

\fairmapping{} can protect the sensitive information better than any other approach. The fidelity values obtained are close to the perfect one of $\fid_{all} = 1$ even though the reconstruction constraint is only applied on the privileged group. 
At the same time, \gansan{}, which reconstructs both groups achieves higher fidelities (at most $1.27\%$ greater than the fidelities of $\fairmapping{}$), but lower protection. This suggests that enforcing the reconstruction constraint only on one group (the privileged group for instance) while ensuring the protection with the original data version of the reconstructed group (i.e. the original privileged group) provide a better structure for protecting sensitive information. It is important to observe that $\gansan{}-OM$ failed to protect the sensitive attribute. Thus, to better protect the sensitive attribute, some degree of freedom (by not constraining the protected group) are necessary.
\fairmapping{} provides both of these requirements. That's why the protection is higher. At the same time, to avoid having the unconstrained group transformed into a completely random distribution, we guide the transformation by ensuring the mapping towards the privileged distribution.

\section{Standard deviation fairness results}
\label{sec:std-fairness-results}
This section presents the standard deviation results of the Table presented in the main section of our article. Table~\ref{tab:fairness-protection-results-std} presents the standard deviation of the fairness results, and Table~\ref{tab:diversity-std} the standard deviation of the diversity.
\begin{table}[]
\caption{Cross validation protection results for all experimented datasets. Standard deviation of fairness results presented in Table~\ref{tab:fairness-protection-results-mean}}
\label{tab:fairness-protection-results-std}
\resizebox{\columnwidth}{!}{%
\begin{tabular}{cccccccccc}
\cline{1-10}
\multicolumn{2}{c}{}                  & \multicolumn{2}{c}{Adult}                  & \multicolumn{2}{c}{Adult2}                 & \multicolumn{2}{c}{German}                 & \multicolumn{2}{c}{Lipton}                 \\ \cline{1-10} 
                              &       & Baseline & \fairmapping{} & Baseline & \fairmapping{} & Baseline & \fairmapping{} & Baseline & \fairmapping{} \\ \cline{2-10} 
\multirow{8}{*}{Metrics}
& $\ber{}_{rc_{prv}}$ &  $0.0022$ &  $0.0298$  &  $0.0073$  &  $0.0409$  & $0.0215$ &  $0.0480$ &  $0.0318$  &  $0.0202$ \\
& $\ber{}_{og_{prv}}$ &  $0.0022$ &  $0.0312$  &  $0.0073$  &  $0.0232$  & $0.0215$ &  $0.0325$ &  $0.0318$  &  $0.0166$ \\
& $\sac{}_{rc_{prv}}$ &  $0.0008$ &  $0.0198$  &  $0.0026$  &  $0.0199$  & $0.0172$ &  $0.0081$ &  $0.0318$  &  $0.0202$ \\
& $\sac{}_{og_{prv}}$ &  $0.0008$ &  $0.0223$  &  $0.0026$  &  $0.0053$  & $0.0172$ &  $0.0201$ &  $0.0318$  &  $0.0166$ \\
& $MI_{rc_{prv}}$ &  $0.0048$ &  $0.0275$  &  $0.0070$  &  $0.0361$  & $0.0120$ &  $0.0182$ &  $0.0461$  &  $0.0068$ \\
& $MI_{og_{prv}}$ &  $0.0048$ &  $0.0753$  &  $0.0070$  &  $0.0171$  & $0.0120$ &  $0.0296$ &  $0.0461$  &  $0.0120$ \\
& $\classification{}_{prot}$ &  $0.0000$ &  $0.1241$  &  $0.0000$  &  $0.3640$  & $0.0000$ &  $0.0972$ &  $0.0000$  &  $0.0877$ \\
& $Fid_{priv}$ &  $0.0001$ &  $0.0022$  &  $0.0001$  &  $0.0242$  & $0.0011$ &  $0.0071$ &  $0.0036$  &  $0.0010$ \\
\end{tabular}
}
\end{table}

\begin{table}[]
    \caption{Standard deviation of the diversity}
    \resizebox{\columnwidth}{!}{%
        \label{tab:diversity-std}
        \begin{tabular}{ccccccc}
        \hline
        \multirow{2}{*}{} & \multicolumn{2}{c}{Whole dataset} & \multicolumn{2}{c}{Protected group} & \multicolumn{2}{c}{Privileged group} \\ \cline{2-7} 
                          & Baseline       & fairmapping      & Baseline        & Fairmapping       & Baseline        & Fairmapping        \\ \hline
        Adult2 & 0.0004 & 0.0221 & 0.0002 & 0.0301 & 0.0006 & 0.0192  \\
        Adult & 0.0002 & 0.0049 & 0.0003 & 0.0074 & 0.0005 & 0.0031  \\
        Lipton & 0.0046 & 0.0100 & 0.0103 & 0.0145 & 0.0014 & 0.0064  \\
        German & 0.0010 & 0.0062 & 0.0066 & 0.0096 & 0.0005 & 0.0069  \\
        \end{tabular}
    }
\end{table}

\section{Fairness classification use case results}
\label{sec:fair-res}
As presented in section~\ref{sec:fairnes-results}, the fairness of our approach can be measure on the \textit{fair classification} use case. 

In this setup, the objective is to build a classification mechanism that is free from discrimination by using the transformed dataset to train a classifier, while testing the model with original (unmodified) data. The rationale of this scenario is to circumvent unusual or unethical issues that are the result of the modifications introduced by the transformation procedure (\emph{e.g.}, to prevent the inference of groups, the value of the attribute \textit{infraction,} from a traffic ticket to a heinous crime), which would not be accepted in critical domains such as predictive justice. Thus, by training the classifier on transformed data in which the sensitive attribute is hidden, it will learn to predict the decision without any information on the sensitive attribute. By using the original test set on the trained classifier, the practitioner avoids those risks of unethical outcomes.

\begin{figure}[ht]
        \includegraphics[width=1\linewidth, height=0.15\textheight]{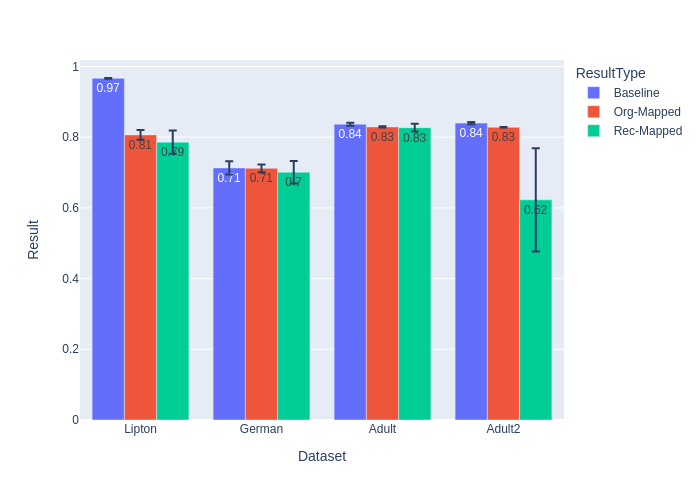}
    \caption{Accuracies achieved with the MLP classifier on the Fair classification scenario. The black vertical bar indicates the standard deviation across all computed folds.
    }
    \label{fig:results-acc-fair-clf}
\end{figure}

\begin{figure}[ht]
        \includegraphics[width=1\linewidth, height=0.15\textheight]{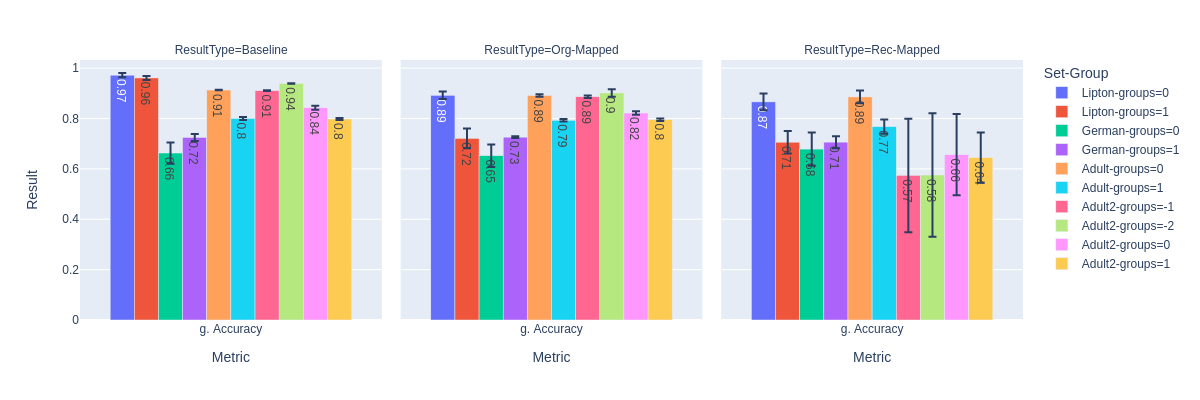}
    \caption{Accuracies computed in each of the dataset group, for the fair classification scenario.
    }
    \label{fig:results-gr-acc-fair-clf}
\end{figure}
In Figure~\ref{fig:results-acc-fair-clf}, we showcase the accuracy achieved with the fair classification scenario. The fair classification accuracies are quite similar to those in the other scenario. This suggests that training a classifier on either the original or transformed and reconstructed dataset to predict the original decision using either the original or the transformed and reconstructed datasets will yield the same results. These datasets therefore share similar characteristics which allow a classifier to behave similarly in all cases.
The group accuracy in Figure~\ref{fig:results-gr-acc-fair-clf} also confirms our observation on the similarity of predictions. 

\begin{figure}[ht]
        \includegraphics[width=1\linewidth, height=0.15\textheight]{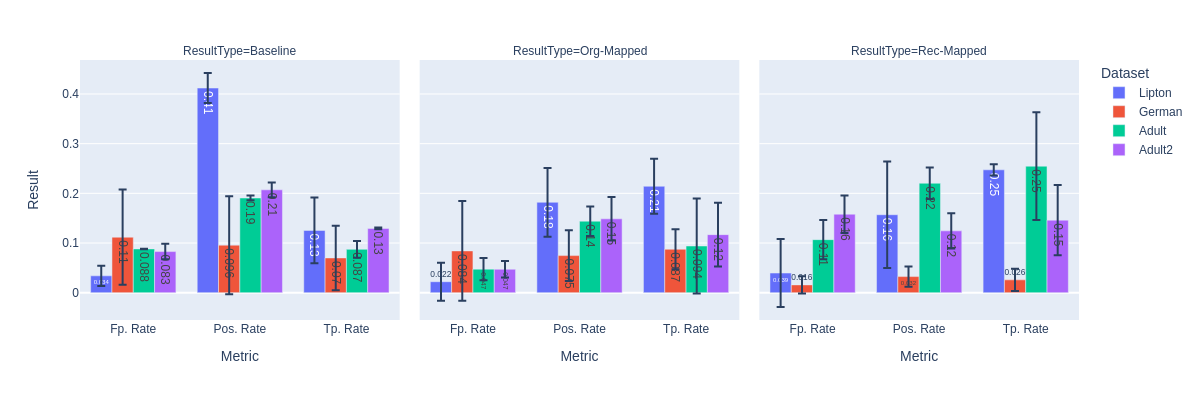}
    \caption{Fairness results for the fair classification.
    }
    \label{fig:results-fair-fair-clf}
\end{figure}
\begin{figure}[ht]
        \includegraphics[width=1\linewidth, height=0.15\textheight]{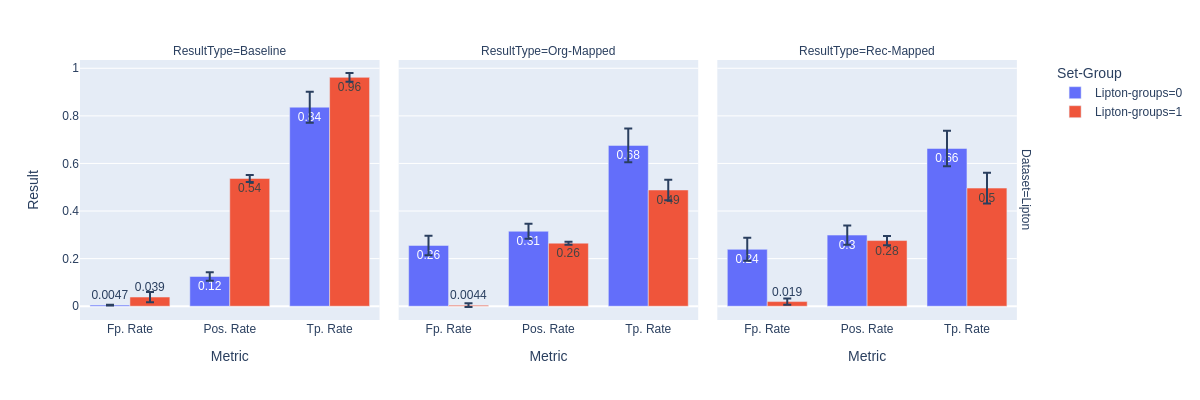}
        \includegraphics[width=1\linewidth, height=0.15\textheight]{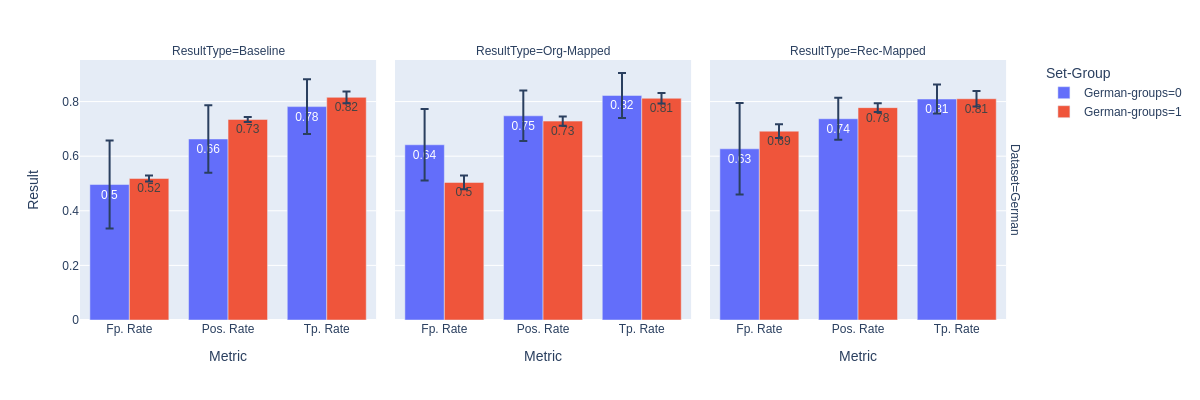}
        \includegraphics[width=1\linewidth, height=0.15\textheight]{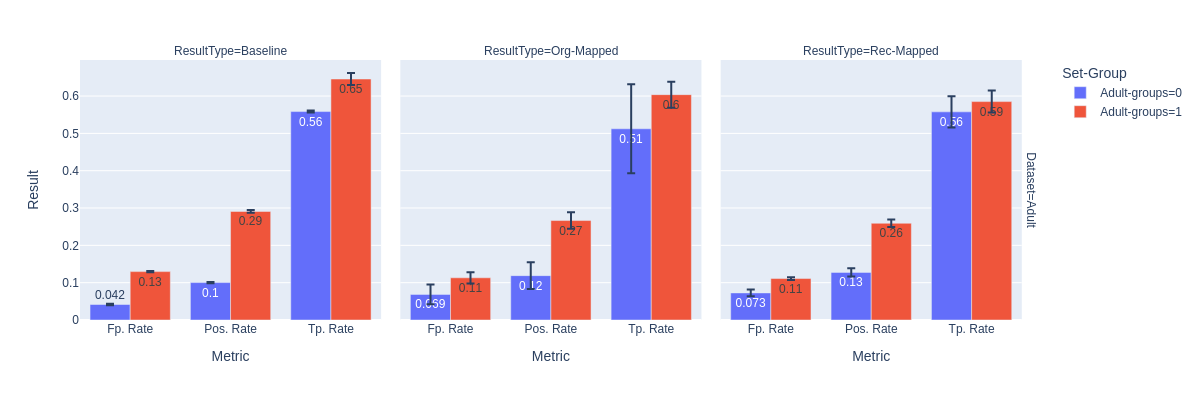}
        \includegraphics[width=1\linewidth, height=0.15\textheight]{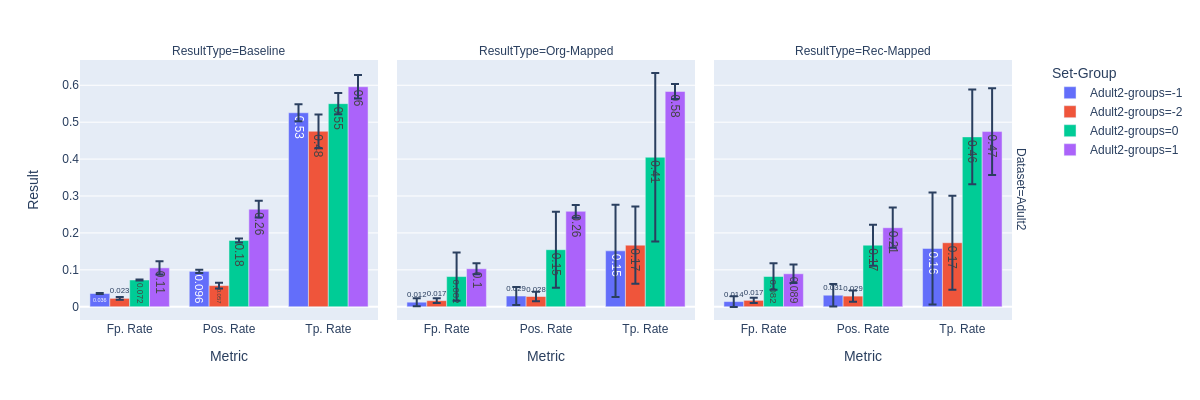}
    \caption{Accuracy of the task decision at each group level for the scenario data publishing.
    }
    \label{fig:results-fair-gr-fair-clf}
\end{figure}

Concerning the fairness results (Figure~\ref{fig:results-fair-fair-clf}), they are not as significantly improved as observed in the other scenario. As a matter of fact, we can observe in Figure~\ref{fig:results-fair-gr-fair-clf} that the metrics on the transformed sets follows the same trend as the baseline.

\section{Metrics computed within each groups}
\label{sec:metrics-per-group}
In this section, we present the fairness and accuracies computed presented in the main article at the group level basis.

\begin{figure}[ht]
        \includegraphics[width=1\linewidth, height=0.15\textheight]{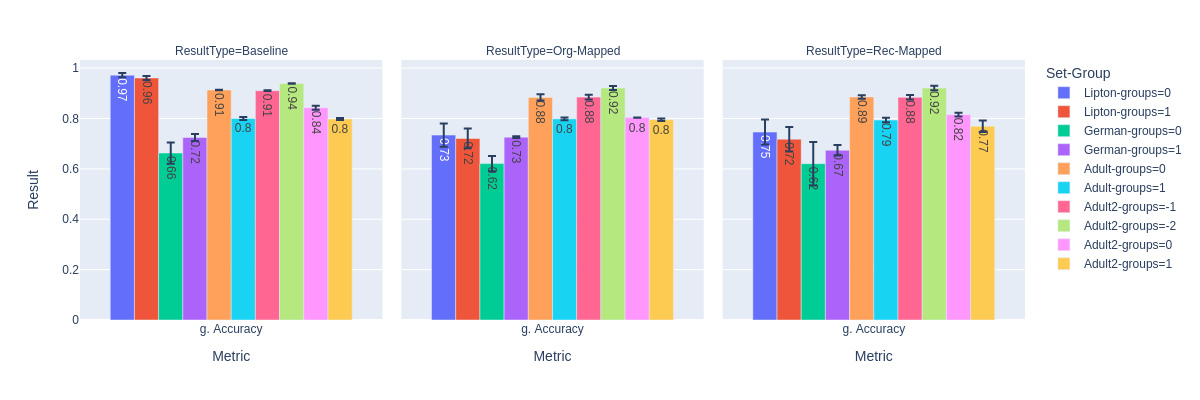}
    \caption{Accuracy of the task decision at each group level for the scenario data publishing.
    }
    \label{fig:results-gr-acc-data-pub}
\end{figure}

Figure~\ref{fig:results-gr-acc-data-pub} present the accuracy of the task decision at each group level for the scenario data publishing.

\begin{figure}[ht]
        \includegraphics[width=1\linewidth, height=0.15\textheight]{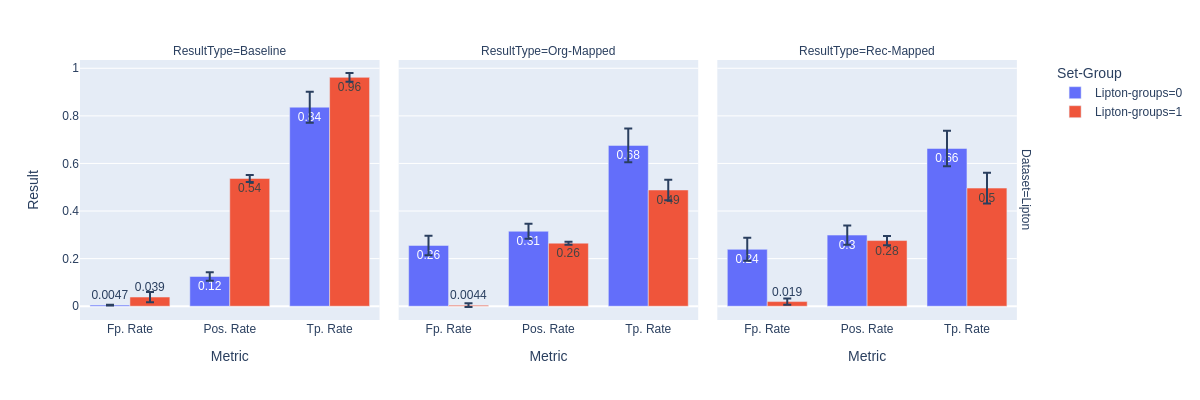}
        \includegraphics[width=1\linewidth, height=0.15\textheight]{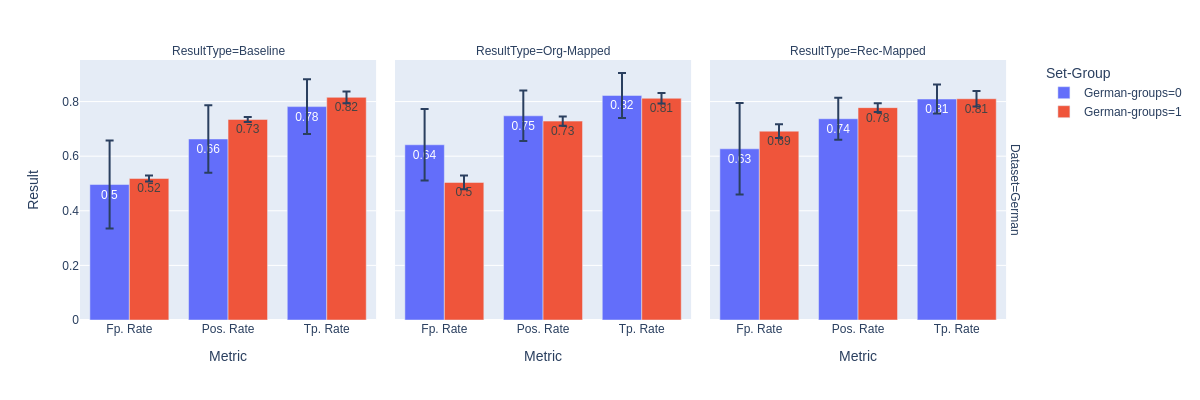}
        \includegraphics[width=1\linewidth, height=0.15\textheight]{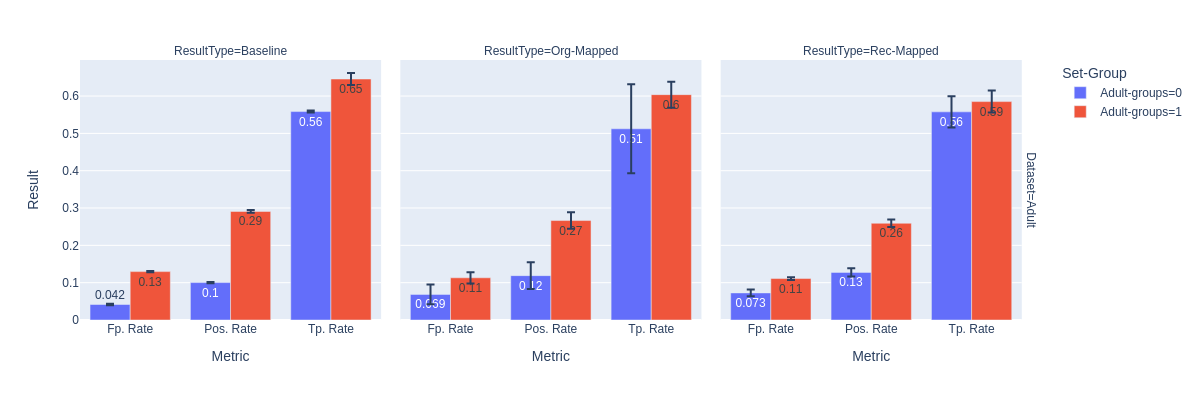}
        \includegraphics[width=1\linewidth, height=0.15\textheight]{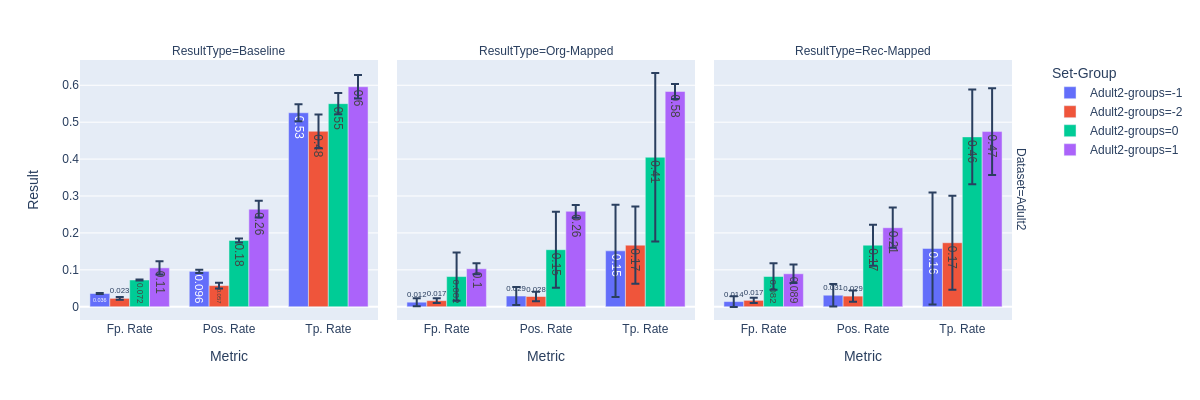}
    \caption{Accuracy of the task decision at each group level for the scenario data publishing.
    }
    \label{fig:results-fair-gr-data-pub}
\end{figure}

Figure~\ref{fig:results-fair-gr-data-pub} presents the fairness metrics computed in each group of the dataset

\begin{figure}[ht]
        \includegraphics[width=1\linewidth, height=0.15\textheight]{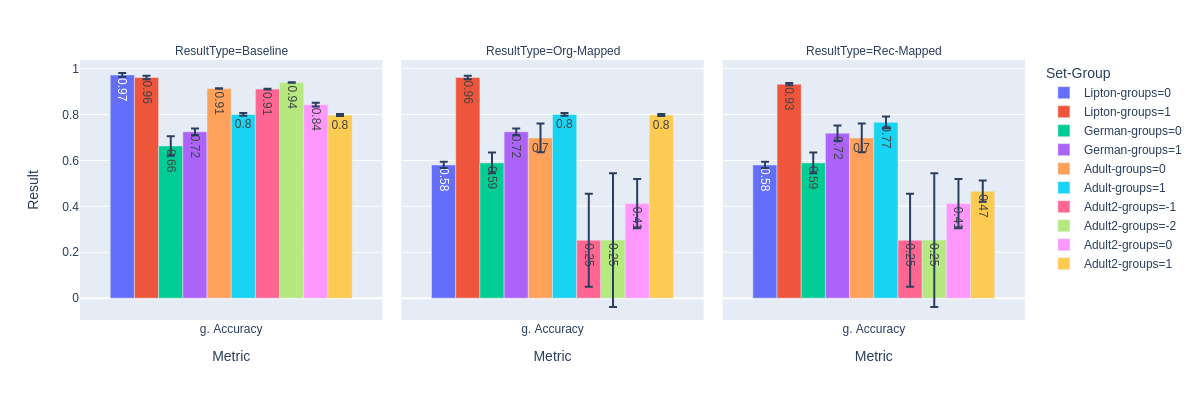}
    \caption{Accuracy of the task decision at each group level for the scenario local sanitization.
    }
    \label{fig:results-gr-acc-local-san}
\end{figure}

Figure~\ref{fig:results-gr-acc-local-san} present the accuracy of the task decision at each group level for the scenario Local sanitization.

Figure~\ref{fig:results-gr-acc-local-san} presents the fairness metrics computed in each group of the dataset

\section{Lower dimensional visualisation of distributions}
\label{sec:distribution-viz}

In this appendix, we present some lower dimensional visualisation of our results.

\begin{figure}[ht]
        \includegraphics[width=1\linewidth, height=0.3\textheight]{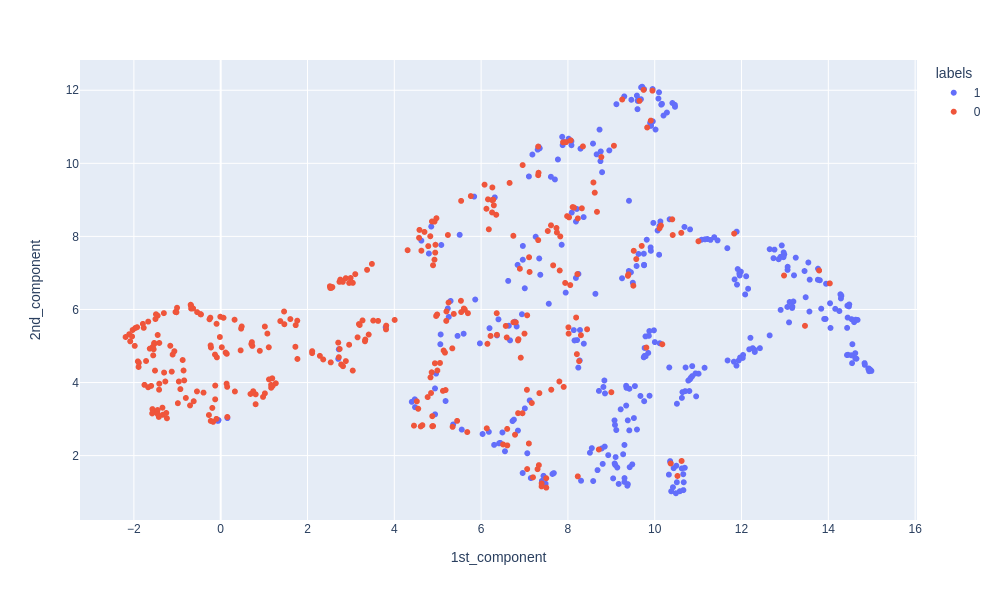}
    \caption{Lower dimensional plot of the original Lipton dataset.}
    \label{fig:dim-red-lipton-orig}
\end{figure}
\begin{figure}[ht]
        \includegraphics[width=1\linewidth, height=0.3\textheight]{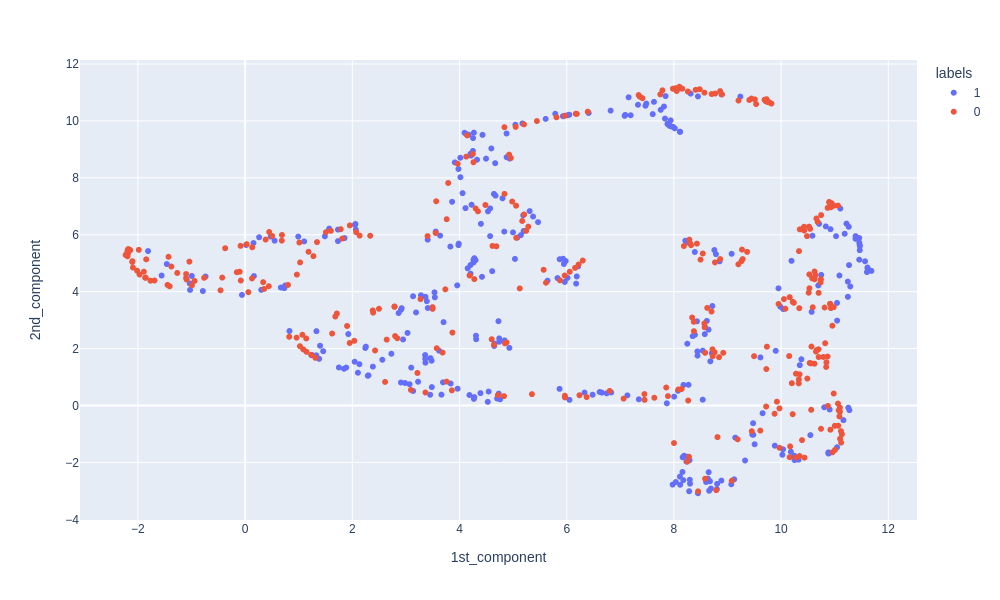}
    \caption{Lower dimensional plot of the Lipton dataset with the transformed protected group data.}
    \label{fig:dim-red-lipton-O-M}
\end{figure}
\begin{figure}[ht]
        \includegraphics[width=1\linewidth, height=0.3\textheight]{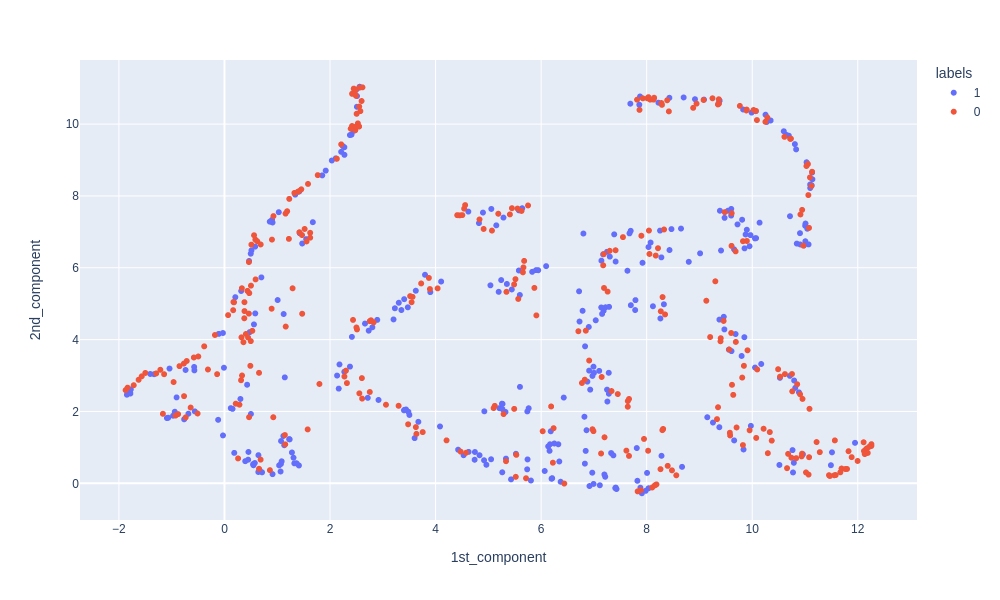}
    \caption{Lower dimensional plot of the Lipton dataset with the transformed protected data and the reconstructed data from the privileged group.}
    \label{fig:dim-red-lipton-R-M}
\end{figure}
In Figure~\ref{fig:dim-red-lipton-orig} we present the lower dimensional plot of the original Lipton dataset, while~\ref{fig:dim-red-lipton-R-M} represent the lower dimensional of the transformed protected group and the reconstructed privileged data. Figure~\ref{fig:dim-red-lipton-O-M} present the same plot, but with the original privileged distribution. We can observe that the transformed protected distribution overlap the privileged distribution, which means that the transformation is successful. 

\begin{figure}[ht]
        \includegraphics[width=1\linewidth, height=0.3\textheight]{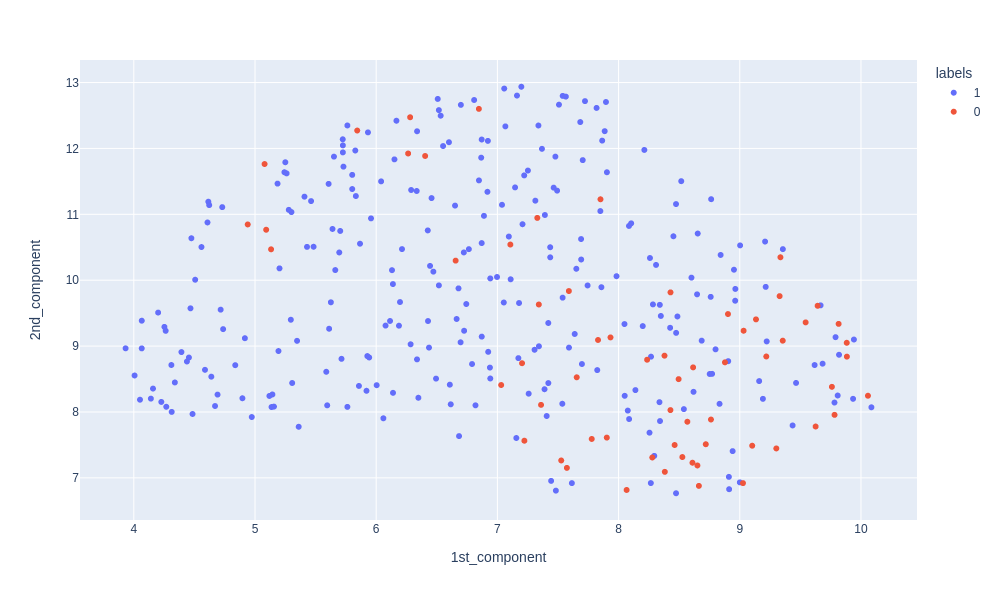}
    \caption{Lower dimensional plot of the original Lipton dataset.}
    \label{fig:dim-red-german-orig}
\end{figure}
\begin{figure}[ht]
        \includegraphics[width=1\linewidth, height=0.3\textheight]{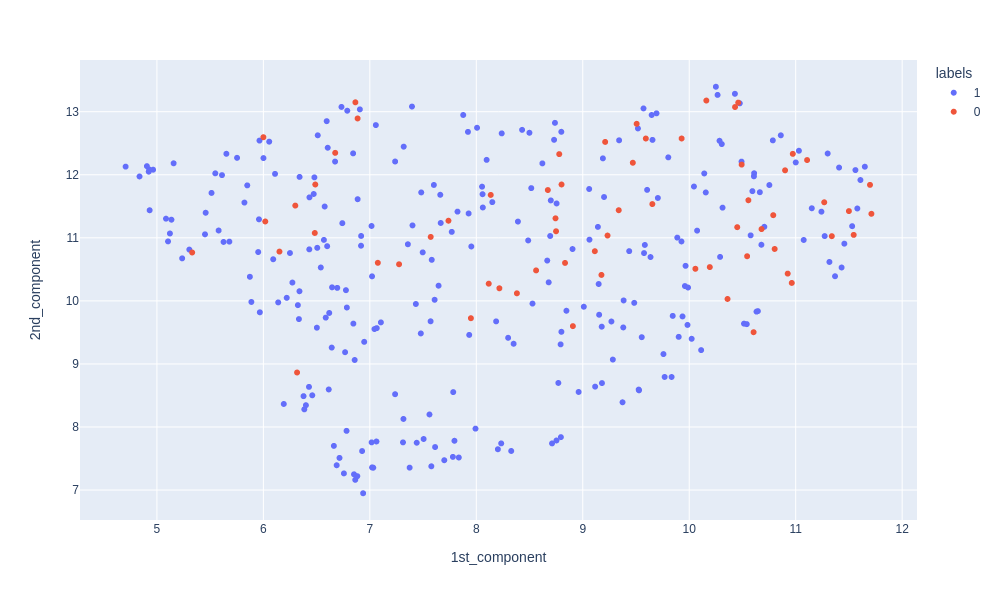}
    \caption{Lower dimensional plot of the Lipton dataset with the transformed protected group data.}
    \label{fig:dim-red-german-R-M}
\end{figure}
On German Credit (Figure~\ref{fig:dim-red-german-orig} and Figure~\ref{fig:dim-red-german-R-M}) the transformation impact is less visible since the original protected distribution is not easily distinguishable from the privileged one. Nevertheless, we can observe a slight improvement in the location of the protected distribution, which is more embedded within the target distribution.

\begin{figure}[ht]
        \includegraphics[width=1\linewidth, height=0.3\textheight]{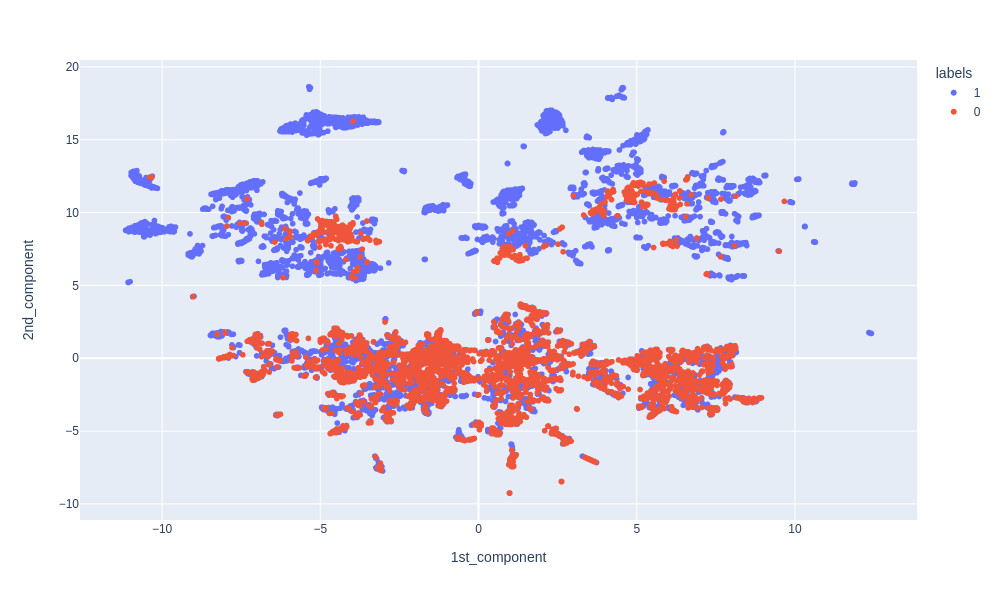}
    \caption{Lower dimensional plot of the original Lipton dataset.}
    \label{fig:dim-red-adult-orig}
\end{figure}
\begin{figure}[ht]
        \includegraphics[width=1\linewidth, height=0.3\textheight]{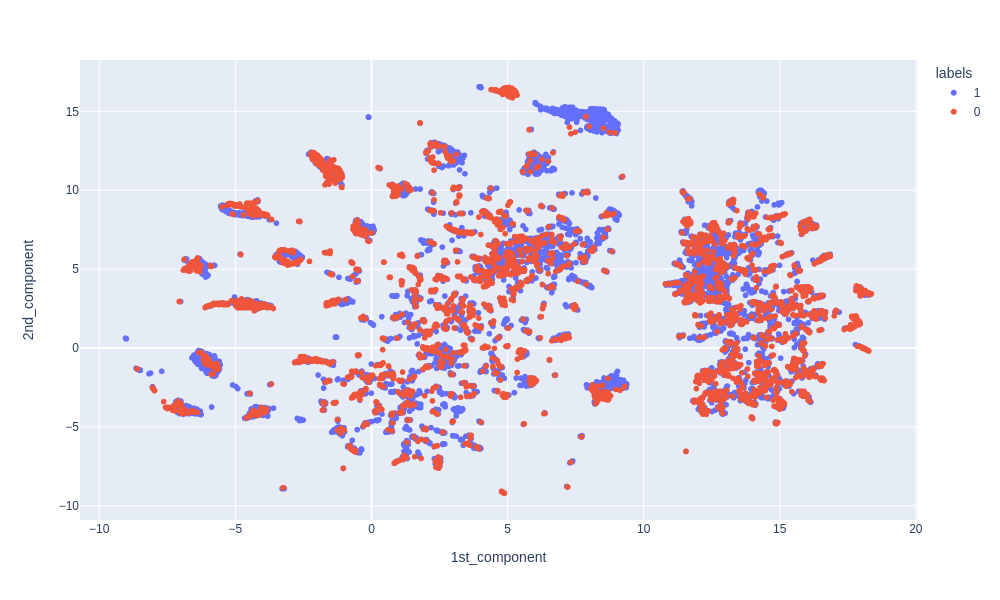}
    \caption{Lower dimensional plot of the Lipton dataset with the transformed protected group data.}
    \label{fig:dim-red-adult-R-M}
\end{figure}
The results on the dataset Adult shows an improvement on the superposition of the transformed protected distribution onto the privileged one (figures~\ref{fig:dim-red-adult-orig}, \ref{fig:dim-red-adult-R-M}).

\bibliographystyle{alpha}
\bibliography{Bib}
\end{document}